\documentclass[11pt]{article}

\usepackage[pdftex]{graphicx}
\usepackage{amsmath,amssymb,subfigure,epsfig,bbm,stmaryrd,MnSymbol,mathrsfs,url}
\usepackage{pstricks}
\usepackage{subfigure}
\graphicspath{{../figures/}} \DeclareGraphicsExtensions{.pdf}
\usepackage{amstext}
\usepackage{array,xcolor}
\usepackage{color,soul}
\usepackage{hyperref}
\DeclareMathOperator{\sign}{sign}
\DeclareMathAlphabet{\mathpzc}{OT1}{pzc}{m}{it}
\DeclareFontFamily{OT1}{pzc}{}
\DeclareFontShape{OT1}{pzc}{m}{it}{<-> s * [1.2] pzcmi7t}{}
\DeclareMathAlphabet{\mathpzc}{OT1}{pzc}{m}{it}
\DeclareMathAlphabet{\mathbbit}{U}{bbm}{m}{sl}

\usepackage{algorithm,mdframed}
\usepackage{algorithmic}

\usepackage{amsmath}
\usepackage[all,cmtip]{xy}
\usepackage{chemfig}

\usepackage{booktabs,makecell}

\usepackage{diagbox}

\newcommand{\balpha}{\boldsymbol{\alpha}}

\newcommand{\Ip}{\mathcal{I}_\oplus}
\newcommand{\In}{\mathcal{I}_\ominus}

\newcommand{\Beta}{{\boldsymbol{\beta}}}

\newcommand{\Blambda}{{\boldsymbol{\Lambda}}}
\newcommand{\Bz}{{\boldsymbol{z}}}
\newcommand{\Ba}{{\boldsymbol{a}}}

\newcommand{\By}{{\boldsymbol{y}}}
\newcommand{\Bh}{{\boldsymbol{h}}}
\newcommand{\supp}{{\mbox{supp}}}
\newcommand{\rank}{{\mbox{rank}}}
\usepackage{mathtools,amssymb,graphicx}

\usepackage[T1]{fontenc}
\usepackage[utf8]{inputenc}
\usepackage[font=small,labelfont=bf,tableposition=top]{caption}

\usepackage{easybmat}
\usepackage{multirow,bigdelim}

\usepackage{makeidx,overpic}
\makeindex

\newcounter{example}

\newtheorem{theorem}{Theorem}
\newtheorem{lemma}{Lemma}

\begin{document}

\title{\vspace{-2cm}\bf{Learning Shapes by Convex Composition}}
\author{Alireza Aghasi and Justin Romberg\thanks{School of Electrical and Computer Engineering, Georgia Institute of Technology, Atlanta, GA. Emails: {\tt aaghasi@ece.gatech.edu} and {\tt jrom@ece.gatech.edu}.}}

\date{}    

\maketitle

\begin{abstract}
We present a mathematical and algorithmic scheme for learning the principal geometric elements in an image or 3D object.  We build on recent work that convexifies the basic problem of finding a combination of a small number shapes that overlap and occlude one another in such a way that they ``match'' a given scene as closely as possible.  This paper derives general sufficient conditions under which this convex shape composition identifies a target composition.  From a computational standpoint, we present two different methods for solving the associated optimization programs.  The first method simply recasts the problem as a linear program, while the second uses the alternating direction method of multipliers with a series of easily computed proximal operators.  
Finally, we present numerical experiments that use the framework to perform image segmentation, optical character recognition, and find multiresolution geometrical descriptions of 3D objects.

\end{abstract}

{\bf Keywords:} Shape Composition, Object Learning, Nonlinear Sparse Recovery, Geometric Packing Problem

\section{Introduction}\label{sec:introduction}

This paper develops a theory and algorithmic methods for decomposing shapes in a two- or three-dimensional scene as a composition of fixed sub-shapes.  We show how these methods can be applied in fundamental image processing and computer vision problems, where we are estimating a shape or region of interest about which some level of prior information is available \cite{aghasi2013sparse, leventon2000statistical, tsai2003shape, aghasi2015convex}.  Example applications include optical character recognition (OCR), where the objects of interest (words) are composed of simpler elements (letters), and characterization of overlapping or occluded objects in computer vision.  Beyond the scope of imaging, these methods can also be used to solve packing problems concerned with the arrangement of given objects inside a specified container \cite{fowler1981optimal}. 
%

Our method, described in full in the next section, is to construct shapes by super-imposing indicator functions and taking the positive part.  Thus places the problem of shape decomposition firmly in the realm of applied harmonic analysis --- we are searching for combinations of elements in a dictionary that explain the observed image.  Finding the best fit that also uses a small number of shapelets is posed as an optimization program (\eqref{eq5} below) that has a natural convex relaxation (\eqref{eq5x} below).

This method gives us a new way to regularize inverse problems in computer vision and computational imaging.  In image reconstruction problems, it is common to  penalize the energy in the reconstruction (Tikhonov), its smoothness (total variation), or its sparsity in a linear transform domain.  In object recovery problems,  penalties on the volume or surface of the reconstructions  are the most standard \cite{chan2001active}.  Here, we are seeking objects that can be constructed through standard shape operations from the smallest number of building blocks.

\begin{figure}[t]
\centering
\includegraphics[width=4in,height = 3.5in]{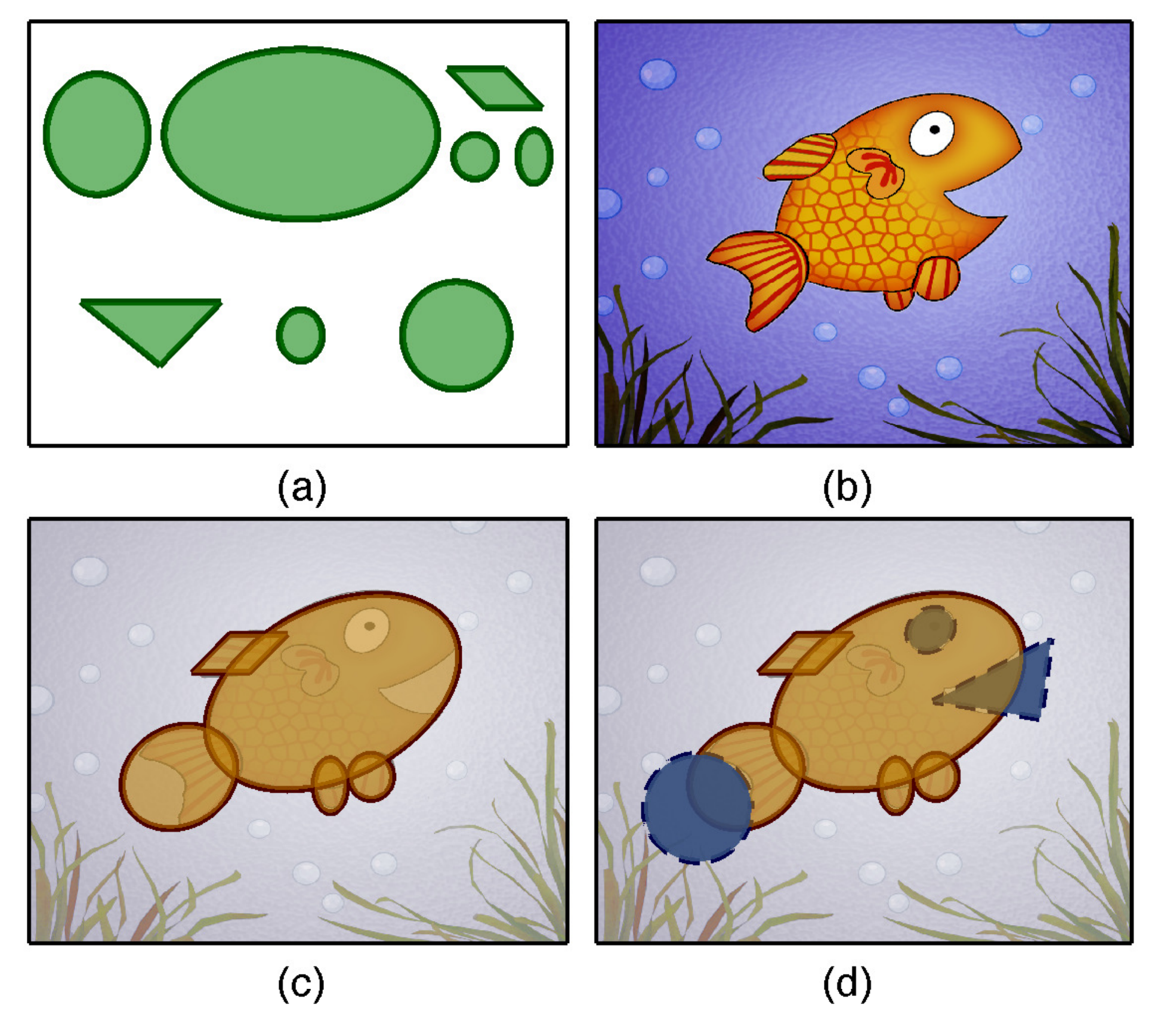}\vspace{-.4cm}
\caption{(a) Reference prototype shapes; (b) A reference image; (c) Object characterization using set union; (d) Finer characterization using union and set difference}
\label{fig_intro}
\end{figure}

\begin{figure}[ht!]
\centering \begin{tabular}{cccc}
\begin{overpic}[trim={0 4cm  0 0},clip, width=1.57in]{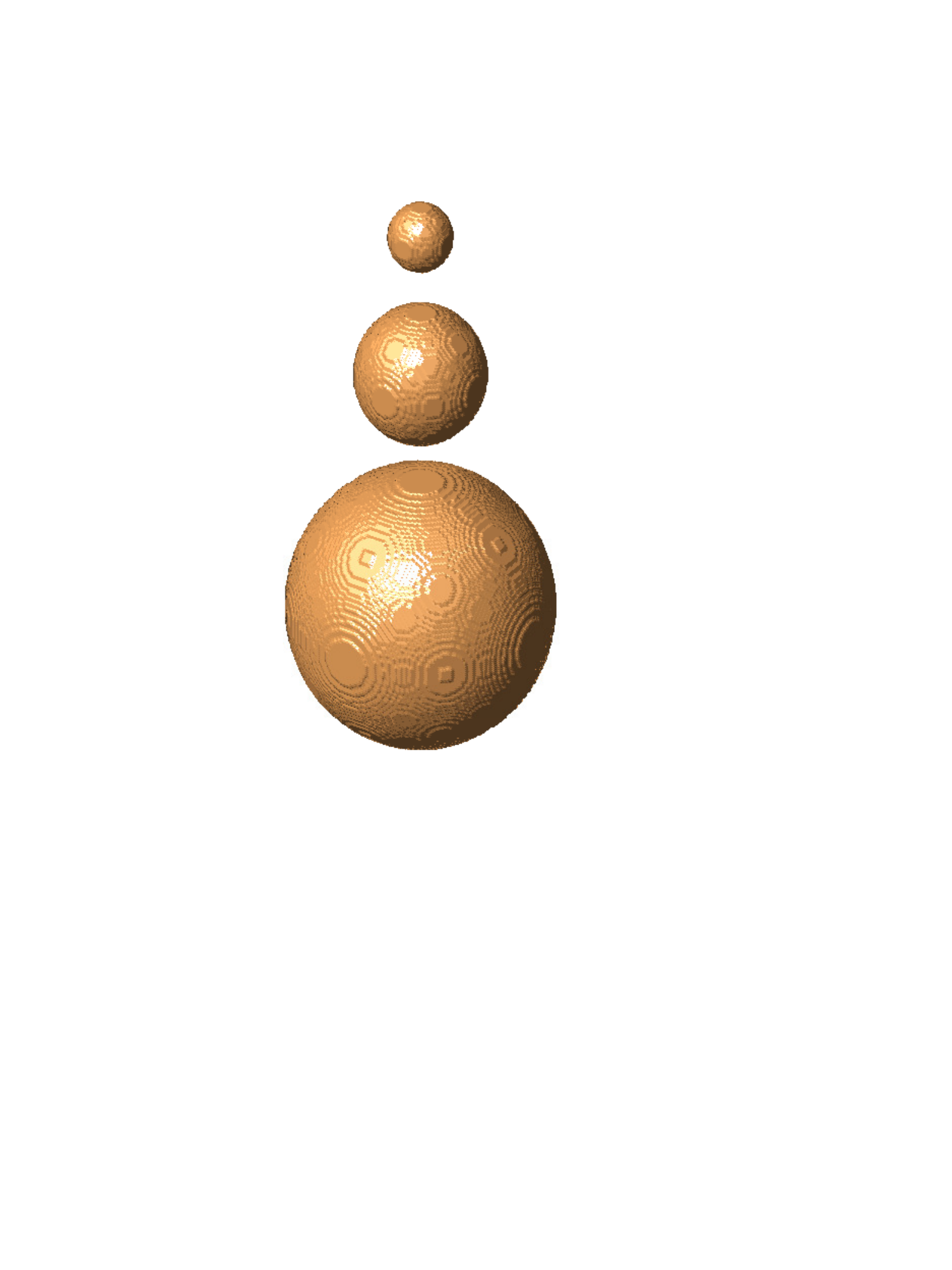}
 \put (37,-2) {\scriptsize(a)} \end{overpic}
 &
\hspace{-1cm}\begin{overpic}[trim={0 -1cm  0 0},clip,width=1.57in]{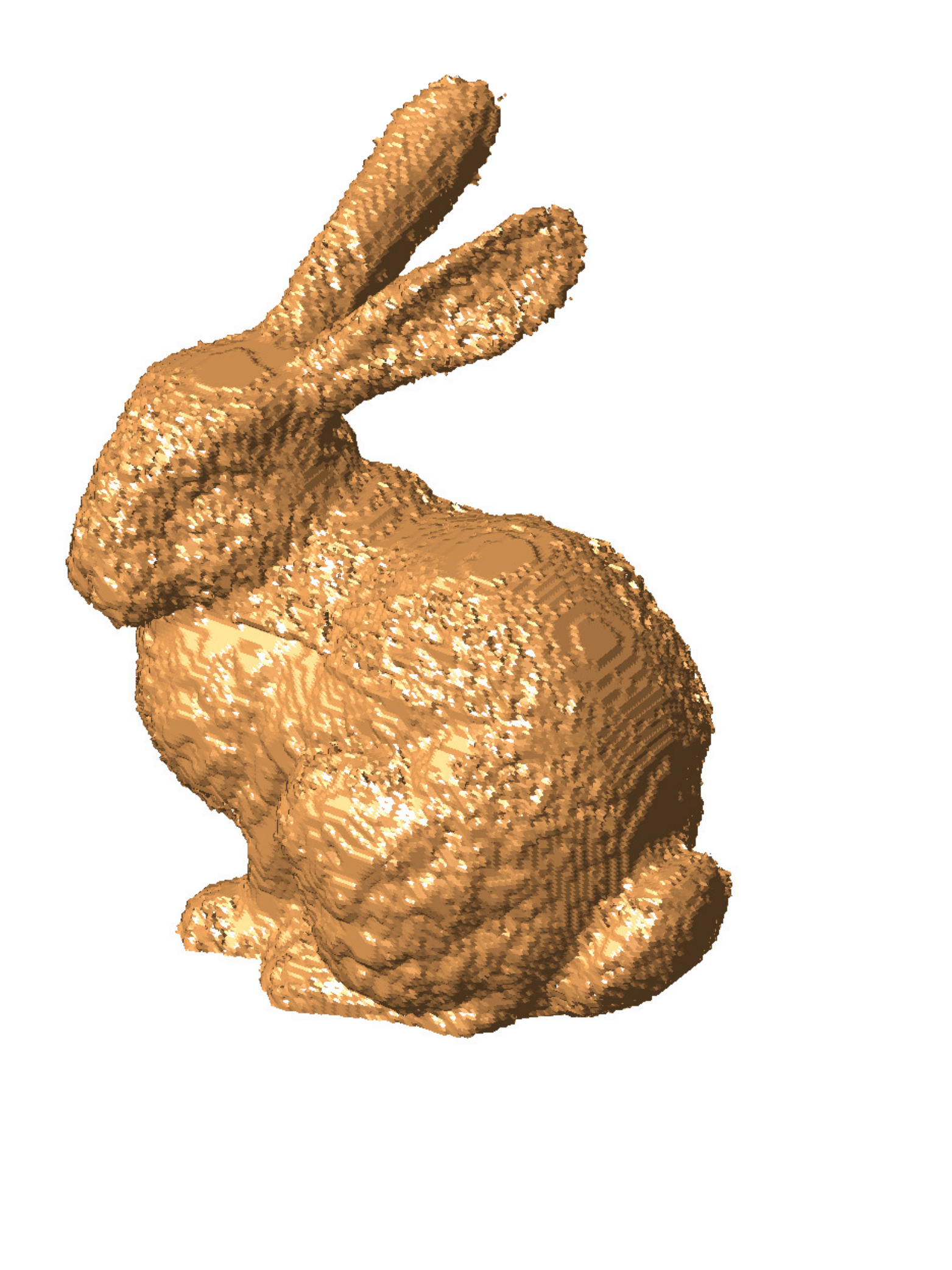} \put (30,-2) {\scriptsize(b)} \end{overpic} 
&
\hspace{-1cm}\begin{overpic}[trim={0 -1cm  0 0},clip,width=1.57in]{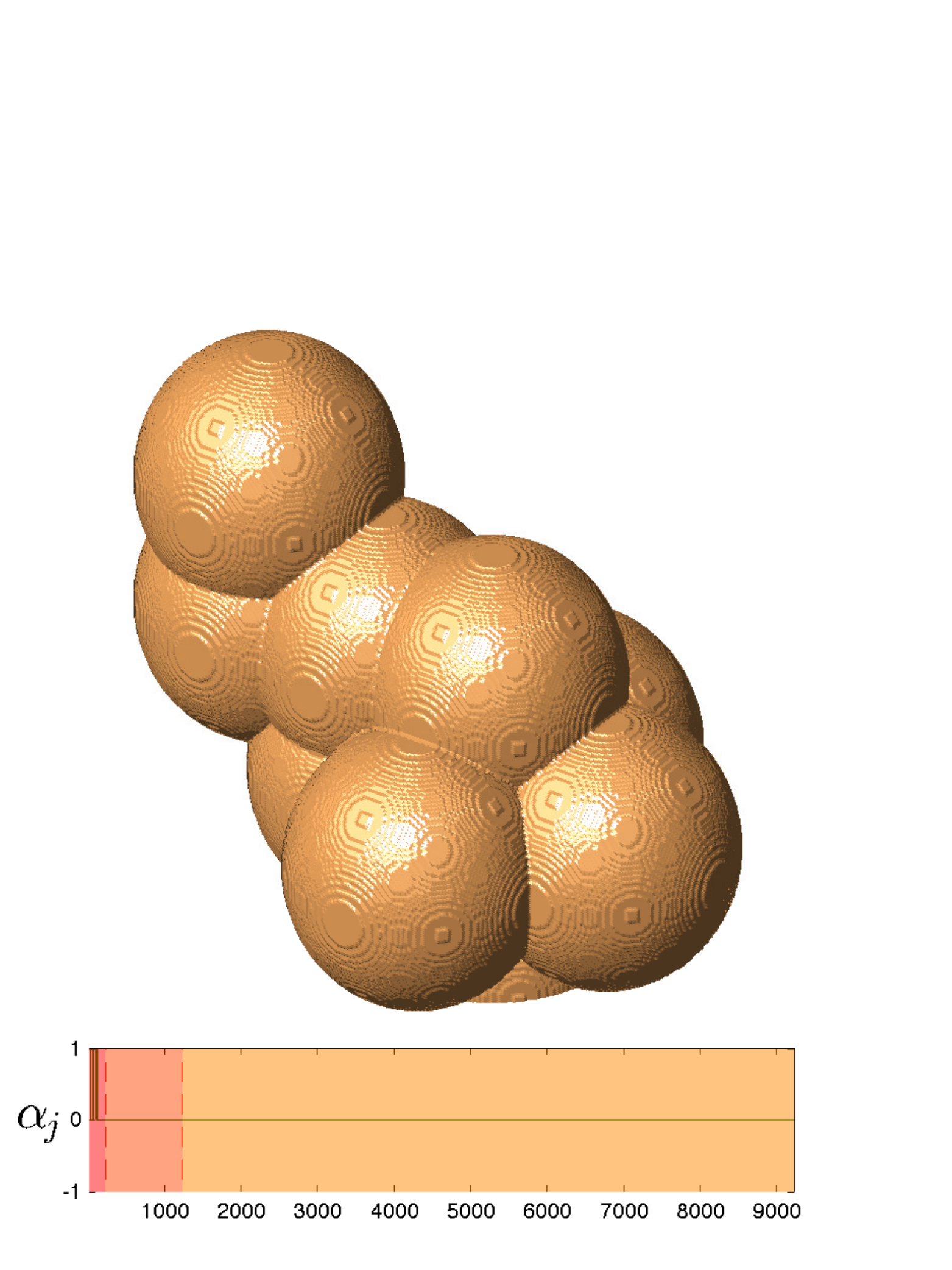}\put (30,-2) {\scriptsize(c)} \end{overpic} 
&
\hspace{-1cm}\begin{overpic}[trim={0 -1cm  0 0},clip,width=1.57in]{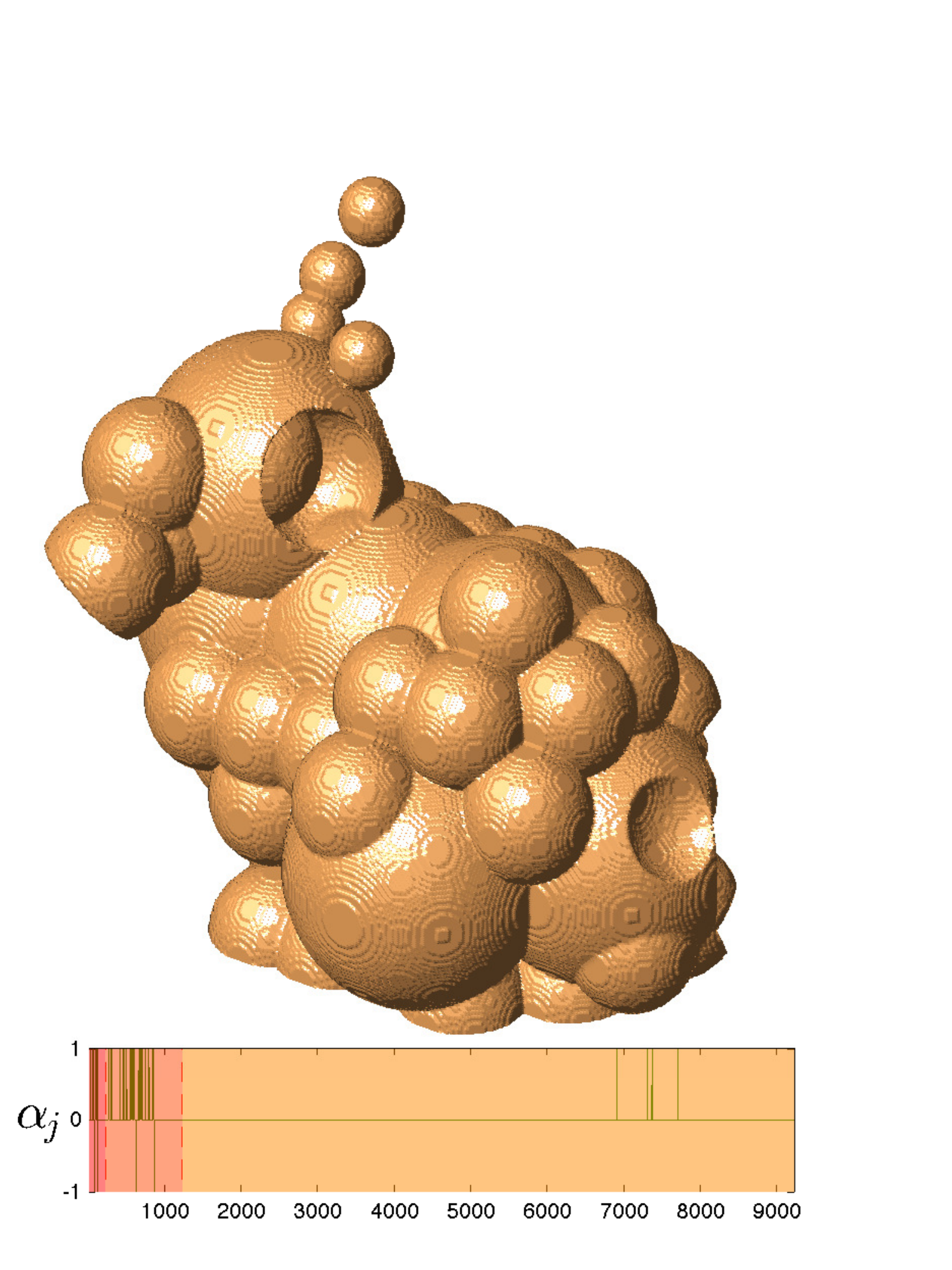}\put (30,-2) {\scriptsize(d)} \end{overpic}
\\
\begin{overpic}[trim={0 -1cm  0 0},clip,width=1.57in]{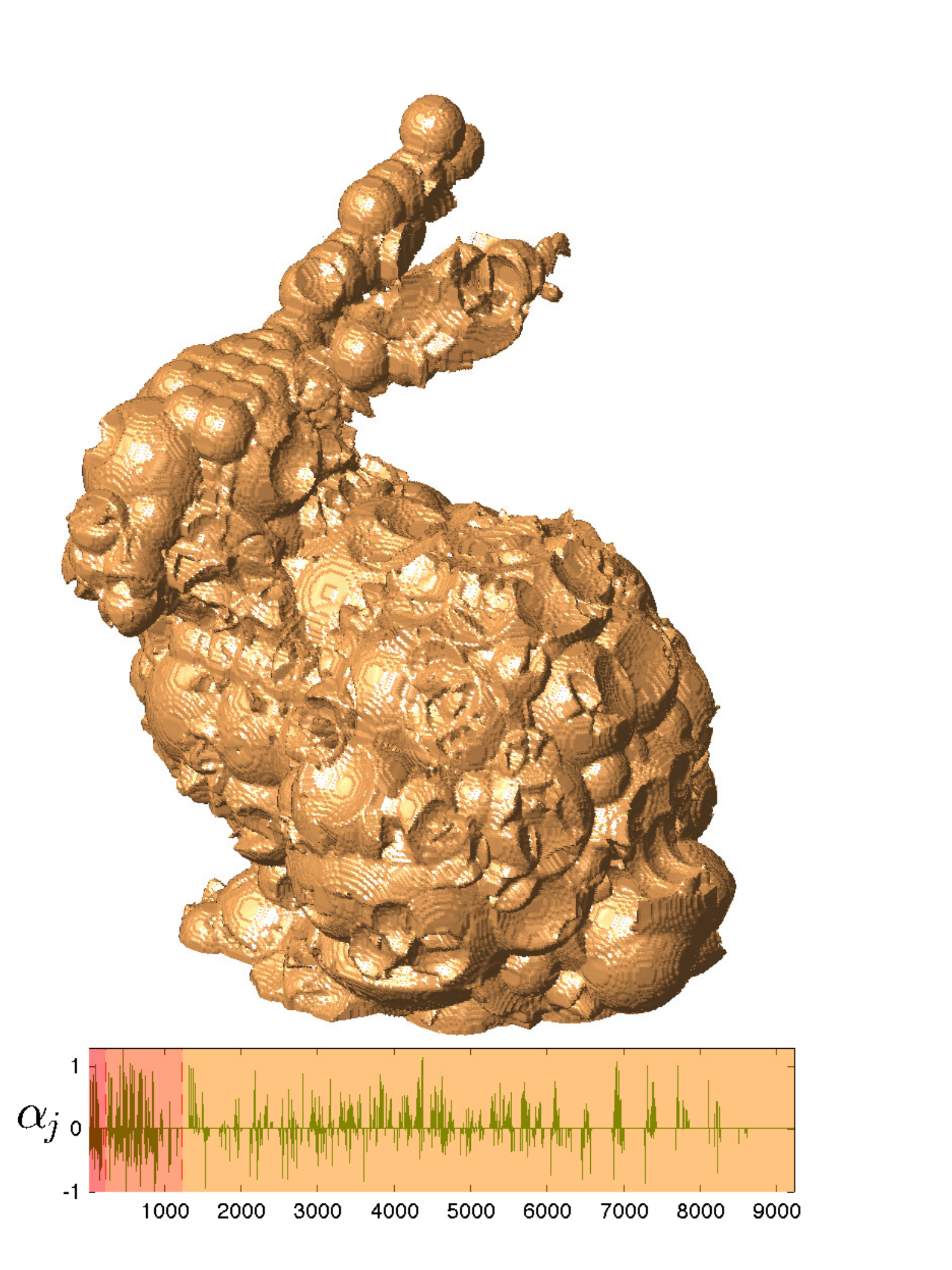}\put (30,-2) {\scriptsize(e)} \end{overpic}
&
\hspace{-1cm}\begin{overpic}[trim={0 -1cm  0 0},clip,width=1.57in]{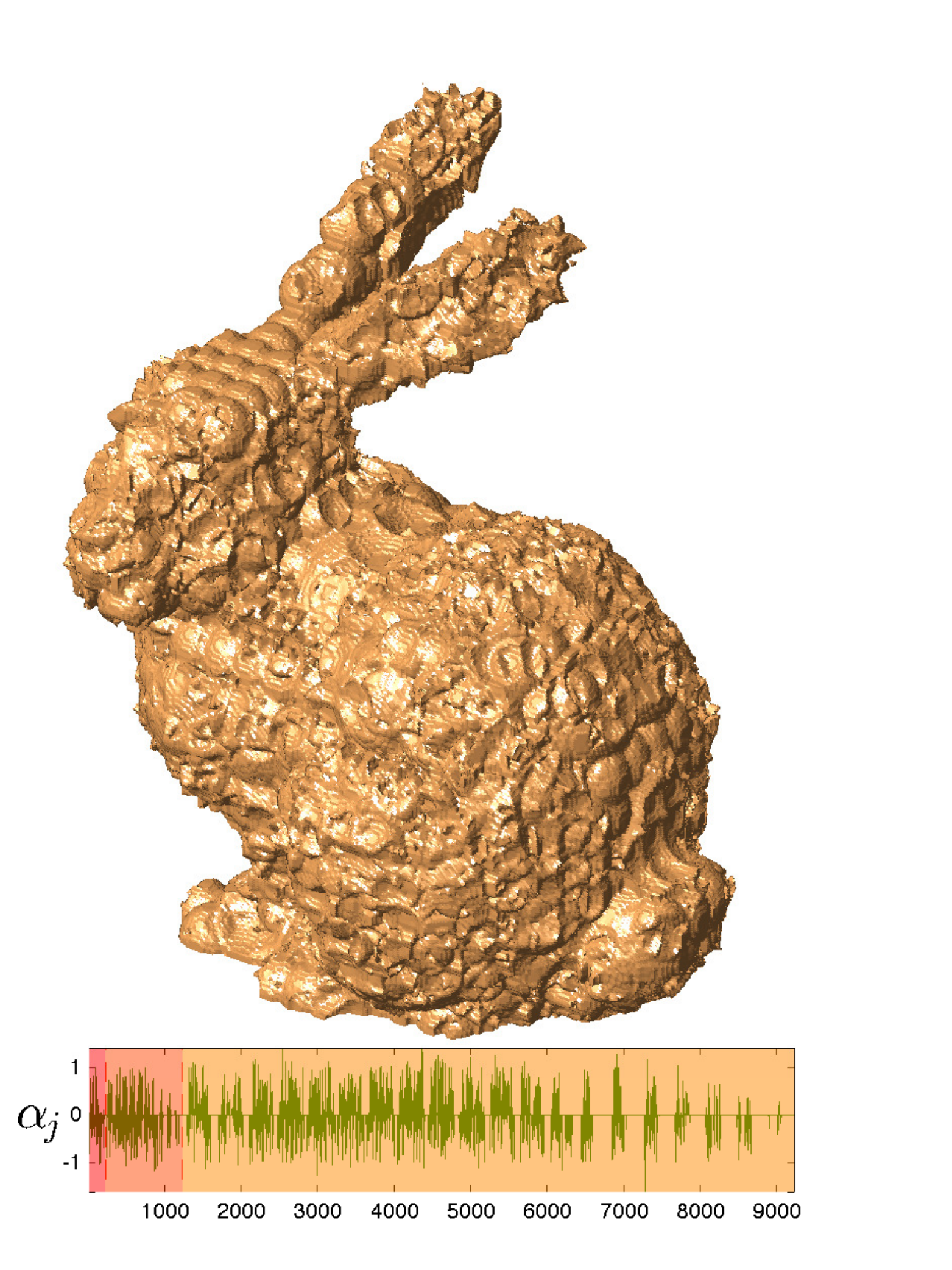}\put (30,-2) {\scriptsize(f)} \end{overpic}
&
\hspace{-1cm}\begin{overpic}[trim={0 -1cm  0 0},clip,width=1.57in]{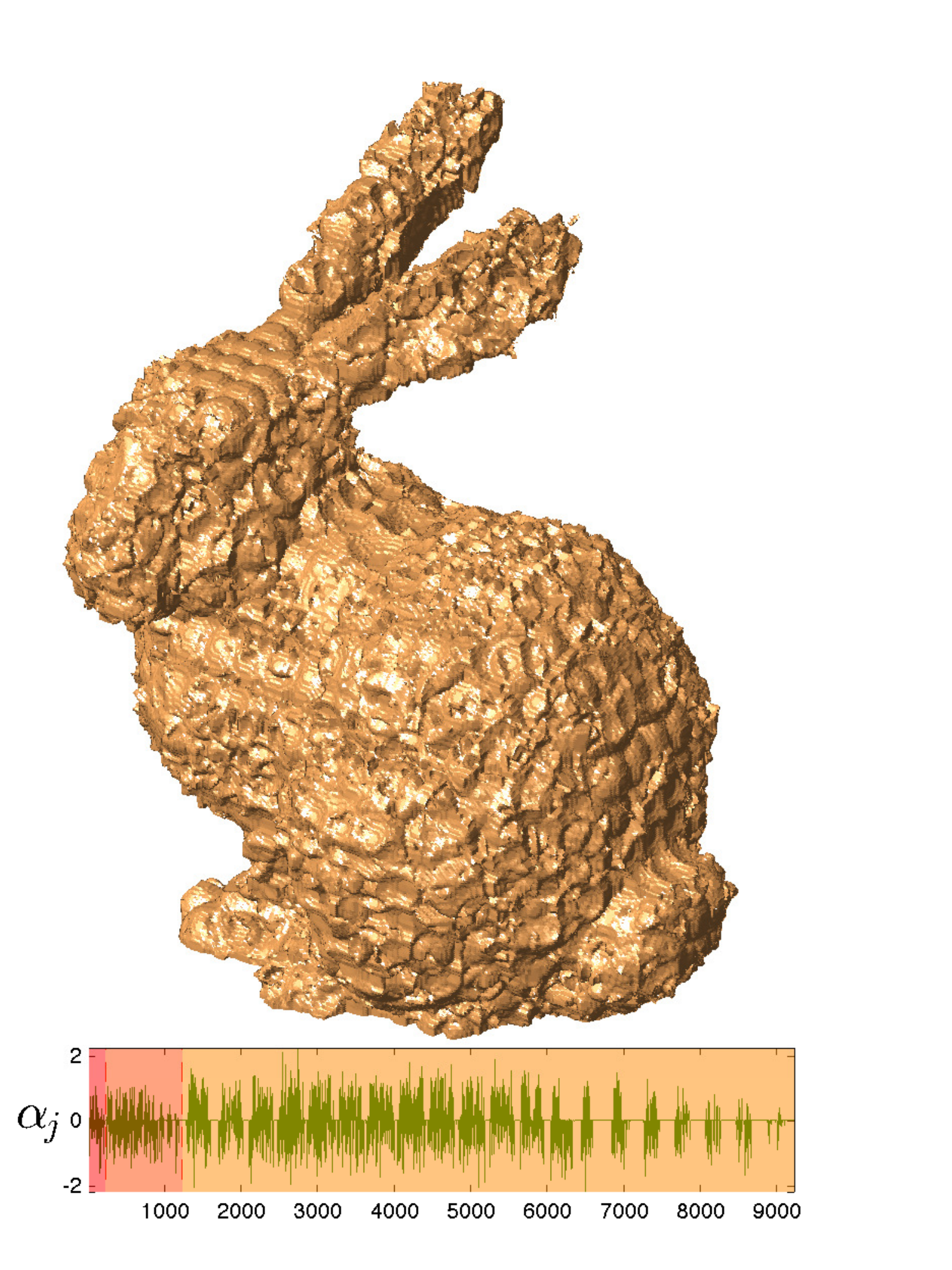}\put (30,-2) {\scriptsize(g)} \end{overpic}
&
\hspace{-1cm}\begin{overpic}[trim={0 -1cm  0 0},clip,width=1.57in]{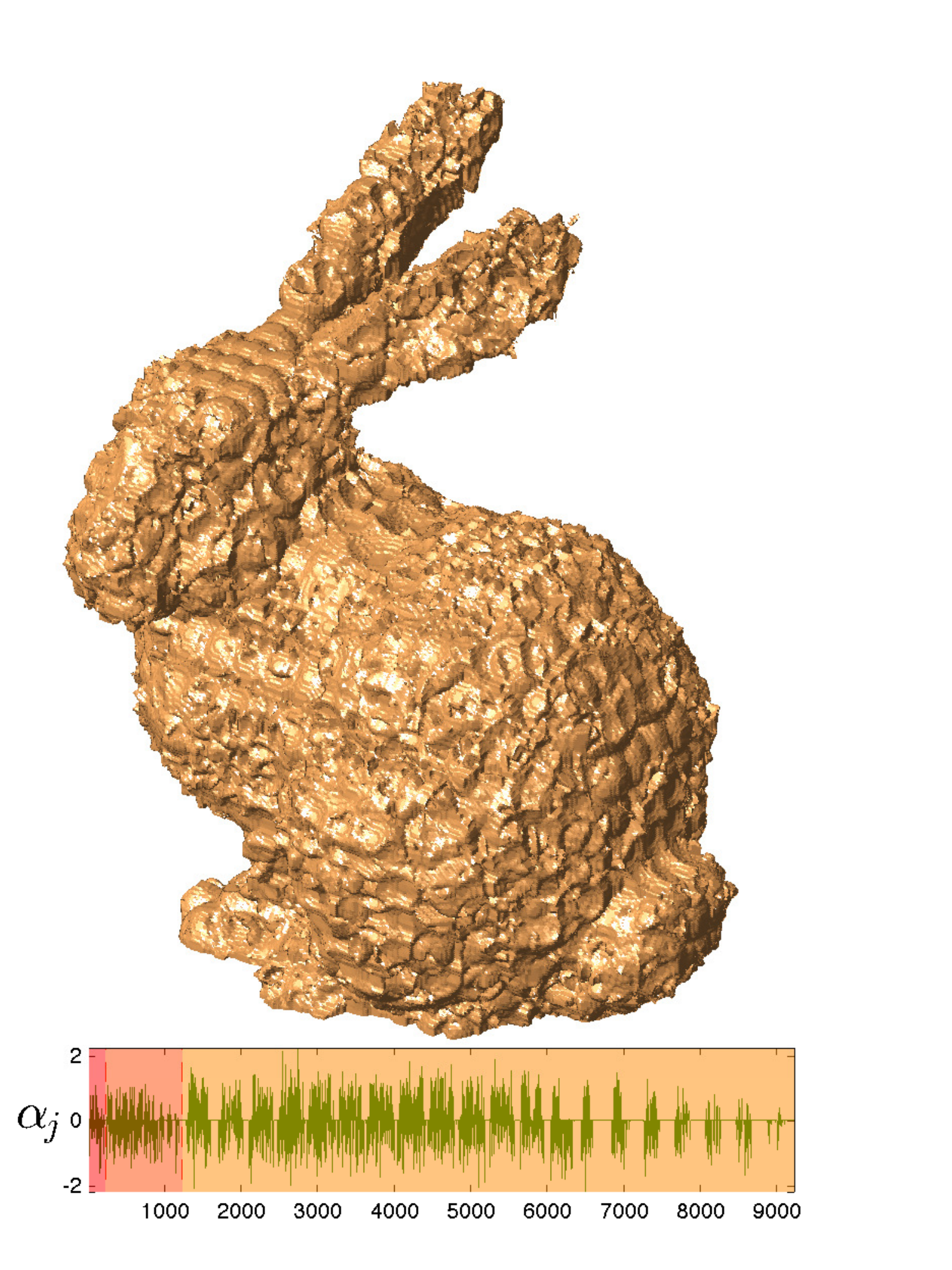}\put (30,-2) {\scriptsize(h)} \end{overpic}
\end{tabular}
\caption{A multi-resolution shape representation of Stanford bunny; (a) three basic elements of the dictionary; (b) reference image; (c-h) reconstructions using the regularized CSC for various values of the regularization parameter. The ribbon below each reconstruction is the recovered $\balpha$, with the indices of the three scales shown by three different colors; (c) $\lambda=10^4$; (d)$\lambda=10^3$; (e) $\lambda=10^2$; (f) $\lambda=10$; (g) $\lambda=10^{-1}$; (h) $\lambda=10^{-3}$}\label{fig3}
\end{figure}

The numerical and theoretical results in this paper can be interpreted as a kind of semantic image segmentation.  Given an image, we separate its domain $D$ into disjoint regions $\Sigma$ and $D\setminus \Sigma$, where $\Sigma$ is chosen so that ``similar'' pixels lie in the same segments.  The region $\Sigma$ is composed through prototype shapes: given a shape dictionary of size $n_s$, $\mathcal{S}_1, \mathcal{S}_2, \cdots \mathcal{S}_{n_s}$, we create $\Sigma$ through a combination of two basic set operations, union and set difference:
\begin{equation}
	\label{eq0}\Sigma = 
	\mathpzc{R}_{\;\Ip,\In}	
	\triangleq\big(\bigcup_{j\in \Ip}\mathcal{S}_j\big) \big\backslash \big(\bigcup_{j\in \In}\mathcal{S}_j\big).
\end{equation}
Using this simple composition rule, rich structure can be developed from a small number of prototypes --- the cartoon example in Figure~\ref{fig_intro} illustrates this.  
Overlapping shapes are formed through the set union, while occlusions come from the set difference.
To account for different rotations an displacements, the dictionary can be populated with instances of the basic elements at different poses. 
By varying the number of shapes used and their sizes, we can obtain a multiscale geometric decomposition of an object.  This is illustrated In Figure~\ref{fig3}, where a reference object (the Stanford bunny) is presented as a composition of spheres of different radii.  In addition, the decomposition can be used to identify the principal shape components, which may carry qualitative information about the content of the image (see, for example, the character recognition experiment in Figure~\ref{fig4}).  This is especially true when we restrict the cardinality of the decomposition by requiring $|\Ip|+|\In|\leq s$ for $s\in\mathbb{N}$.

These decompositions are found by solving a linear program (see Section~\ref{sec:lp} below) that executes in a fraction of a second for the two dimensional examples in this paper, and in a few minutes for the 3D example just described.  More details on these experiments are given in Section~\ref{sec:Sim}.

\subsection{Previous work and contributions}

The problem of finding an optimal shape decomposition of the form \eqref{eq0} was first formally posed in \cite{aghasi2015convex}.  It was shown there that this hard combinatorial problem has a natural convex proxy, and that the solutions to the combinatorial problem and the proxy agree under certain restrictive conditions.  As discussed in more detail in Section~\ref{secPrem} below, the composition of shapes was modeled by superimposing their corresponding characteristic functions.  The key analytical concept was the \emph{disjoint shape decomposition} that maps a dictionary of overlapping shapes into a collection of non-overlapping shapelets.  The minimizers of the convex proxy were characterized by establishing a bijective map between the two representations, and this characterization was used to derive sufficient conditions for the unique optimality of the proposed program; for an extreme case (referred to as the \emph{lucid object condition}), conditions were derived on the overlapping structure of the dictionary elements under which a target composition is recovered by the convex program.  Methods for computing the solution to the convex program were left undiscussed, and experiments were performed using standard convex programming packages \cite{cvx}.
%

From a theoretical standpoint, this paper addresses the main problems which remained open in the previous work. The unique optimality result proposed in \cite{aghasi2015convex} (Proposition 4.6) requires verifying two different conditions: unique intersection of the cost sublevel set with a separating hyperplane, and a tangent cone property to avoid degenerate cases. In this paper, by taking a different analysis path, we propose in Theorem \ref{thunique} a unified set of conditions for the unique optimality. The new result is less restrictive in establishing the optimality conditions. Using this as a tool, we derive sufficient conditions under which the convex program recovers a target composition $\mathpzc{R}_{\;\Ip,\In}$ in a \emph{general} setup. The conditions are stated in terms of the lucidity of the object in the image and the overlapping structure of the elements in the dictionary (geometric coherence). We also present some results on how well the convex solution approximates the original combinatorial solution.

From a computational standpoint, this paper provides two different and very practical computational schemes to address the convex program. We propose an algorithm based on the alternating direction method of multipliers (ADMM) \cite{boyd2011distributed}, which supports distributed processing. This framework suits large-scale problems, where less accuracy is required. We also provide an alternative reformulation of the problem as a linear program, which can be solved accurately in short time. Some of the challenging examples which took more than 8 minutes in \cite{aghasi2015convex}, can be addressed in fractions of a second using the formulation in this paper. 

From an application standpoint, the combination of this paper and \cite{aghasi2015convex} provides a new object learning scheme, applicable and extendable to a wide range of imaging and vision problems. For example, for the first time (to the best of our knowledge) problems such as multi-resolution shape representation can be addressed in a convex framework. Moreover, learning objects based on the constituting geometric elements is a more natural way of approaching object characterization problems than introducing features and regressors which do not provide meaningful interpretations. This work can be considered as an analogous of sparse recovery in harmonic analysis for vision applications. We continue the introduction section by briefly overviewing the formal problem formulation and the proposed convex relaxation in \cite{aghasi2015convex}.

\subsection{Background on the Convex Composition Model}
In the context of binary image segmentation, a variational technique to partition an image $D\subset\mathbb{R}^d$ into $\Sigma$ and $D\setminus\Sigma$ is finding the optimal partitioner via the minimization
\begin{equation}\label{eq1}
\Sigma^*=\operatorname*{arg\,min}_{\Sigma}\;\;\gamma(\Sigma)+\int_{\Sigma}\Pi_{in}(x)\;\mbox{d}x + \int_{D\setminus \Sigma}\Pi_{ex}(x)\;\mbox{d}x,
\end{equation}
where $\gamma(\Sigma)$ is a regularization term promoting a desired structure, and $\Pi_{in}(.)\geq 0$ and $\Pi_{ex}(.) \geq 0$ are some image-dependent inhomogeneity measures. For an image with pixel values $u(x)$, a widely-used measure is the one proposed by Chan and Vese \cite{chan2001active}, which takes $\Pi_{in}(x)=(u(x)-\tilde u_{in})^2$ and $\Pi_{ex}(x)=(u(x)-\tilde u_{ex})^2$, for scalar values $\tilde u_{in}$ and $\tilde u_{ex}$. Other choices of inhomogeneity measures allow us to perform the segmentation based on features such as texture and color \cite{cremers2007review}. We keep the choice of the inhomogeneity measures general as presented in (\ref{eq1}).

It often happens that $\Pi_{in}$ and $\Pi_{ex}$ are fixed or can be estimated a priori. For instance, in the case of classic Chan-Vese model, applicable to grayscale images with almost bimodal histograms, the quantities $\tilde u_{in}$ and $\tilde u_{ex}$ can be roughly estimated from the histogram \cite{ncom}. Defining the functional 
\begin{equation*}
E(\Sigma) \triangleq \int_{\Sigma}\Delta(x)\;\mbox{d}x,\quad\mbox{where}\;\;\;\Delta(x) \triangleq \Pi_{in}(x) - \Pi_{ex}(x),
\end{equation*}
an equivalent formulation of (\ref{eq1}) becomes
\begin{equation}\label{eq2}
\Sigma^*=\operatorname*{arg\,min}_{\Sigma}\;\;\gamma(\Sigma)+E(\Sigma).
\end{equation}

Given a dictionary of shapes $\mathfrak{D}=\{\mathcal{S}_1, \mathcal{S}_2, \cdots ,\mathcal{S}_{n_s}\}$, the problem of interest corresponds to addressing (\ref{eq2}), when $\Sigma$ is restricted to a non-redundant form as (\ref{eq0}). Technically, the goal would be to address 
\begin{align}
\label{eq5}
\min_{\Ip, \In} \;\;E(\mathpzc{R}_{\;\Ip,\In})\quad \quad\mbox{s.t.}\quad \quad |\Ip|+|\In|\leq s,
\end{align}
where the focus is exploring compositions of the form (\ref{eq0}) with limited number of elements (bounded by $s$). Such cardinality restriction promotes simpler representations and prevents shape redundancy. Unfortunately, formally addressing (\ref{eq5}) requires an exhaustive search among a large number of possibilities, growing exponentially with $s$ (specifically $\Omega\big((2n_s/s)^s\big)$ possibilities \cite{aghasi2015conf}).

It is discussed in \cite{aghasi2015convex} that basic set operations among arbitrary shapes can be modeled by superimposing the corresponding characteristic functions. Based on this fact, the authors in \cite{aghasi2015convex} propose the following convex model, namely the convex cardinal shape composition (CSC), as a proxy to (\ref{eq5}):
\begin{align}\nonumber 
&\underset{\balpha}{\min} \;\; G(\balpha)\triangleq \int_{D}\max\Big(\Delta(x) \mathcal{L}_{\balpha}(x), \Delta(x)^-   \Big)\;\mbox{d}x\\[-.15cm]  & s.t.\qquad  \qquad \qquad \qquad   \|\balpha\|_1\leq \tau.\label{eq5x}
\end{align}
In this formulation $\mathcal{L}_{\balpha}(x) \triangleq \sum_{j=1}^{n_s} \alpha_j \chi_{\mathcal{S}_j}(x)$, and $\Delta^-$ returns $\Delta$ when negative and zero otherwise. Moreover, $\chi_\mathcal{S}(x)$ represents the characteristic function of a given shape $\mathcal{S}$, i.e., $\chi_\mathcal{S}(x) = 1$ for $x\in \mathcal{S}$ and $\chi_\mathcal{S}(x) = 0$ for $x\notin \mathcal{S}$. Equivalently, a regularized version of the convex proxy is
\begin{equation}\label{eq5xr}
\min_{\balpha}\;\;G(\balpha) + \lambda\|\balpha\|_1. 
\end{equation}

In this convex model the shape interactions are modeled by the combination of characteristic functions. The number of active shapes are controlled by the $\ell_1$ restriction on $\balpha$. Among the active components of the solution, the components with positive values identify the $\Ip$ set and the negative values represent $\In$. In (\ref{eq5x}), $\tau$ is a free parameter and often an integer quantity. In fact in many interesting scenarios this value can simply be taken to be the same as $s$. That is why the constrained form (\ref{eq5x}) is more desirable and convenient to work with, than the regularized form (\ref{eq5xr}). The regularized form is applies to problems where simple shape descriptors are sought, whereas (\ref{eq5x}) makes a more direct connection with (\ref{eq5}) in terms of controlling the number of active elements. The interested reader is referred to \cite{aghasi2015convex} for a more detailed discussion.

The remainder of the paper is structured as follows. In Section \ref{optsec}, we present two numerical schemes to address the CSC. Section \ref{secPrem} overviews some preliminary notions from \cite{aghasi2015convex}, which are essential tools for the analysis of CSC. Section \ref{secUniq} presents the unique optimality result, which compared to Proposition 4.6 and Theorem 4.7 in \cite{aghasi2015convex} presents a unified and less restrictive optimality framework. Finally in Section \ref{sec:accrecovery} we derive sufficient conditions under which CSC recovers a target composition and relate the convex solution to the solution of (\ref{eq5}). In Section \ref{sec:Sim} we report some numerical experiments and provide the concluding remarks. The theoretical content is presented in Sections \ref{secPrem}, \ref{secUniq} and \ref{sec:accrecovery}. Sections \ref{sec:introduction}, \ref{optsec} and \ref{sec:Sim} can be studied independently and would provide the general formulation required for implementation purposes.

\subsection{Notation}
Our presentation mainly relies on multidimensional calculus. We use bold characters to denote vectors and matrices. Considering a matrix $\boldsymbol{A}$ and the index sets $\Gamma_1$, and $\Gamma_2$, we use $\boldsymbol{A}_{\Gamma_1,:}$ to denote the matrix obtained by restricting the rows of $\boldsymbol{A}$ to $\Gamma_1$. Similarly, $\boldsymbol{A}_{:,\Gamma_2}$ denotes the restriction of $\boldsymbol{A}$ to the columns specified by $\Gamma_2$, and $\boldsymbol{A}_{\Gamma_1,\Gamma_2}$ is the submatrix with the rows and columns restricted to $\Gamma_1$ and $\Gamma_2$, respectively.

For a vector $\boldsymbol{a}=[a_1,\cdots,a_n]^T$, we use $\boldsymbol{a}_i$ to denote the $i$-th element of $\boldsymbol{a}$, i.e., $\boldsymbol{a}_i=a_i$. In indexing vectors of the same type, such as $\boldsymbol{a}$, we use the notation $\boldsymbol{a}_{\langle i \rangle}$ to avoid confusion with the $i$-th entry of $\boldsymbol{a}$. Such indexing would be only needed in Section \ref{optsec}. The support of a vector $\boldsymbol{a}$ is denoted by $\supp(\boldsymbol{a})$. Following the MATLAB convention, the vertical concatenation of two vectors $\boldsymbol{a}$ and $\boldsymbol{b}$ (i.e., $[\boldsymbol{a}^T,\boldsymbol{b}^T]^T$) is frequently denoted by $[\boldsymbol{a};\boldsymbol{b}]$. 

Finally, in the analysis of the sets, the cardinality of a set $A$ is denoted by $|A|$. For a given set $\mathcal{S}\in\mathbb{R}^d$, $\mathcal{S}^o$ and $\overline{\mathcal{S}}$ respectively denote the interior and the closure of the set, and $\mathcal{S}^c$ denotes the set complement.

\section{Convex Programming for CSC}\label{optsec}

We propose two different methods to solve the proposed convex problem. The first method is based on increasing the number of optimization variables to reformulate the problem as a linear program (LP), which suits both formulations (\ref{eq5x}) and (\ref{eq5xr}). Once the problem is cast as a standard LP, variety of fast computational schemes may be considered \cite{mittelmann2002benchmarks}. The second approach uses an alternating direction method of multipliers (ADMM) \cite{boyd2011distributed}, and best suits a parallel computing framework for large-scale problems.  This approach is a better fit to the regularized form of CSC, where the focus is mainly obtaining simple compositions and not necessarily accurate solutions.

\subsection{CSC Reformulation as an LP}\label{sec:lp}
In a real-world application, the imaging domain can be discretized into a collection of pixels, here denoted by $\{x_i\}_{i=1}^N$. A similar quantization applies to the elements of the dictionary to produce a discrete representation of (\ref{eq5x}) as
\begin{equation}\label{eqimp1}
\min_{\balpha} \;\;\;\tilde{G}(\balpha)\triangleq \sum_{i=1}^N\max(\boldsymbol{a}_{\langle i\rangle}^T\balpha,b_i)\quad s.t.\quad \|\balpha\|_1\leq \tau.
\end{equation}
Here, for $j=1,\cdots,n_s$, the $j$-th entry of the vector $\boldsymbol{a}_{\langle i\rangle}$ is $\Delta(x_i)\chi_{\mathcal{S}_j}(x_i)$, and $b_i=\Delta(x_i)^-$.

To reformulate (\ref{eqimp1}) as an LP, consider a variable $\Bz\in\mathbb{R}^N$ with the entries 
\begin{equation}\label{eqimp2}
z_i=\max(\boldsymbol{a}_{\langle i\rangle}^T\balpha,b_i)-b_i.
\end{equation}
Equation (\ref{eqimp2}) naturally imposes the constraints $\boldsymbol{a}_{\langle i\rangle}^T\balpha\leq z_i+b_i$, and $b_i\leq z_i+b_i$ (or simply $\Bz\succeq \boldsymbol{0}$). We also introduce two positive variables $\Bz' = \balpha^+$ and $\Bz'' = -\balpha^-$, which essentially yield $\balpha = \Bz'-\Bz''$ and $\|\balpha\|_1 = \boldsymbol{1}^T(\Bz'+\Bz'')$. By stacking up the vectors $\boldsymbol{a}_{\langle i\rangle}^T$ into a matrix $\boldsymbol{A}\in\mathbb{R}^{N\times n_s}$ (that is, $\boldsymbol{A}_{i,:} = \boldsymbol{a}_{\langle i\rangle}^T$), an equivalent representation of (\ref{eqimp1}) in terms of $\Bz$, $\Bz'$ and $\Bz''$ becomes
\begin{equation}\label{eqimp3}
\begin{array}{lc} \underset{{\Bz,\Bz',\Bz''}}{\min}&\boldsymbol{1}^T\Bz\\ s.t. & \begin{bmatrix}-\boldsymbol{I} & \boldsymbol{A}& -\boldsymbol{A}\\ \boldsymbol{0}&\boldsymbol{1}^T&\boldsymbol{1}^T\end{bmatrix}\begin{bmatrix}\Bz\\ \Bz'\\ \Bz''\end{bmatrix} \preceq \begin{bmatrix}\boldsymbol{b} \\ \tau \end{bmatrix}, \quad \hspace{-.1cm} \begin{bmatrix}\Bz \\ \Bz' \\ \Bz''\end{bmatrix}\succeq \boldsymbol{0},
\end{array}
\end{equation}
where $\boldsymbol{b}\in\mathbb{R}^N$ is a vector with entries $b_i$.

It is worth noting that the constraint matrix in (\ref{eqimp3}) is sparse, and the level of sparsity varies by the average number of pixels used to represent the dictionary elements. Moreover, since the number of variables in the dual LP is less than the variable count in the primal ($N+1$ vs. $N+2n_s$), depending on the LP algorithm, it might be computationally faster to address the dual problem:
\begin{equation*}
\begin{array}{lc} \underset{\By}{\min}&\begin{bmatrix}\boldsymbol{b}; \tau \end{bmatrix}^T\!\!\By\\ s.t. & -\begin{bmatrix}-\boldsymbol{I} & \boldsymbol{0}\\ \boldsymbol{A}^T & \boldsymbol{1}\\ -\boldsymbol{A}^T & \boldsymbol{1}\end{bmatrix}\By \preceq \begin{bmatrix}\boldsymbol{1}\\\boldsymbol{0}\\ \boldsymbol{0} \end{bmatrix},\quad  \By\succeq \boldsymbol{0}.
\end{array}
\end{equation*}
By the complementary slackness \cite{bertsimas1997introduction}, the dual solution can identify the underlying active constraints and the nonzero components of the primal solution. Accordingly, the primal minimizer may be retrieved by addressing a linear system of equations.

For the regularized CSC in (\ref{eq5xr}), a similar path makes the LP conversion straightforward (the details of which are skipped here).

\subsection{Distributed Optimization for CSC}
A main component of the ADMM is an efficient method for applying the appropriate proximal operators.  Given a convex function $g(\balpha):\mathbb{R}^{n_s}\to\mathbb{R}$, the proximal operator of $g$ scaled by a factor $\xi$ is defined by
\begin{equation}\label{eqimp4}
\mbox{\textbf{prox}}_{\xi g}(\boldsymbol{\rho})\triangleq\operatorname*{arg\,min}_{\balpha} \;\;g(\balpha)+\frac{1}{2\xi}\|\balpha-\boldsymbol{\rho}\|^2.
\end{equation}

For the quantized CSC in (\ref{eqimp1}), the objective may be cast as
\begin{equation}\label{eqimp7}
\min_{\balpha} \;\;\; \sum_{i=1}^N g_i(\balpha) \;+\;f(\balpha),
\end{equation}
where $g_i(\balpha)=\max(\boldsymbol{a}_{\langle i\rangle}^T\balpha,b_i)$ and $f(\balpha) = \mathbbit{I}_{\hspace{.02cm}\mathcal{C}_{\tau}\!}(\balpha)$ represents the indicator function of the convex set $\mathcal{C}_{\tau}=\{\balpha\in\mathbb{R}^{n\!_s}:\;\|\balpha\|_1\leq {\tau}\}$. That is,
\begin{equation}\label{eqimp8}
\mathbbit{I}_{\hspace{.02cm}\mathcal{C}_{\tau}\!}(\balpha) = \left \{ \begin{array}{lc} 0&\|\balpha\|_1\leq {\tau}\\+\infty & \|\balpha\|_1> {\tau} \end{array} \right..
\end{equation}
Problem (\ref{eqimp7}) can be classified as an instance of the \emph{global variable consensus problem with regularization} (see \S7.1.1 of \cite{boyd2011distributed}), which is equivalent to the convex program 
\begin{equation}\label{eqimp9}
\min_{\balpha_{\langle 1 \rangle},\cdots \balpha_{\langle N \rangle},\boldsymbol{\rho}}  \sum_{i=1}^N g_i(\!\balpha_{\langle i \rangle}\!) + f(\boldsymbol{\rho}),\;\; s.t. \;\; \balpha_{\langle 1 \rangle}\!=\!\cdots\!=\!\balpha_{\langle N \rangle}\!=\!\boldsymbol{\rho}.
\end{equation}
As elaborated in \cite{boyd2011distributed}, an ADMM iterative process to solve (\ref{eqimp9}) takes the following alternating form:  
\begin{eqnarray}\label{eqimp10}
\balpha^{k+1}_{\langle i\rangle }&=& \mbox{\textbf{prox}}_{\xi g_i}(\boldsymbol{\rho}^ k-\boldsymbol{\omega}^k_{\langle i \rangle}),\\ \label{eqimp11}
\boldsymbol{\rho}^{ k+1}&=& \mbox{\textbf{prox}}_{(\xi/N) f}(\bar{\balpha}^{ k+1}+\bar{\boldsymbol{\omega}}^{ k}),\\ 
\label{eqimp12}
\boldsymbol{\omega}^{k+1}_{\langle i\rangle}&=& \boldsymbol{\omega}^k_{\langle i\rangle}+\balpha^{k+1}_{\langle i\rangle}-\boldsymbol{\rho}^{k+1}.
\end{eqnarray}
Here, the superscript $k$ denotes the iteration index, the variable $\boldsymbol{\omega}_{\langle i\rangle}$ measures the deviation of $\balpha_{\langle i\rangle }$ and $\boldsymbol{\rho}$, and the variables with the bar denote the mean vectors, i.e., $\bar{\boldsymbol{\omega}}^{ k}=\frac{1}{N}\sum_{i=1}^N \boldsymbol{\omega}^k_{\langle i\rangle}$ and $\bar{\balpha}^{ k}=\frac{1}{N}\sum_{i=1}^N \balpha^k_{\langle i\rangle}$. While a warm start can always help speeding up the convergence, zero initialization for the underlying variable would be a straightforward choice. 

The proposed iterative scheme can be efficiently carried out in a distributed computing framework. Each subsystem is in charge of computing the proximal update on $\balpha_{\langle i \rangle}$ and an evaluation of $\boldsymbol{\omega}_{\langle i\rangle}$, as suggested by (\ref{eqimp10}) and (\ref{eqimp12}). The outcomes are averaged and passed to a single computing unit to perform the proximal update on $\boldsymbol{\rho}$.

Our main motivation to use the proposed ADMM scheme is the possibility of deriving a closed form expression for $\mbox{\textbf{prox}}_{\xi g_i}(.)$. Specifically, for $g(\balpha) = \max(\Ba^T\balpha,b)$, following the derivation in Appendix \ref{AppendProx}, we have
\begin{equation}\label{prox_derived}
\mbox{\textbf{prox}}_{\xi g}({\boldsymbol{\rho}})  = \left \{\begin{array}{ll} {\boldsymbol{\rho}}-\xi\Ba &  \Ba^T{\boldsymbol{\rho}} \in( b + \xi\|\Ba\|^2\;,\;\infty)\\ {\boldsymbol{\rho}}-\frac{\Ba\Ba^T}{\|\Ba\|^2}{\boldsymbol{\rho}}+\frac{b}{\|\Ba\|^2} \Ba & \Ba^T{\boldsymbol{\rho}}\in[b\;,\; b + \xi\|\Ba\|^2]\\ {\boldsymbol{\rho}} & \Ba^T{\boldsymbol{\rho}}\in(-\infty\;,\;b)
 \end{array} \right. \!\!.
\end{equation}
In other words, the proximal updates on the vectors $\balpha_{\langle i \rangle}$ can be computed very cheaply, and the proposed scheme can be conveniently applied to large-scale problems. 

We would also like to note that for $f(\balpha) = \mathbbit{I}_{\hspace{.02cm}\mathcal{C}_{\tau}\!}(\balpha)$, the proximity function in (\ref{eqimp11}) reduces to $\mbox{\textbf{prox}}_{(\xi/N) f}(\boldsymbol{\rho}) = \mathcal{P}_{\mathcal{C}_{\tau}}(\boldsymbol{\rho})$, the projection onto $\mathcal{C}_{\tau}$. In the regularized form of CSC where $f(\balpha) = \lambda\|\balpha\|_1$, the proximity function reduces to the soft thresholding operator (see \cite{aghasi2015conf} for more details).

Generally speaking, the ADMM requires a large number of iterations to recover accurate solution of a convex problem \cite{boyd2011distributed, aghasi2015conf}, and therefore more suitable for the regularized version of CSC, where the focus is mainly concision of the representation rather than accuracy. For the purpose of experiments in this paper we employ the LP formulation which is addressable on a desktop computer.

\section{Preliminary Notions for Analysis}\label{secPrem}

To present the main results, we need to overview some basic definitions and two major concepts.  The first concept is that of a disjoint shape decomposition, discussed in Section~\ref{sec:dsd} below.  This gives an alternative way to analyze shape compositions using non-overlapping ``shapelets''.  The second concept is a stable procedure that relates a shape composition to an $\balpha$-representation. This process plays a key role in developing the main results by relating the conditions for the recovery of a composition to the overlapping pattern of its elements.  A more detailed presentation of the material in this section is available in \cite{aghasi2015convex}.

In the context of the present theoretical framework, a {\bf shape} in $\mathbb{R}^d$ is a closed set (hence, the union of finitely many closed sets) with positive Lebesgue measure. 

Two shapes $\mathcal{S}_1$ and $\mathcal{S}_2$ are called {\bf disjoint} if $\mathcal{S}_1^o \cap \mathcal{S}_2^o = \emptyset$. Therefore, two disjoint shapes can at most share a set with zero Lebesgue measure.

A composition $\mathpzc{R}_{\;\Ip,\In}$ is called {\bf non-redundant} if excluding any shape from the composition changes the Lebesgue measure of the outcome. 
%
%

\subsection{Disjoint Shape Decomposition}
\label{sec:dsd}

The overlapping pattern of the elements in the dictionary can in general be very complex. The process outlined below allows us to decompose overlapping shapes into smaller non-overlapping elements and decode the entire overlap pattern into a binary matrix.  We emphasize that this decomposition, which greatly expands the cardinality of the representation of a shape, is for the purpose of analysis only; it plays no role in our computational framework.

Given $n$ overlapping shapes $\mathcal{S}_1,\mathcal{S}_2,\cdots,\mathcal{S}_n$, consider an $n$-dimensional binary vector $\boldsymbol{b}\in\{0,1\}^n\setminus \{0\}^n$. Corresponding to the constructor vector $\boldsymbol{b}$, a set $\Omega$ with nonempty interior is called a {\bf shapelet} when it is representable as
\begin{equation}  \label{eq10}
\Omega = \bigcap_{j=1}^n\Theta(\mathcal{S}_j),
\;\;\;\;\mbox{where}\;\;\;\;
 \Theta(\mathcal{S}_j) =
\left\{
\begin{array}{lr}
       \mathcal{S}_j& \boldsymbol{b}_j=1\\[.2cm]
       \overline{\mathcal{S}_j^c}& \boldsymbol{b}_j =0
     \end{array}.
   \right.
\end{equation}
Theoretically, we can have up to $2^n-1$ shapelets. However, in assessing all possible $\boldsymbol{b}\in\{0,1\}^n\setminus \{0\}^n$, the response of (\ref{eq10}) to some vectors is simply a null set  and not technically counted as a shapelet. We denote the total number of shapelets by $n_\Omega$, which is certainly upper-bounded by $2^n-1$. 

The disjoint shape decomposition (DSD) is referred to the process which generates the possible $n_\Omega$ shapelets and their corresponding constructor vectors by assessing the response of (\ref{eq10}) to the elements of $\{0,1\}^n\setminus \{0\}^n$. We succinctly denote the process by 
\begin{equation}\label{eq12}
\{\Omega_i,\boldsymbol{B}_{i,:}\}_{i=1}^{n_\Omega} = \mbox{DSD}\Big( \{\mathcal{S}_j\}_{j=1}^{n}\Big),
\end{equation}
where the $i$-th row of the binary matrix $\boldsymbol{B}\in \{0,1\}^{n_\Omega\times n}$ corresponds to the $i$-th shapelet constructor vector. When $\boldsymbol{B}$ is resulted from the DSD process over a specific set of shapes, we call it the corresponding {\bf bearing matrix}, and when its construction involves all the elements of the dictionary,  we refer to it as the {\bf dictionary bearing matrix}.

While the shapes $\mathcal{S}_1,\mathcal{S}_2,\cdots,\mathcal{S}_n$ can overlap arbitrarily or even cover one another, the set of shapelets they generate are disjoint:
\[
	\Omega_{i_1}^o\cap\Omega_{i_2}^o = \emptyset, 
	\quad \mbox{for} \quad i_1\neq i_2,\;\;i_1,i_2\in\{1,2,\cdots,n_\Omega\}. 
\] 
Moreover, using the outcome of the DSD process, each shape $\mathcal{S}_j$ enjoys a unique shapelet representation as 
\begin{equation}\label{eqIj}
\mathcal{S}_j = \bigcup_{i\in \mathcal{I}_j} \Omega_i,\;\; \mbox{where}\;\;\;\mathcal{I}_j \triangleq \Big\{i: \boldsymbol{B}_{i,j}=1, i\in\{1,\cdots,n_\Omega\}\Big\}.
\end{equation}

The separability of the elements in the shapelet domain will later assist us in the analysis of the CSC problem. This separability is automatically transferred to the $\balpha$-domain and $\mathcal{L}_{\balpha}(x)$ will take a constant value within each $\Omega_i$. If the $i$-th element of the vector $\Beta\in\mathbb{R}^{n_\Omega}$ denotes the value of $\mathcal{L}_{\balpha}(x)$ over $\Omega^o_i$, then for an arbitrary $\balpha$
\begin{equation}
\Beta = \boldsymbol{B}\balpha.
\end{equation}

\subsection{Mapping Shape Compositions to Functions} 

The CSC framework relies on recasting shape compositions of the form \eqref{eq0} as the positive support of a linear combination of indicator functions.  Given $\mathpzc{R}_{\;\Ip,\In}$ as \eqref{eq0},  we can find many $\{\alpha_j\}$ such that the function
\[	
	\mathcal{L}_{\balpha}(x) = \sum_{j\in \Ip \cup \In} \alpha_j \chi_{\mathcal{S}_j}(x)
\]
has positive support that matches $\mathpzc{R}_{\;\Ip,\In}$:
\[
	\supp\big(\mathpzc{L}_{\boldsymbol{\alpha}}(x)^+ \big) = \mathpzc{R}_{\;\Ip,\In}.
\]
To choose one of these possibilities, we take $\balpha_{\Ip}=\boldsymbol{1}$, and then $\balpha_{\In}$ to be the solution to the optimization program
\begin{equation*}
	\min_{\balpha_{\In}}\; -\!\!\sum_{j\in \In}\!\alpha_j
	\; \text{s.t.} \; \sum_{j\in \In}\! \alpha_j \chi_{\mathcal{S}_j}(x)\leq -\sum_{j\in \Ip}\! \chi_{\mathcal{S}_j}(x), \;\forall x\in \mathpzc{R}_{\;\Ip,\In}^c.
\end{equation*}
Under certain technical conditions on the bearing matrix for $\mathrm{DSD}\left(\{\mathcal{S}_j\}_{j\in\Ip\cup\In}\right)$, the solution to this program is unique (see \cite[Thm.\ 4.5]{aghasi2015convex}).  In the sequel, we will assume that this is indeed true for the shape composition we are trying to recover, and we call such shape compositions {\bf basic}.  In general, non-redundant compositions with simple overlaps among the elements meet the criteria for being basic.

The optimization program above gives us a bijective map from a basic, non-redundant shape composition $\mathpzc{R}$  to a coefficient sequence $\alpha_{\mathpzc{R}}$; we use $\mathcal{A}(\cdot)$ to denote this map:
\begin{equation*}
	\label{alphar}
	\balpha_\mathpzc{R} \triangleq \mathcal{A}\big(\{\mathcal{S}_j\}_{j\in\Ip\cup\In};\Ip,\In\big).
\end{equation*}
Inverting the map is straightforward.  For a given $\balpha_\mathpzc{R}$, the index sets $\Ip$ and $\In$ correspond to the positive and negative components of $\balpha_\mathpzc{R}$. 

The components of the linkage outcome, $\balpha_\mathpzc{R}$, often take integer values which later allows us to sweep integer quantities for the CSC free parameter $\tau$.  This is demonstrated in the experiment in Figure~\ref{fig2}, for example.

\section{Unique Optimality}\label{secUniq}
The material presented in this section discusses the unique optimality conditions in an abstract level. Basically, we focus on the conditions for a given vector $\balpha^*$ to be the unique minimizer of the CSC program (\ref{eq5x}). The conditions mainly depend on the mean inhomogeneity measures and the value of $\mathcal{L}_{\balpha^*}(x)$ over the shapelets resulted from a DSD on the entire dictionary. 

An important outcome of this section is Theorem \ref{thunique}, which presents such conditions. Later in Section \ref{sec:accrecovery} we show how this result translates into the recovery of a target composition in terms of the lucidity of the underlying object in the image and the standing of the composition elements in the dictionary. To facilitate the presentation we proceed by first discussing the notion of lucidity.

\subsection{Lucid Objects}\label{sec:loc}
The lucid object condition (LOC) corresponds to the condition where the object and the background are completely separable by the inhomogeneity measures.
More specifically, given a region $\Sigma\subset D$, the LOC holds for $\Sigma$ if
\begin{equation}\label{eqloccon}\left\{\begin{array}{lc}
\!\!\Delta(x)<0 & x\in \Sigma\\
\!\!\Delta(x)>0 & x\in D\setminus\Sigma
\end{array},
\right.viz.:\;  
\Pi_{ex}(x)\overset{x \in \Sigma}{\underset{x \in D\setminus \Sigma}{\gtrless}}\Pi_{in}(x). 
\end{equation}
 As an example, for the Chan-Vese model, LOC holds for an object $\Sigma$ when the intensity variations around $\tilde u_{in}$ and $\tilde u_{ex}$ are sufficiently small in the corresponding regions.

The main analysis theme in \cite{aghasi2015convex} is when LOC holds for a region $\Sigma$ in the image and there exists a composition of the dictionary elements which perfectly matches $\Sigma$. Thanks to Theorem \ref{thunique} later presented in this section, we are able to develop stronger results which not only improve upon \cite{aghasi2015convex}, but also accounts for a tolerable level of LOC violation in recovering a target composition.  

\subsection{Convex Analysis and Certificate of Duality}
In this section, we derive sufficient conditions under which a particular vector $\balpha^*$ is the solution to the convex program (\ref{eq5x}).  These conditions are not only milder than those in previous work \cite{aghasi2015convex}, they also allow a unified analysis that trades off object lucidity in the image against the geometrical coherence of the target region.

Consider $\{\Omega_i\}_{i=1}^{n_\Omega}$ and $\boldsymbol{B}$ to be the outcomes of a DSD process on the dictionary elements $\{\mathcal{S}_j\}_{j=1}^{n_s}$. The unique optimality conditions for a given $\balpha$ heavily rely on the entries of $\Beta = \boldsymbol{B}\balpha$ and the mean inhomogeneity quantities: 
\begin{equation}\label{inhom}
p_i \triangleq \int_{\Omega_i}  \Delta(x)^+ \; \mbox{d}x,\quad \mbox{and} \quad q_i \triangleq \int_{\Omega_i}  \big(-\Delta(x)\big)^+ \; \mbox{d}x.
\end{equation}
We proceed by introducing the main notations to more conveniently present the result . Given a set of points $A\subset \mathbb{R}$, for a vector $\boldsymbol{v}\in\mathbb{R}^n$ we will extensively make use of the following notation:
\[I_{A}(\boldsymbol{v}) \triangleq \Big\{i:\boldsymbol{v}_i\in A, \; i\in\{1,2,\cdots,n\}\Big\}.
\]
Basically, $I_{A}(\boldsymbol{v})$ returns the indices of the components of $\boldsymbol{v}$, which belong to the set $A$. In a singleton case, for $a\in\mathbb{R}$, we use the notation $I_{a}(\boldsymbol{v}) \triangleq \big\{i:\boldsymbol{v}_i=a, \; i\in\{1,2,\cdots,n\}\big\}$. For instance, for a sparse vector $\boldsymbol{v}$, $I_{\mathbb{R}\setminus\{0\}}(\boldsymbol{v})=\supp(\boldsymbol{v})$.

As will be detailed in the proof of Theorem \ref{thunique}, a key component of the analysis is a separable reformulation of $G(\balpha)$ by the variable change $\Beta = \boldsymbol{B}\balpha$. Similar to the majority of the literature in convex analysis (e.g., see \cite{candes2013simple}) we need to consider support sets
\begin{align}\label{ee-7}
\Gamma_{0^-} \triangleq I_{(-\infty,0)}(\Beta),\quad \mbox{and}\quad 
\Gamma_{1^+} \triangleq I_{(1,\infty)}(\Beta), 
\end{align}
and the off-support sets
\begin{align}\label{ee-8}
\Gamma_0 \triangleq I_{0}(\Beta),\quad \mbox{and}\quad 
\Gamma_1 \triangleq I_{1}(\Beta).
\end{align}
We also introduce the bounding vectors $\boldsymbol{l}$ and $\boldsymbol{u} \in \mathbb{R}^{|\Gamma_0\cup\Gamma_1|}$ with entries
\begin{equation}\label{ee-9}
\boldsymbol{l}_{m(\ell)} =  \left\{\begin{array}{ll}
\!\!q_\ell-p_\ell & \ell\in \Gamma_0\\
\!\!-p_\ell &\ell \in \Gamma_1
\end{array}
\right. , \boldsymbol{u}_{m(\ell)} =  \left\{\begin{array}{ll}
\!\!q_\ell & \ell\in \Gamma_0\\
\!\!q_\ell - p_\ell &\ell \in \Gamma_1
\end{array}
\right. ,
\end{equation}
where $m(.): \Gamma_0\cup\Gamma_1 \mapsto \{1,2,\cdots ,|\Gamma_0\cup\Gamma_1|\}$ is a simple bijective index map that allows filling in the bounding vector entries in a desired order.

\begin{theorem}[Unique Optimality]\label{thunique}
\emph{Given the dictionary bearing matrix $\boldsymbol{B}\in \{0,1\}^{n_\Omega\times n_s}$, consider a target vector $\balpha^*\in\mathbb{R}^{n_s}$ such that $\|\balpha^*\|_1=\tau$ and correspondingly for $\Beta^* = \boldsymbol{B}\balpha^*$, all entries of $\Beta^*$ lie outside the interval $(0,1)$. Further, consider $\boldsymbol{e}\in\mathbb{R}^{n_s}$ such that 
\begin{equation}\label{ee-10}
\boldsymbol{e}_j \triangleq\!\! \sum_{\ell\in\mathcal{I}_j \cap \Gamma_{1^+}}\!\!\!\!p_\ell -\!\!\sum_{\ell\in\mathcal{I}_j \cap \Gamma_{0^-}}\!\!\! q_\ell, \quad j=1,2,\cdots,n_s,
\end{equation}
and $\boldsymbol{c}\in\mathbb{R}^{n_s}$ such that $\boldsymbol{c}_j = \sign(\balpha_j^*)$ for $j\in \Gamma_{\!\balpha^*} \triangleq \supp(\balpha^*)$ and $|\boldsymbol{c}_j|< 1$ for $j\in \Gamma_{\balpha^*}^c$. If the matrix $\boldsymbol{B}_{\Gamma_0\cup\Gamma_1,\Gamma_{\!\balpha^*}}$ has full column rank and there exist $\boldsymbol{\eta}\in\mathbb{R}^{|\Gamma_0\cup\Gamma_1|}$ and a scalar $\eta_c>0$ such that\footnote{The rows of $\boldsymbol{B}_{\Gamma_0\cup\Gamma_1,:}$ also need to be arranged according to the index map $m(.)$ used in (\ref{ee-9}). More specifically $\boldsymbol{B}_{\Gamma_0\cup\Gamma_1,:} = \boldsymbol{B}_{\{m^{-1}(\ell): \ell\in 1,2,\cdots,|\Gamma_0\cup\Gamma_1|\},:}$.}
\begin{align}\label{eqthunique1}
\boldsymbol{B}_{\Gamma_0\cup\Gamma_1,:}^T\boldsymbol{\eta} =  \eta_c\boldsymbol{c} + \boldsymbol{e},
\quad\mbox{and}\quad 
\boldsymbol{l}\prec\boldsymbol{\eta}\prec \boldsymbol{u},
\end{align}
then $\balpha^*$ is the unique minimizer of the convex program (\ref{eq5x}).}
\end{theorem}
This result expresses the general unique optimality conditions in an entirely abstract way, by assessing the properties of the dictionary bearing matrix over the predefined support and off-support sets. The discussion in the next section translates this result into the recovery of a target composition in terms of the lucidity of the underlying object in the image and the standing of the composition elements in the dictionary.

\section{Recovery and Tolerable LOC Violation}
\label{sec:accrecovery}
In this section, we derive sufficient conditions under which CSC successfully identifies the elements of a target composition $\mathpzc{R}_{\Ip,\In}$.  We start by fixing some conventions for notation.

Consider a dictionary of shape elements $\{\mathcal{S}_j\}_{j=1}^{n_s}$ and an image domain $D$. The target composition is a basic non-redundant composition $\mathpzc{R}_{\Ip,\In}$, such that $\Ip,\In\subset\{1,\cdots,n_s\}$. To avoid indexing complications, we simply assume that $\Ip = 1,\cdots, n_\oplus$ and $\In = n_\oplus+1, \cdots,n_\oplus+n_\ominus$. We will refer to the remaining off-target dictionary elements, index by $\{n_\oplus+n_\ominus+1,\cdots,n_s\}$, as the {\bf exterior} shapes. 

With reference to the basic composition, the outcome of the linkage process is denoted by
\begin{equation}\label{eqlink}
\balpha_\mathpzc{R} = \mathcal{A}\big(\{\mathcal{S}_j\}_{j\in\Ip\cup\In};\Ip,\In\big).
\end{equation}
For the parameter selection $\tau = \big\|\mathcal{A}\big(\{\mathcal{S}_j\}_{j\in\Ip\cup\In};\Ip,\In\big)\big\|_1$, we aim to discuss conditions that the CSC outcome 
\begin{equation}\label{eqcon2}
\balpha^*=\operatorname*{arg\,min}_{\balpha} \;\;G(\balpha) \quad s.t. \quad \|\balpha\|_1\leq \tau
\end{equation}
satisfies 
\begin{equation}\label{eqalph}
\balpha^*_{\Ip\cup\In}=\balpha_\mathpzc{R}\qquad \mbox{and}\qquad  \balpha^*_{(\Ip\cup\In)^c}=\boldsymbol{0},
\end{equation}
which declares a successful identification of the target composition elements. 

In the ideal scenario that $\overline{\mathpzc{R}_{\Ip,\In}}=\Sigma\subset D$ and the LOC holds for $\Sigma$,  the analysis in \cite{aghasi2015convex} affirms that a successful identification is guaranteed, as long as the exterior shapes maintain a restricted level of overlap (called the geometric coherence) with the elements of $\mathpzc{R}_{\Ip,\In}$.
There assumptions are too strong to be useful in general scenarios.  For instance, they do not hold when $\Sigma$ and $\mathpzc{R}_{\Ip,\In}$ are misaligned and/or the clutter and noise in the image causes (\ref{eqloccon}) to be violated in measurable portions of the image. We will assert that when the violation of the ideal scenario is ``sufficiently small'', under a limited level of geometric coherence between the exterior shapes and $\mathpzc{R}_{\Ip,\In}$, a successful identification of $\Ip$ and $\In$ is still possible.

To more technically present the result, consider $\{\Omega^\mathpzc{R}_\ell\}_{\ell=1}^{n_{\Omega^\mathpzc{R}}}$ and $\boldsymbol{B}^\mathpzc{R} \in \{0,1\}^{n_{\Omega^\mathpzc{R}}\times (n_\oplus+n_\ominus)}$ to be the shapelets and the bearing matrix associated with the DSD process
\begin{equation}\label{eqmeshr}
\{\Omega^\mathpzc{R}_\ell,\boldsymbol{B}^\mathpzc{R}_{\ell,:}\}_{\ell=1}^{n_{\Omega^\mathpzc{R}}} = \mbox{DSD}\Big( \{\mathcal{S}_j\}_{j\in\Ip\cup\In}\Big).
\end{equation}
When an exterior shape is added to the collection, depending on its overlap with the present elements $\{\mathcal{S}_j\}_{j\in\Ip\cup\In}$, the DSD process may produce finer partitions. Strictly speaking, when $n_s > n_\oplus+n_\ominus$ and
\begin{equation*}
\{\Omega_i,\boldsymbol{B}_{i,:}\}_{i=1}^{n_{\Omega}} = \mbox{DSD}\Big( \{\mathcal{S}_j\}_{j=1}^{n_s}\Big),
\end{equation*}
there exist index sets $\mathcal{J}_\ell$, such that
\begin{equation}\label{eqmesh}
\Omega^\mathpzc{R}_\ell = \bigcup_{i\in \mathcal{J}_\ell}\Omega_i, \qquad \ell = 1,2,\cdots, n_{\Omega^\mathpzc{R}}.
\end{equation}

We  refer to the finer shapelets $\{\Omega_i\}_{i=1}^{n_\Omega}$ as {\bf cells}. As stated before, successful identification of a composition relies on its coherence with the exterior shapes. However, adding an exterior element to the dictionary requires an entire update of the cellular architecture. For this reason, it is more convenient to look into a reversed process, where a collection of disjoint closed sets, yet referred to as cells, is fixed and each shape in the dictionary is produced by by making a union over a number of them (for such setting (\ref{eqmesh}) still holds). As a non-exclusive case, the cells could be considered as the image pixels.

As another contributing element of the geometric coherence, we denote by $\Gamma_1^\mathpzc{R}$ and $\Gamma_0^\mathpzc{R}$, the unit and null-valued shapelets corresponding to the linkage process (\ref{eqlink}). The partial bearing matrix $\boldsymbol{B}_{\Gamma_0^\mathpzc{R}\cup\Gamma_1^\mathpzc{R},:}$ is a non-singular square matrix of width $n_\oplus+n_\ominus$, as a basic non-redundant composition $\mathpzc{R}_{\Ip,\In}$ has exactly $n_\oplus$ unit-valued, and $n_\ominus$ null-valued shapelets.
The following result is adopted from \cite{aghasi2015convex} with a slight sign modification.

\begin{theorem}[Bearing Constants]  \label{lemcons} Let $\boldsymbol{c}\in \mathbb{R}^{n_\oplus +n_\ominus}$, where $\boldsymbol{c}_{\Ip} = \boldsymbol{1}$ and $\boldsymbol{c}_{\In} = -\boldsymbol{1}$. Considering the basic non-redundant composition $\mathpzc{R}_{\Ip,\In}$, the linear system
\begin{equation}\label{eqbw}
\big(\boldsymbol{B}_{\Gamma_0^\mathpzc{R}\cup\Gamma_1^\mathpzc{R},:}^\mathpzc{R}\big)^T\boldsymbol{w}= \boldsymbol{c}
\end{equation}
has a unique solution, which satisfies $\boldsymbol{1} \preceq \boldsymbol{w}_{\Gamma_1^\mathpzc{R}}\preceq (1+n_\ominus)\boldsymbol{1} $ and $-\boldsymbol{1} \preceq \boldsymbol{}\boldsymbol{w}_{\Gamma_0^\mathpzc{R}}\prec \boldsymbol{0} $.
\end{theorem}
Solving (\ref{eqbw}) for $\boldsymbol{w}$ assigns strictly positive quantities to the unit-valued shapelets, and strictly negative quantities to the null-valued shapelets. The entries of $\boldsymbol{w}$ merely depend on the architecture of $\mathpzc{R}_{\Ip,\In}$ and for this reason referenced as the {\bf bearing constants}. 

The {\bf geometric coherence} between an exterior shape and the target composition $\mathpzc{R}_{\Ip,\In}$ is defined by
\begin{equation}
	\label{eqcoherence}
\mbox{Coh}(\mathcal{S}_j,\mathpzc{R}_{\Ip,\In}) \triangleq \big|  \sum_{\ell\in \Gamma_0^\mathpzc{R}\cup \Gamma_1^\mathpzc{R}} \gamma_{\ell,j}\boldsymbol{w}_\ell\big |, \qquad j\in (\Ip\cup\In)^c,
\end{equation}
where the quantities $\gamma_{\ell,j}\in[0,1]$ measure the cellular overlap between an exterior shape and a null or unit-valued shapelet, and are calculated as
\begin{equation*}
 \gamma_{\ell,j} \triangleq \frac{1}{|\mathcal{J}_\ell|}    \sum_{i\in\mathcal{J}_\ell}   1_{\{\Omega_i\subset\mathcal{S}_j\}},\quad  \ell\in \Gamma_0^\mathpzc{R}\cup\Gamma_1^\mathpzc{R} , \;\; j= n_\oplus + n_\ominus + 1, \cdots, n_s.
\end{equation*}
For instance, an exterior shape that does not overlap with any elements of $\mathpzc{R}_{\Ip,\In}$ has a zero geometric coherence with the target composition.  

In general the construction of the linkage process enforces that $\mathcal{L_{\balpha_\mathpzc{R}}}(x)\geq 1$ inside $\mathpzc{R}_{\;\Ip,\In}$ and $\mathcal{L_{\balpha_\mathpzc{R}}}(x)\leq 0$ outside $\mathpzc{R}_{\;\Ip,\In}$. As a result, when LOC holds for $\overline{\mathpzc{R}_{\;\Ip,\In}}$, the entries of the vector $\boldsymbol{e}$ (defined in (\ref{ee-10}) and referred to as the {\bf LOC violation vector}) are all zero. Also, over the null and unit-valued cells:
\begin{equation}
\left\{\begin{array}{lc}p_i>0=q_i & \forall i\in\mathcal{J}_\ell, \ell\in \Gamma_0 \\ q_i>0=p_i & \forall i\in\mathcal{J}_\ell, \ell\in \Gamma_1 \end{array} \right..
\end{equation}
It is reasonable to refer to the entries of $\boldsymbol{e}$ and the quantities $\{q_i:i\in\mathcal{J}_\ell, \ell\in \Gamma_0\}$ and $\{p_i:i\in\mathcal{J}_\ell, \ell\in \Gamma_1\}$ as the LOC violation quantities, since they vanish when LOC holds for the target composition. These quantities may take nonzero values when there is a misalignment between the object present in the image and $\overline{\mathpzc{R}_{\Ip,\In}}$, and/or the image clutter or noise cause (\ref{eqloccon}) to be violated in measurable portions of the image. The following theorem warrants the recovery of a target composition under a sufficiently small LOC violation and limited coherence of the exterior shapes with $\mathpzc{R}_{\Ip,\In}$:

\begin{theorem}[Unique Recovery of a Target Composition] \label{thglast}
\emph{Following the preceding setup, suppose the LOC violation is sufficiently small for $\mathpzc{R}_{\;\Ip,\In}$. Corresponding to each cell $\Omega_i$, $i\in\mathcal{J}_\ell$, $\ell\in \Gamma_1^\mathpzc{R}\cup\Gamma_0^\mathpzc{R}$, and each exterior shape $\mathcal{S}_j$, $j\in\{ n_\oplus+n_\ominus+1, \cdots, n_s \}$, there exist $\epsilon_i$ and $\delta_j$ directly related to the LOC violation quantities, that if
\begin{equation}\label{eqcond}
\left\{\begin{array}{lc}p_i>\frac{|w_\ell|}{|\mathcal{J}_\ell|}(\eta_c-\epsilon_i) & \forall i\in\mathcal{J}_\ell, \ell\in \Gamma_0 \\ q_i>\frac{|w_\ell|}{|\mathcal{J}_\ell|}(\eta_c-\epsilon_i) & \forall i\in\mathcal{J}_\ell, \ell\in \Gamma_1 \end{array} \right.,
\end{equation}
and
\begin{align}\label{eqcoh}
\mbox{Coh}(\mathcal{S}_j,\mathpzc{R}_{\Ip,\In})<1-\frac{\delta_j}{\eta_c},\; \forall j\in\{ n_\oplus\!+\!n_\ominus\!+\!1, \cdots, n_s \}
\end{align}
for a fixed $\eta_c> \max\Big\{0, \big\{\epsilon_i \big\}_{i\in\mathcal{J}_\ell, \ell\in \Gamma_1^\mathpzc{R}\cup\Gamma_0^\mathpzc{R}}\Big\}$, then the unique minimizer of the convex program (\ref{eqcon2}) is $\balpha^*$, obeying $\balpha^*_{\Ip\cup\In}=\balpha_\mathpzc{R}$ and $\balpha^*_{(\Ip\cup\In)^c}=\boldsymbol{0}$.}
\end{theorem}

Theorem \ref{thglast} poses a stronger result compared to Theorem 4.10 of \cite{aghasi2015convex}, in two main aspects. First, the latter makes a perfect LOC assumption for  $\overline{\mathpzc{R}_{\;\Ip,\In}}$, while 
here a general setup is considered. Second, even under the LOC assumption that the quantities $\epsilon_i$ and $\delta_j$ vanish, the counterpart is in hold of an additional rank constraint (imposed on the overall dictionary bearing matrix row-supported on $\Gamma_0\cup\Gamma_1$), while here this requirement is eliminated thanks to Theorem \ref{thunique}. This pruning is important since the previous result was valid
when $n_s<|\Gamma_0\cup \Gamma_1|$ (see the discussion in \cite{aghasi2015convex}), while here no such limitation holds on $n_s$, and the number of the dictionary elements can be arbitrarily large (even larger than the number of cells or image pixels). 

To complete the discussion, we proceed by elaborating on the dependence of $\epsilon_i$ and $\delta_j$ in Theorem \ref{thglast} to the LOC violation quantities. In this regard 
\begin{equation}
\epsilon_i = \left\{\begin{array}{rll}
 \frac{1}{|\boldsymbol{w}_\ell|}(  \boldsymbol{\varepsilon}^{LV}_\ell - q_i|\mathcal{J}_\ell|) & i\in \mathcal{J}_\ell,& \ell \in \Gamma_0^\mathpzc{R}\\
-\frac{1}{|\boldsymbol{w}_\ell|}(  \boldsymbol{\varepsilon}^{LV}_\ell + p_i|\mathcal{J}_\ell|) & i\in \mathcal{J}_\ell,& \ell \in \Gamma_1^\mathpzc{R}
 \end{array}
\right.
\end{equation}
where,
\begin{equation}\label{locviol}
\boldsymbol{\varepsilon}^{LV} \triangleq \big(\boldsymbol{B}_{\Gamma_0^\mathpzc{R}\cup\Gamma_1^\mathpzc{R},:}^\mathpzc{R}\big)^{-T} \boldsymbol{e}_{\Ip\cup\In},
\end{equation}
and $(.)^{-T}$ denotes the inverse of the transpose matrix. The vector $\boldsymbol{\varepsilon}^{LV}\in\mathbb{R}^{n_\oplus + n_\ominus}$,  depends on the LOC violation over the elements of the target composition and the conditioning of $\boldsymbol{B}_{\Gamma_0^\mathpzc{R}\cup\Gamma_1^\mathpzc{R},:}^\mathpzc{R}$. For small values of $s=|\Ip\cup\In|$ a better conditioning is expected. Specifically, in the case of $\In=\emptyset$, the underlying matrix reduces to a permutation matrix, which essentially offers the best conditioning.

To present $\delta_j$, let $T$ denote the index set associated with the cells that do not overlap with the constituting elements of $\mathpzc{R}_{\;\Ip,\In}$, i.e.,
\begin{equation}\label{Tdef}
T = \Big\{i:\Omega_i \subset \Big(\bigcup_{j=1}^{n_s} \mathcal{S}_j\Big) \big\backslash \Big(\bigcup_{j=1}^{n_\oplus+n_\ominus} \mathcal{S}_j\Big)  ,i\in\{1,2,\cdots,n_\Omega\}\Big\}.
\end{equation}
As derived in the proof, for $j\in\{n_\oplus+n_\ominus+1,\cdots,n_\Omega\}$, 
\begin{equation}\label{eqdeltaj}\delta_j = \Big|
        -\boldsymbol{e}_j
        +   \sum_{\ell\in \Gamma_0^\mathpzc{R}\cup \Gamma_1^\mathpzc{R}} \gamma_{\ell,j}\boldsymbol{\varepsilon}^{LV}_\ell\Big| + \sum_{i\in T\cap\mathcal{I}_j} (q_i-p_i)^+ ,
\end{equation}
where, following (\ref{eqIj}), the index set $\mathcal{I}_j$ indicates the cells within each $\mathcal{S}_j$.

For a dictionary which only consists of $\{\mathcal{S}_j\}_{j\in\Ip\cup\In}$, condition (\ref{eqcond}) guarantees that the solution to the CSC is $\balpha_\mathpzc{R}$. For a larger dictionary which consists of exterior shapes in addition to $\{\mathcal{S}_j\}_{j\in\Ip\cup\In}$, the geometric coherence condition (\ref{eqcoh}) must be met for each exterior element to warrant an accurate recovery of the target composition.  

The conditions stated in Theorem \ref{thglast} allow us to characterize the solutions of the CSC program. We conclude this section by providing some general discussions relating the outcome of the CSC to the minimizer of the original shape composition problem (\ref{eq5}). From (\ref{eq5x}) it is straightforward to see that 
\begin{equation*}
G(\balpha) =\int_{D}\Big(\Delta^+\max\big(\mathpzc{L}_{\boldsymbol{\alpha}},0\big) + \Delta^-\min\big(\mathpzc{L}_{\boldsymbol{\alpha}},1\big)\Big)\;\mbox{d}x.
\end{equation*}
Suppose that $\balpha$ characterizes a composition $\mathpzc{R}_{\;\Ip,\In}$, i.e., $\mathpzc{L}_{\balpha}(x)\geq 1$, for $x\in \mathpzc{R}_{\;\Ip,\In}^o$ and $\mathpzc{L}_{\balpha}(x)\leq 0$, for $x\in D\setminus\overline{\mathpzc{R}_{\;\Ip,\In}}$.
We can define $D_1 = \{x:\mathpzc{L}_{\boldsymbol{\alpha}}(x)=1\}$, $D_0 = \{x:\mathpzc{L}_{\boldsymbol{\alpha}}(x)=0\}$, $D_{1^+} = \{x:\mathpzc{L}_{\boldsymbol{\alpha}}(x)>1\}$ and $D_{0^-} = \{x:\mathpzc{L}_{\boldsymbol{\alpha}}(x)<0\}$, and use $\int_D = \int_{D_1}+\int_{D_0}+\int_{D_{1^+}}+\int_{D_{0^-}}$ to verify that
\begin{equation}\label{eqpos}
\int_D\Delta^+\max(\mathpzc{L}_{\boldsymbol{\alpha}},0) \mbox{d}x= \int_{D_1\cup D_{1^+}}\Delta^+\mbox{d}x + \varepsilon_{1^+}(\balpha),
\end{equation}
where $\varepsilon_{1^+}(\balpha) = \int_{D_{1^+}}\Delta^+(\mathpzc{L}_{\boldsymbol{\alpha}}-1)\mbox{d}x\geq 0 $, and 
\begin{equation}\label{eqneg}
\int_D\Delta^-\min(\mathpzc{L}_{\boldsymbol{\alpha}},1) \mbox{d}x= \int_{D_1\cup D_{1^+}}\Delta^-\mbox{d}x + \varepsilon_{0^-}(\balpha),
\end{equation}
where $\varepsilon_{0^-}(\balpha) = \int_{D_{0^-}}(-\Delta)^+|\mathpzc{L}_{\boldsymbol{\alpha}}|\mbox{d}x\geq 0$. A combination of (\ref{eqpos}) and (\ref{eqneg}) yields
\begin{equation}\label{concomb}
G(\balpha) - E(\mathpzc{R}_{\;\Ip,\In}) = \varepsilon_{1^+}(\balpha)+\varepsilon_{0^-}(\balpha)\triangleq \varepsilon_{1^+,0^-}(\balpha) \geq 0. 
\end{equation}
When LOC holds for $\overline{\mathpzc{R}_{\;\Ip,\In}}$, or $D_{1^+}\cup D_{0^-}=\emptyset$  we get $\varepsilon_{1^+}(\balpha) = \varepsilon_{0^-}(\balpha) = 0$ and the inequality in (\ref{concomb}) turns into an equality. 

Now suppose $\{\hat{\mathcal{I}}_\oplus, \hat{\mathcal{I}}_\ominus\}$ is the unique outcome of the original shape composition problem (\ref{eq5}), $\hat\balpha_\mathpzc{R}$ is the outcome of the linkage process in response to $\{\hat{\mathcal{I}}_\oplus, \hat{\mathcal{I}}_\ominus\}$ (not necessarily unique) and $\hat{\balpha}\in\mathbb{R}^{n_s}$ is an $s$-sparse vector which matches $\hat\balpha_\mathpzc{R}$ over $\{\hat{\mathcal{I}}_\oplus, \hat{\mathcal{I}}_\ominus\}$. For the corresponding CSC program (\ref{eq5x}), consider $\balpha^*$ to be the unique solution which is $s$-sparse and identifies the index sets $\{\Ip^*,\In^*\}$. If $\|\hat\balpha\|_1\leq\tau=\big\|\mathcal{A}\big(\{\mathcal{S}_j\}_{j\in\Ip^*\cup\In^*};\Ip^*,\In^*\big)\big\|_1$, based on (\ref{concomb}) and the optimality of the solutions, the following relationships hold: (I) $G(\balpha^*) - E(\mathpzc{R}_{\;\Ip^*,\In^*})=\varepsilon_{1^+,0^-}(\balpha^*)$; (II) $G(\hat{\balpha}) - E(\mathpzc{R}_{\;\hat{\mathcal{I}}_\oplus,\hat{\mathcal{I}}_\ominus})=\varepsilon_{1^+,0^-}(\hat\balpha)$; (III) $G(\balpha^*)\leq G(\hat\balpha)$; (IV) $E(\mathpzc{R}_{\;\hat{\mathcal{I}}_\oplus,\hat{\mathcal{I}}_\ominus})\leq E(\mathpzc{R}_{\;\Ip^*,\In^*})$. A combination of these four relations yields 
\begin{equation}\label{eqbound}
0\leq E(\mathpzc{R}_{\;\Ip^*,\In^*}) - E(\mathpzc{R}_{\;\hat{\mathcal{I}}_\oplus,\hat{\mathcal{I}}_\ominus})\leq \varepsilon_{1^+,0^-}(\hat\balpha) - \varepsilon_{1^+,0^-}(\balpha^*), 
\end{equation}
which quantifies the CSC performance in approximating the solution of (\ref{eq5}). The most right-side expression in (\ref{eqbound}) is controlled by the level of element overlap and LOC violation of the solutions.
Broadly speaking, the conditions stated in Theorem \ref{thglast} provide sufficient conditions for the recovery of a composition with cardinality $s$, the cost value of which is at most $\varepsilon_{1^+,0^-}(\hat\balpha) - \varepsilon_{1^+,0^-}(\balpha^*)$ more than the minimizer of (\ref{eq5}).

The discussion above along with inequality (\ref{eqbound}) can be framed into a more technical context. We omit such extensions due to the article length restrictions and only discuss a simple scenario. Consider $\hat{\mathcal{I}}_\oplus$ to be the unique solution of (\ref{eq5}) and $\balpha^*$ to be the unique CSC solution which identifies $\Ip^*$. In this case $\hat{\balpha}$ and $\balpha^*$ are both $s$-sparse and take unit values over $\hat{\mathcal{I}}_\oplus$ and $\Ip^*$, respectively. If $\varepsilon_{1^+,0^-}(\hat\balpha)=0$ (e.g., this happen when LOC holds for $\mathpzc{R}_{\;\hat{\mathcal{I}}_\oplus}$ or the elements of $\mathpzc{R}_{\;\hat{\mathcal{I}}_\oplus}$ have no overlaps), we must have $\hat{\mathcal{I}}_\oplus = \Ip^*$. This is simply because, if $\hat{\mathcal{I}}_\oplus \neq \Ip^*$, the inequalities in (\ref{eqbound}) become strict and form  the contradiction $0<-\varepsilon_{1^+,0^-}(\balpha^*)$.

\section{Numerical Experiments}\label{sec:Sim}
While the focus of this paper is the CSC underlying theory and the presentation of computational tools, in this section we present some numerical experiment highlighting the general performance of CSC. Several other experiments in applications such as OCR, principal shape extraction and solving jigsaw puzzles, have been presented in \cite{aghasi2015convex} using prototype code and the popular CVX package \cite{cvx}. In this paper, we present some new large-scale experiments which include noisy object identification, multi-resolution shape representation (in 3D), and OCR with an automated way of constructing the shape dictionary.  

For the experiments presented, we use the LP reformulation of the CSC and use Gurobi \cite{gurobi} to address the resulting problem\footnote{Instances of the MATLAB code are available at the authors' webpage, currently: http://web.mit.edu/aghasi/www/software.html.}. The proposed reformulation allows addressing the CSC
hundreds of times faster than the previous  implementation.  As a concrete example, to address the experiment in \S 5.3 of \cite{aghasi2015convex}, an average runtime of approximately 8 minutes is reported, while using the dual LP formulation presented in Section \ref{sec:lp}, an identical problem can be addressed in less than a second on a standard desktop computer (3.4 GHz CPU and 16GB memory).

\subsection{Noisy Object Identification}

As discussed earlier, the proposed framework can be considered as a new way of regularizing object identification problems, beyond the standard surface or volume regularizers. The proposed scheme limits the reconstructions to objects constituting of fewer geometric components. 

\begin{figure}[htb]
\centering\begin{tabular}{cccc}
 \begin{overpic}[trim={0 -1cm  0 0},clip,width=1.4in]{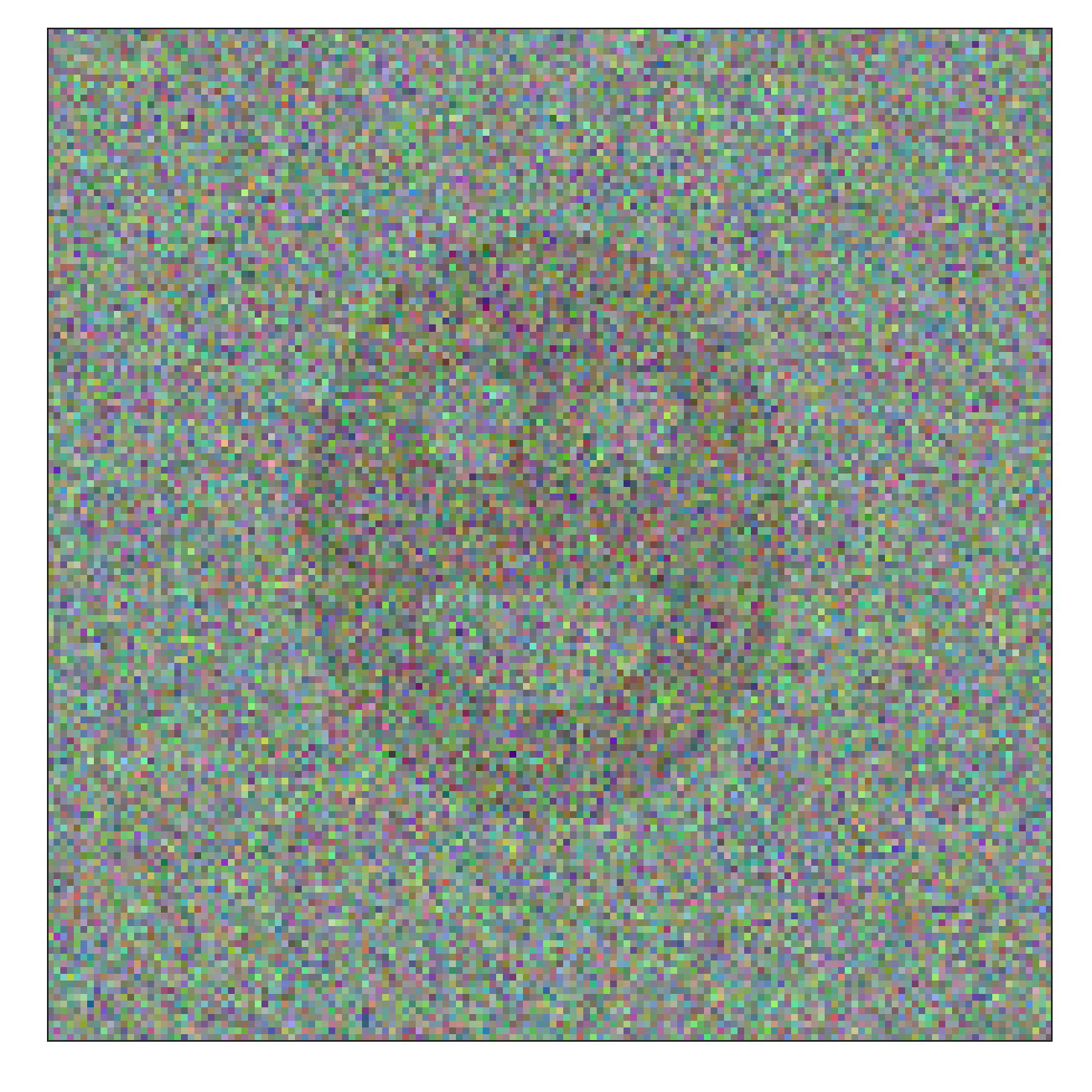}\put (40,-2) {\scriptsize(a)} \end{overpic}
&
\hspace{-.42cm}\begin{overpic}[trim={0 -1cm  0 0},clip,width=1.4in]{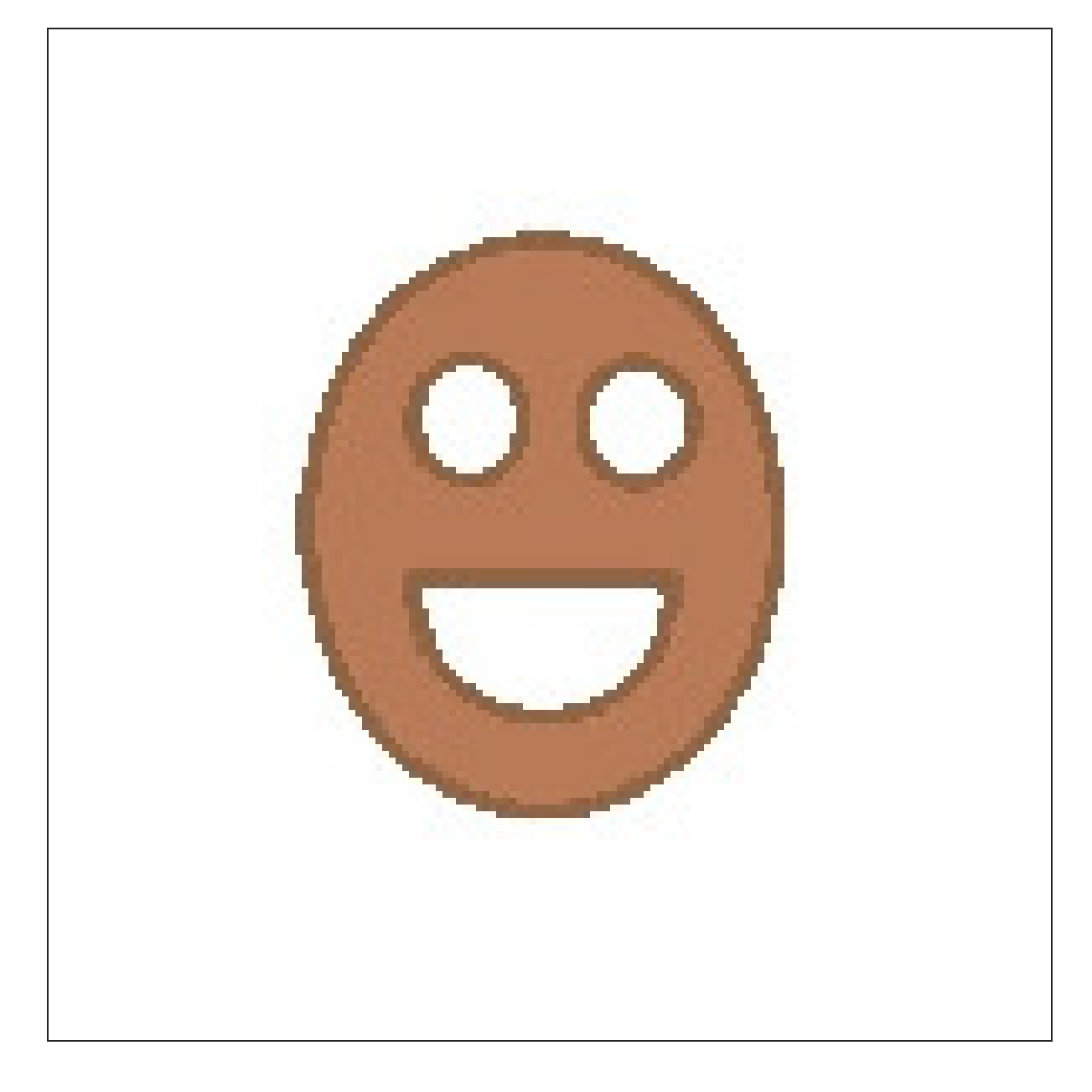}\put (40,-2) {\scriptsize(b)} \end{overpic}
&
\hspace{-.42cm}\begin{overpic}[trim={0 -1cm  0 0},clip,width=1.4in]{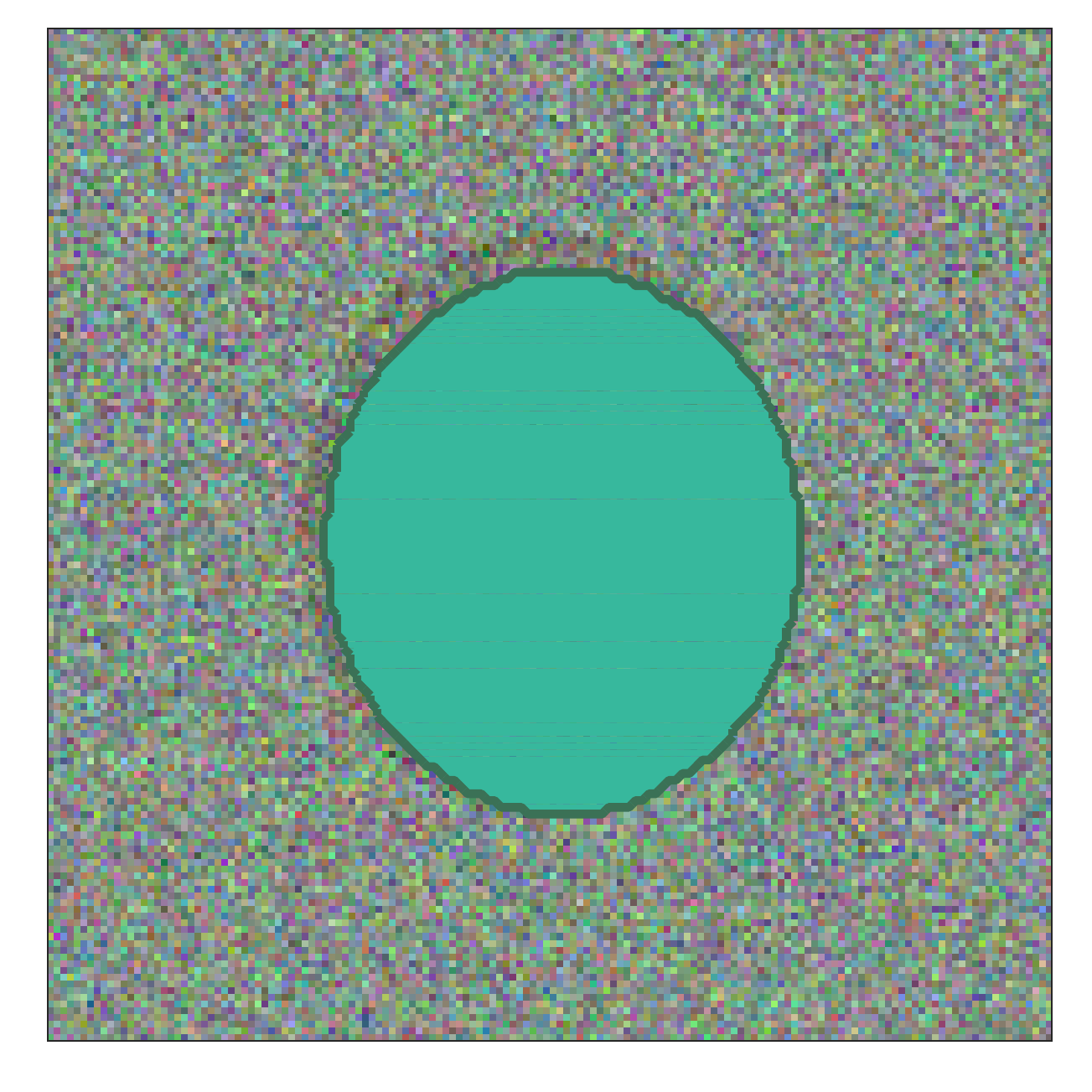}\put (40,-2) {\scriptsize(c)} \end{overpic}
&
\hspace{-.42cm}\begin{overpic}[trim={0 -1cm  0 0},clip,width=1.4in]{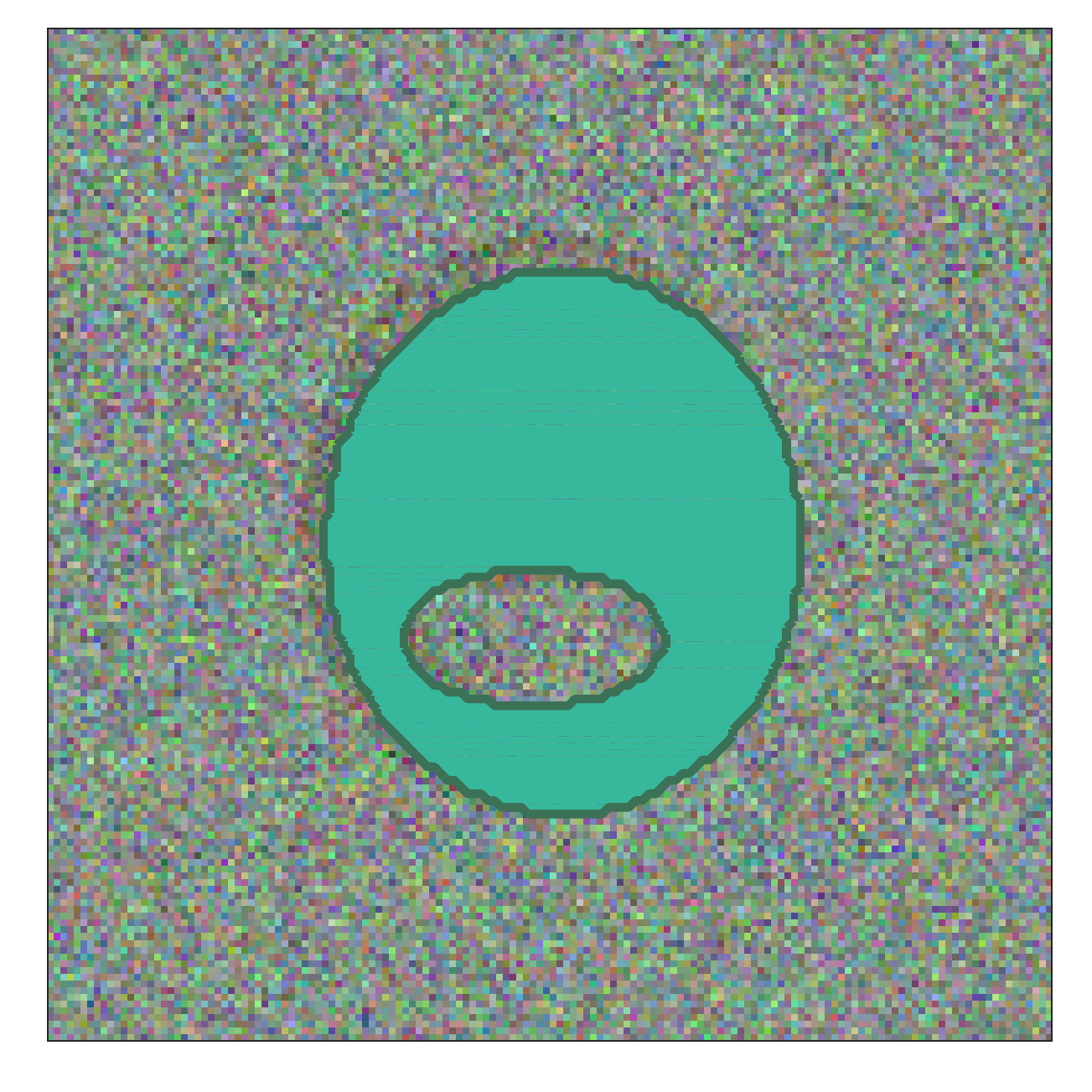}\put (40,-2) {\scriptsize(d)} \end{overpic}
\\
 \begin{overpic}[trim={0 -1cm  0 0},clip,width=1.4in]{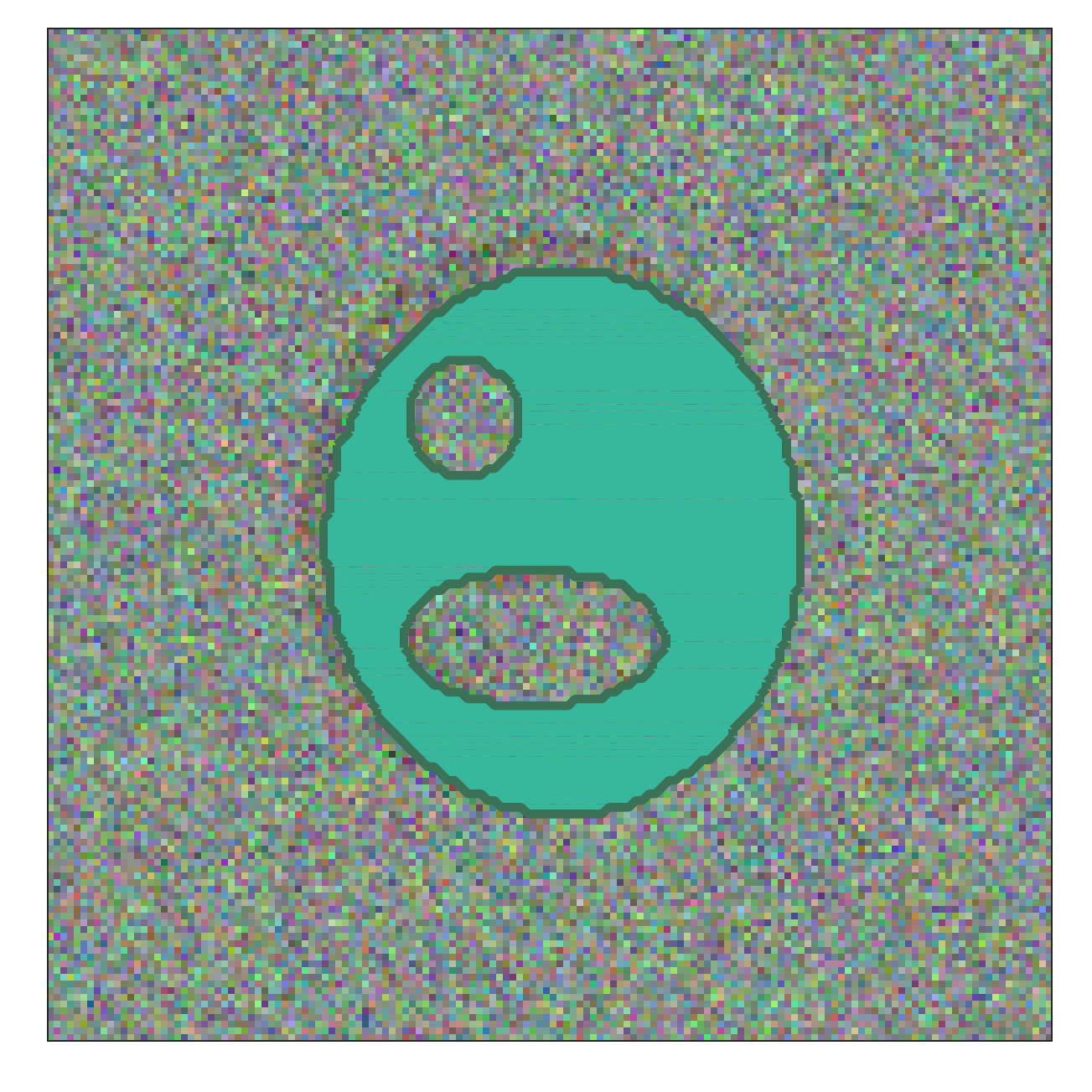}\put (40,-2) {\scriptsize(e)} \end{overpic}
&
\hspace{-.42cm}\begin{overpic}[trim={0 -1cm  0 0},clip,width=1.4in]{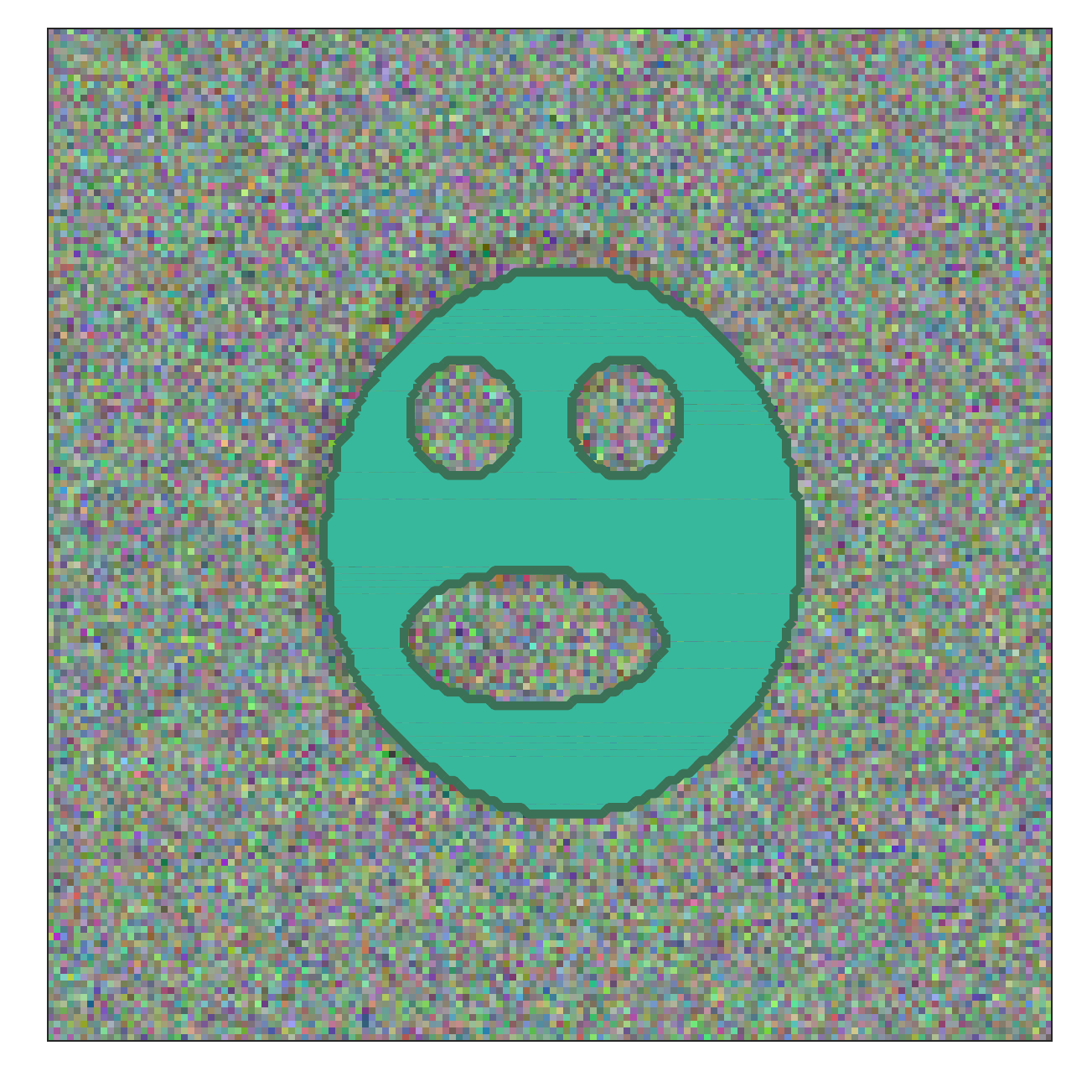}\put (40,-2) {\scriptsize(f)} \end{overpic}
&
\hspace{-.42cm}\begin{overpic}[trim={0 -1cm  0 0},clip,width=1.4in]{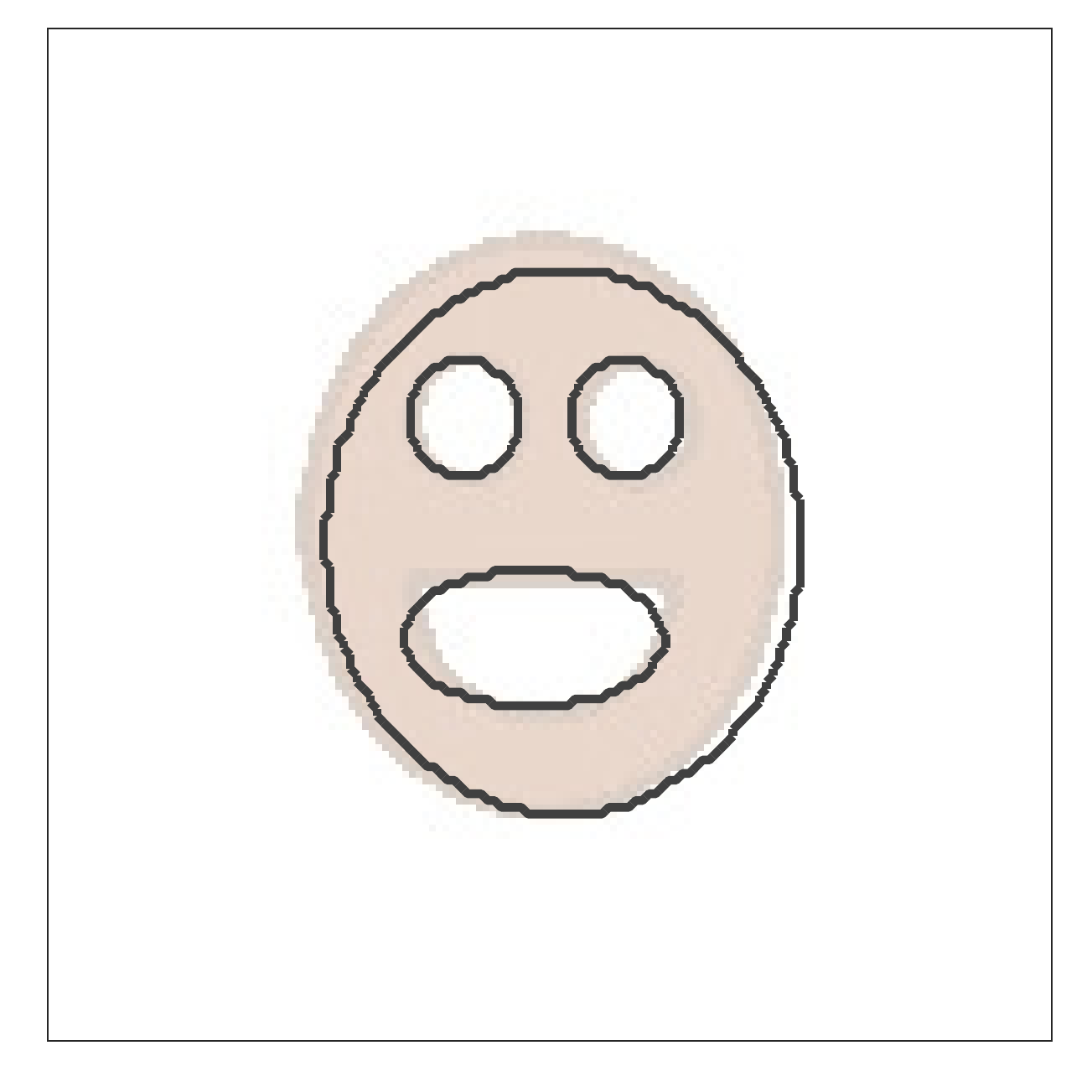}\put (40,-2) {\scriptsize(g)} \end{overpic}
&
\hspace{-.42cm}\begin{overpic}[trim={0 -1cm  0 0},clip,width=1.4in]{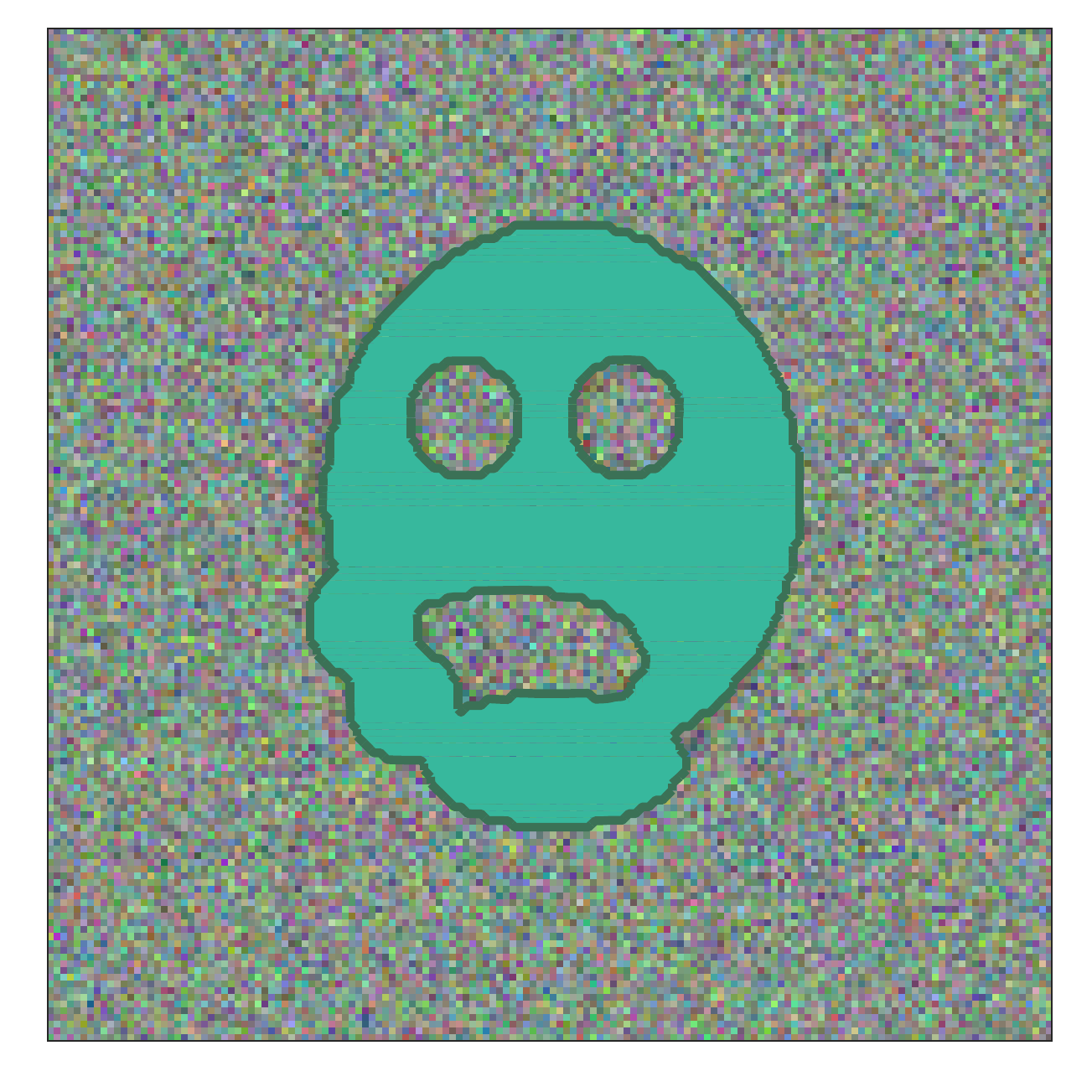}\put (40,-2) {\scriptsize(h)} \end{overpic}
\end{tabular}
\caption{Object identification applied to a noisy image; (a) the noisy image; (b) original image; (c-f) reconstructions using the constrained CSC for respectively $\tau=1,2,3$ and $4$; (g) comparing the alignment of the recovered object with the true object; (h) CSC recovery for $\tau=8$}\label{fig2}
\end{figure}

Figure \ref{fig2}(a) shows a color smiley image of size $150\times 150$ pixels corrupted by Gaussian noise. The signal to noise ratio (SNR) is -14dB. The clean actual image is shown in Figure \ref{fig2}(b). The shape dictionary used for this example consists of circular and elliptical disks at different scales, centered at regular grid points throughout the imaging domain. The dictionary consists of 3651 elements. 

The ultimate goal is the recovery of the object with reference to the prior information embedded into the construction of the shape dictionary. The measure of inhomogeneity used for the color image consists of the terms $\Pi_{in/ex}(x)=\sum_{c}(u^{(c)}(x) - {\tilde u}_{in/ex}^{(c)})^2$, where $c$ sweeps the RGB channels. The mean intensities $\tilde u_{in}$ and $\tilde u_{ex}$ are acquired using a standard binary k-means clustering.  

The outcomes of the constrained CSC for various values of $\tau=1,2,3,4$, are shown in Figures \ref{fig2}(c-f). The average runtime for these experiments is less than 3 seconds. We can see that progressively increasing the value of $\tau$ allows us to identify the main geometric components of the underlying object one after the other. It is also worth noting that the identification follows an ordered pattern, where the more major components are identified prior to the smaller objects. Figure \ref{fig2}(g) compares the alignment of the identified object with the original reference smiley. Stepping the value of $\tau$ beyond the level 4 allows the algorithm to capture more details, but at the expense of less meaningful outcomes due to the large image noise (Figure \ref{fig2}(h)). 

\subsection{Multi-Resolution Segmentation and Object Representation} An implication of the previous example is the use of CSC for multi-resolution shape representations. Basically, when the $\ell_1$ penalty is large, the CSC tends to identify the more bulky compositions which contribute more to the underlying objective. As the $\ell_1$ penalty reduces, the algorithm finds its way to contribute more element and capture more structural details. This bahvior, in a sense, evokes the wavelet recovery and denoising using $\ell_1$ regularization. In such a scheme, a larger $\ell_1$ penalty on the wavelet coefficients causes the coarser structures to be identified, while reducing the penalty identifies the signal components at finer scales. 

To demonstrate this behavior, we consider the reconstruction of a 3D object in an imaging domain of size $150\times 150\times 150$ voxels. The basic elements used in the dictionary are spheres of diameters 60, 30 and 15 voxels (Figure \ref{fig3}(a)). The spacing between the spheres is larger for the larger elements. The balls are placed on regular grids of sizes 6, 10 and 20 points (respectively, corresponding to the large, medium and small balls). This selection produces a dictionary of $6^3+10^3+20^3 = 9216$ elements.  

As the reference object to be reconstructed, we used the Stanford bunny depicted in Figure \ref{fig3}(b). For various values of $\lambda$, ranging from $10^4$ to $10^{-3}$, the outcomes of the regularized CSC are depicted in Figures \ref{fig3}(c-h). Below each reconstruction, the resulting coefficient vector $\balpha$ is plotted. The index range is partitioned into three intervals corresponding to different dictionary element scales (indication by different colors: red, pink and orange). For larger values of $\lambda$ most of the $\balpha$ energy is concentrated in the red and pink area, corresponding to the bulkier objects. For smaller $\lambda$ the CSC allows the recovery of finer details.

\subsection{OCR and Reducing the Dictionary Size}
We conclude the experiments with an OCR problem, as an application also considered in \cite{aghasi2015convex}. As the CSC allows identifying the principal shape elements through a segmentation task, it would fit well into an OCR framework, where the primary objective is the identification of letters present in an image. The basic dictionary elements for this problem are simply the characters that may appear in the image.

\begin{figure}[htb]
\centering\begin{tabular}{ccc}
 \begin{overpic}[trim={0 -1cm  0 0},clip,width=1.76in]{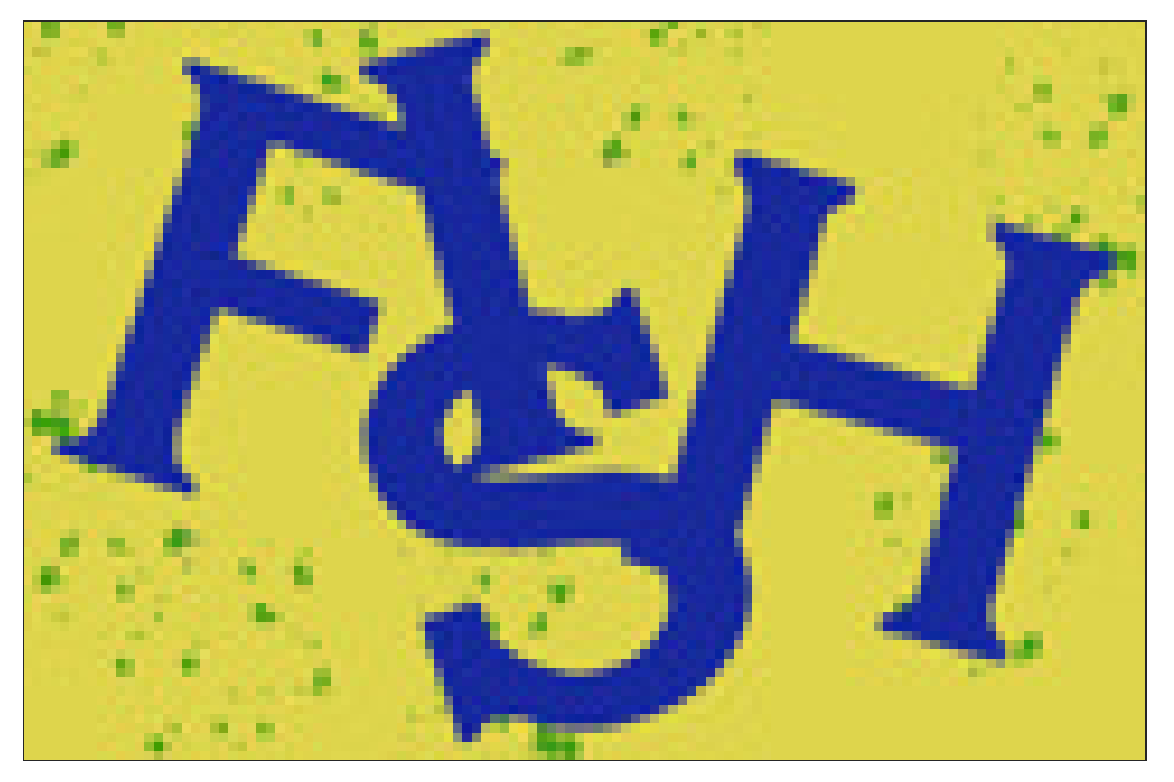}\put (45,-1) {\scriptsize(a)} \end{overpic}
&
\hspace{-.4cm}\begin{overpic}[trim={0 -1cm  0 0},clip,width=1.76in]{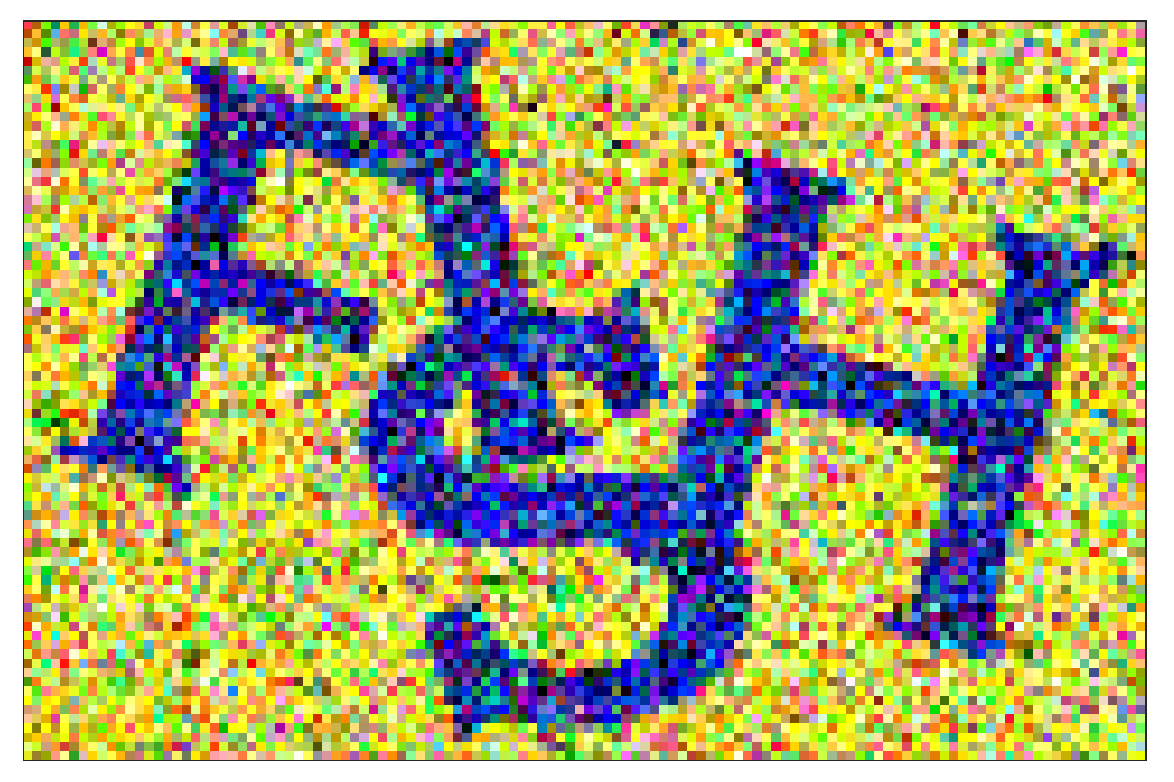}\put (45,-1) {\scriptsize(b)} \end{overpic}
&
\hspace{-.4cm}\begin{overpic}[trim={0 -1cm  0 0},clip,width=1.76in]{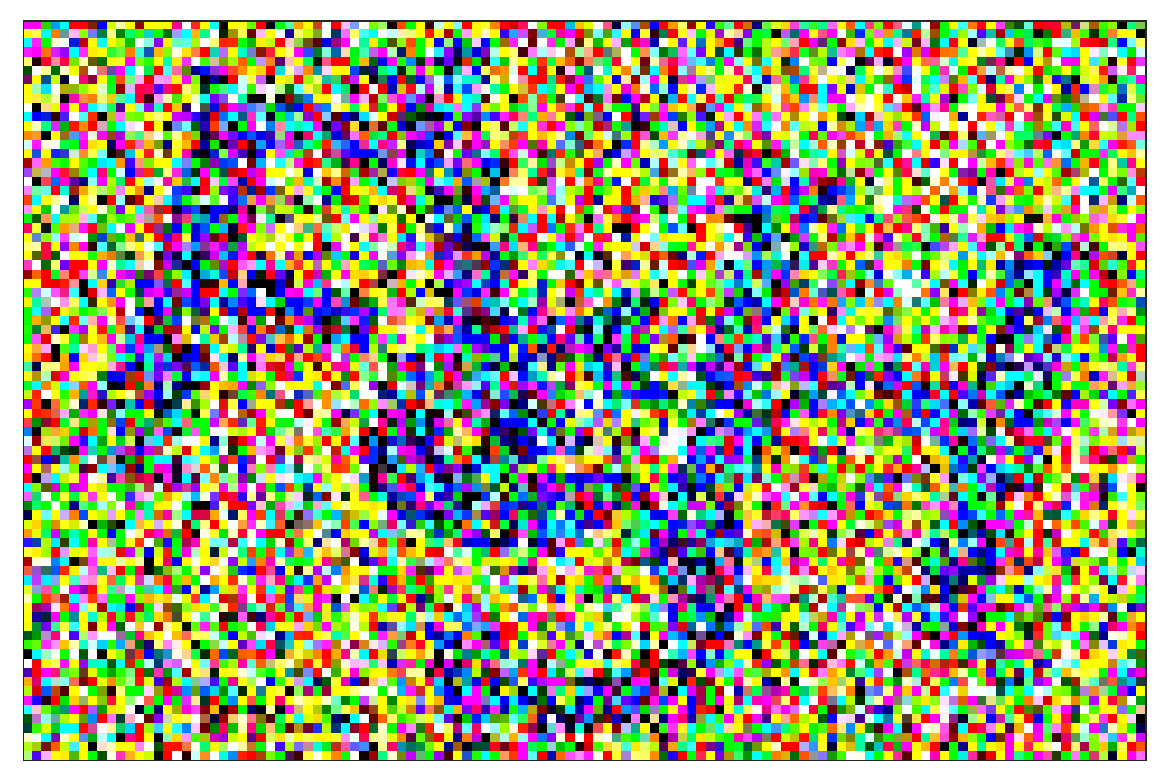}\put (45,-1) {\scriptsize(c)} \end{overpic}
\\
 \begin{overpic}[trim={0 -1cm  0 0},clip,width=.945in]{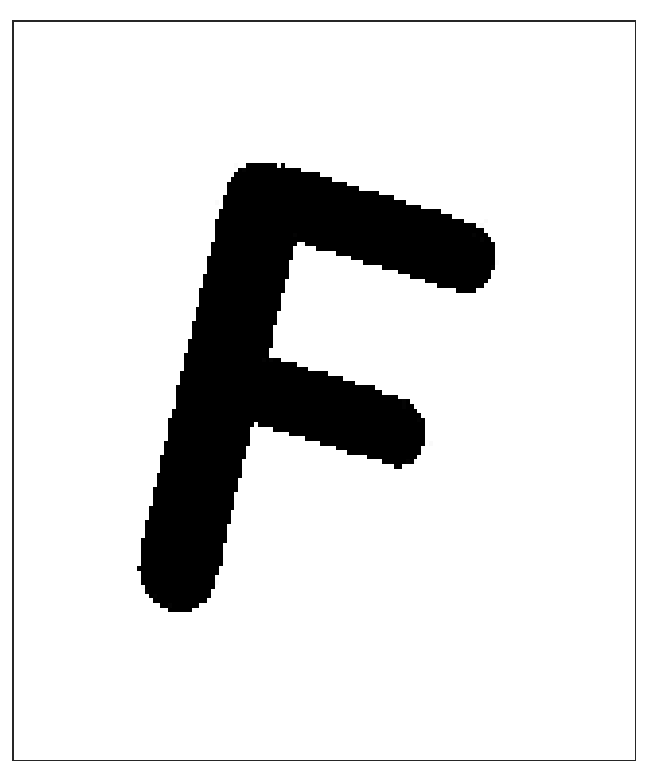}\put (30,-1) {\scriptsize(d)} \end{overpic}
&
\hspace{-.4cm}\begin{overpic}[trim={0 -1cm  0 0},clip,width=1.76in]{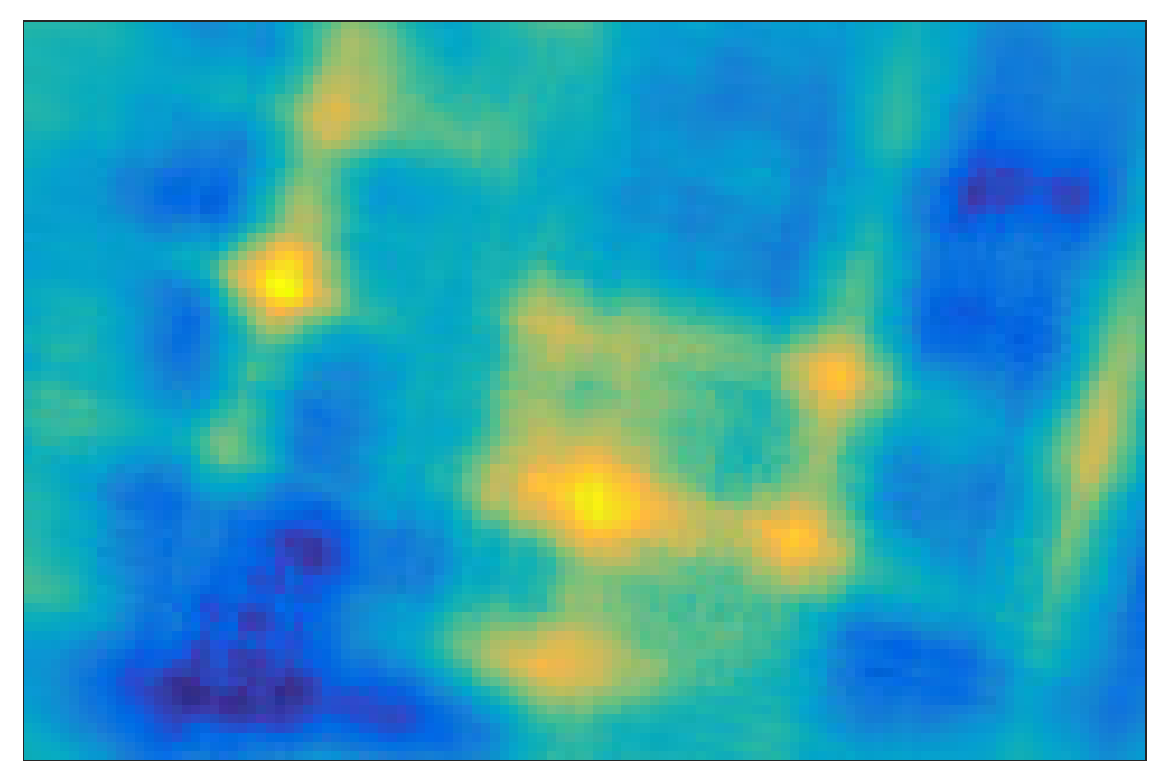}\put (45,-1) {\scriptsize(e)} \end{overpic}
&
\hspace{-.4cm}\begin{overpic}[trim={0 -1cm  0 0},clip,width=1.76in]{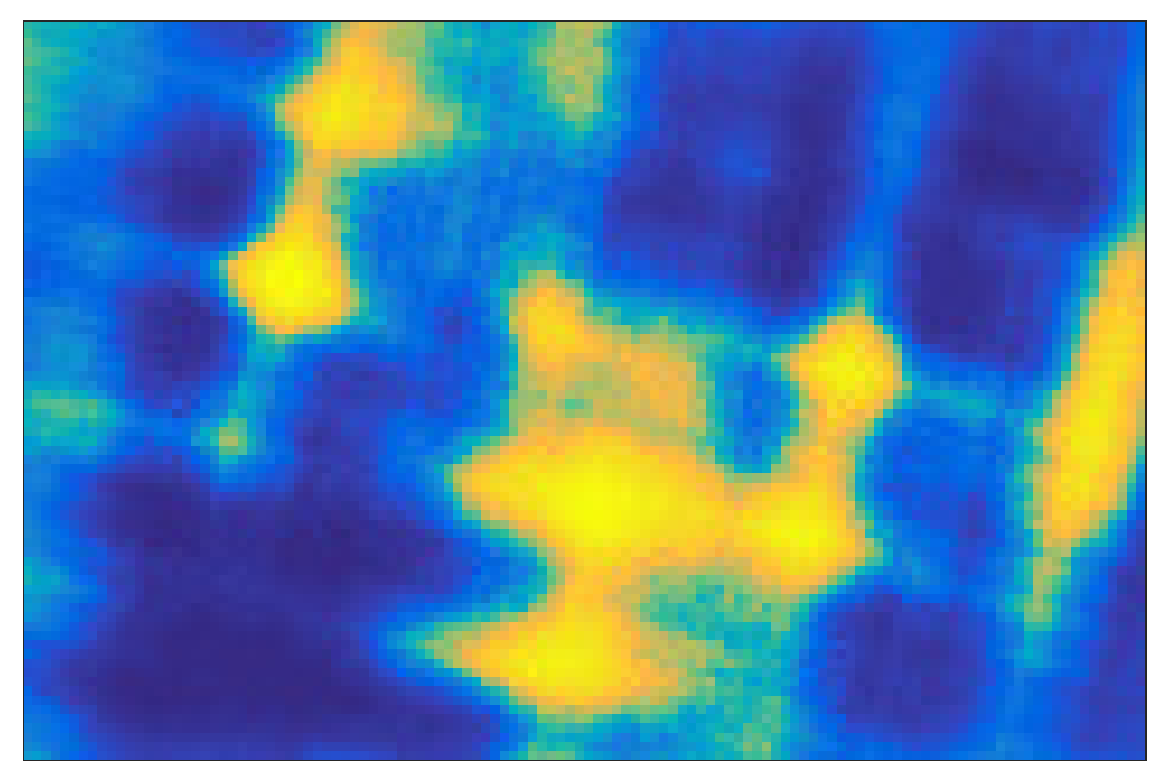}\put (45,-1) {\scriptsize(f)} \end{overpic}
\\
 \begin{overpic}[trim={0 -1cm  0 0},clip,width=1.76in]{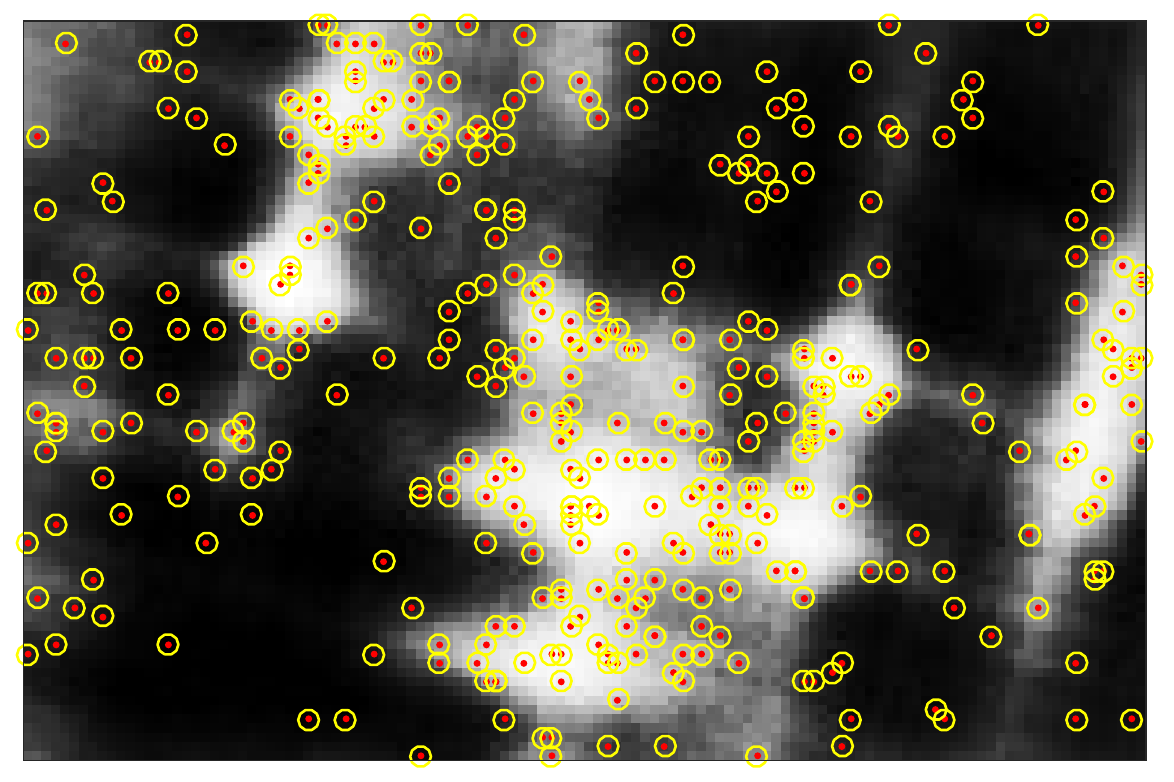}\put (45,-1) {\scriptsize(g)} \end{overpic}
&
\hspace{-.4cm}\begin{overpic}[trim={0 -1cm  0 0},clip,width=1.76in]{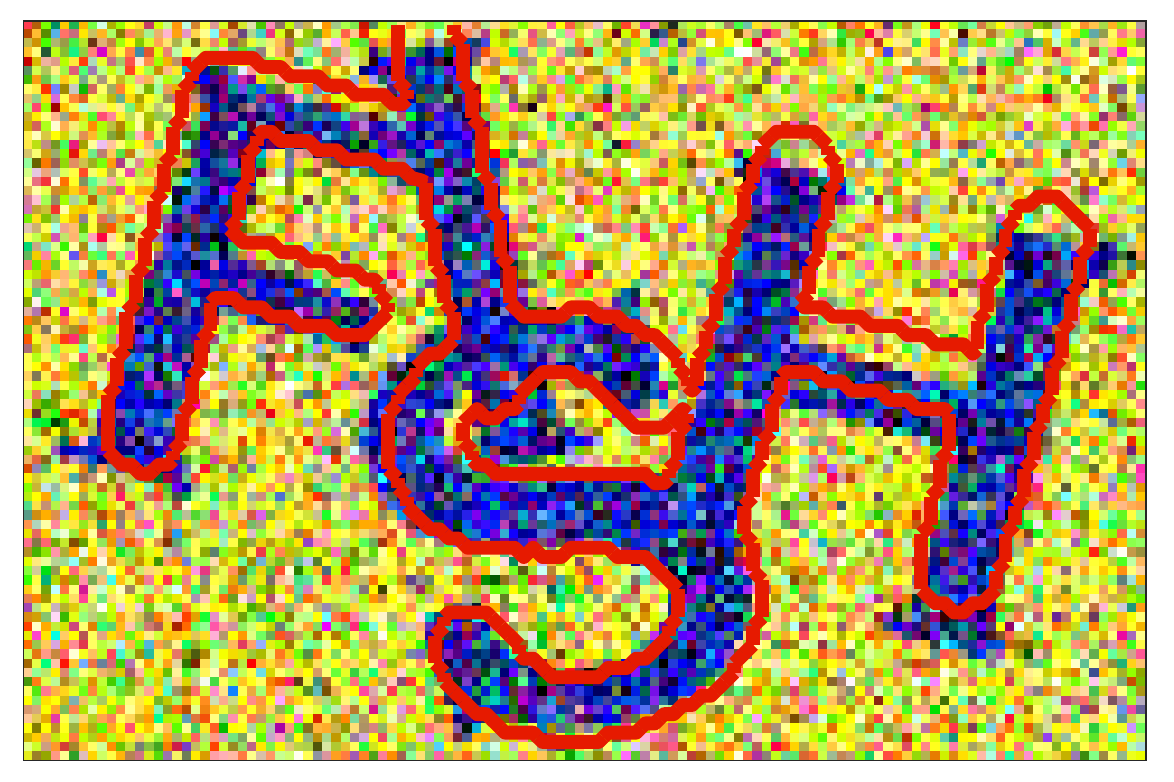}\put (45,-1) {\scriptsize(h)} \end{overpic}
&
\hspace{-.4cm}\begin{overpic}[trim={0 -1cm  0 0},clip,width=1.76in]{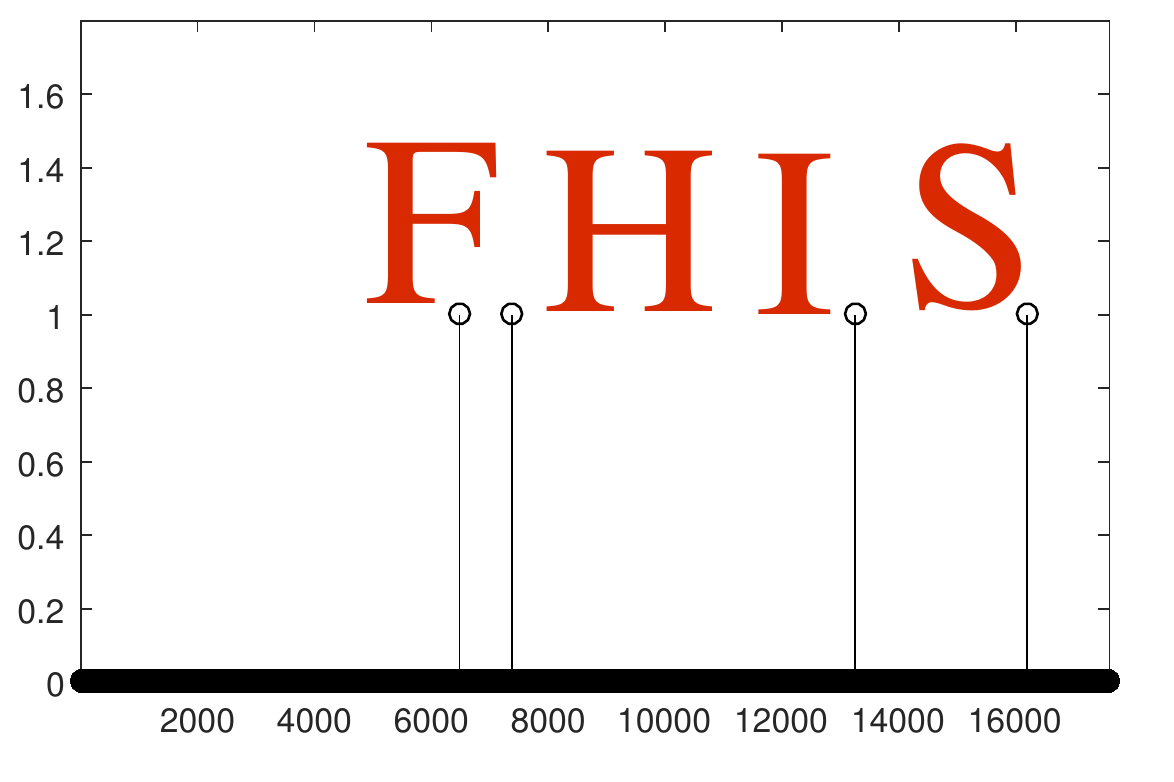}\put (45,-1) {\scriptsize(i)} \end{overpic}
\end{tabular}
\caption{An OCR problem: (a) the reference image; (b,c) noisy image versions (SNR = 0, SNR = -10 dB); (d) an element of the dictionary; (e) the cross correlation of the $\Delta$-image in panel (c) with the character in panel (d); (f) the outcome of applying a smooth rounding function ($\epsilon_r=0.3$) to the image in panel (e); (g) samples generated with reference to the pdf in panel (f); (h) CSC outcome for $\tau=4$, the red contours indicate the positive support of $\mathcal{L}_{\balpha^*}(x)$; (i) a plot of the $\balpha^*$ vector}\label{fig4}
\end{figure}

Unlike the previous two experiments, where shifted versions of the main shape elements were located throughout the imaging domain, in this experiment we use an automated way of generating the dictionary. Figures \ref{fig4}(a-c) show a reference OCR image and two noisy versions of the image for different SNR values. The characters in the underlying word are overlapping, which can make the OCR process challenging. As the elements of the dictionary, we use 26 uppercase letters, where to build up the dictionary instances of each letter are placed in different locations of the image. The font in the image and the font used in the dictionary are different (although somehow similar). Figure \ref{fig4}(d) shows a sample letter from the dictionary. 

In order to populate the dictionary with shifted and rotated versions of each character, we consider placing the characters in locations where there is more chance of such object presence. A simple cross correlation between the image and the character can provide us with such information. Specifically, consider constructing an image with pixel values $-\Delta(x)=\Pi_{ex}(x)-\Pi_{in}(x)$. Cross correlation of such image with a character characteristic function, $\chi_{\mathcal{S}}$, returns the values  $-E(\mathcal{S})$, when the center of $\mathcal{S}$ sweeps the entire domain. As an example, figure \ref{fig4}(e), shows the result of cross correlating the ``$\Delta$-image'' associate with panel (c) with the character shown in panel (d). The higher valued regions (yellow color) in panel (e) correspond to locations with higher chance of character matching. We can add a sufficiently large offset to the outcome of panel (e) and use a scaled version of the resulting function to generate a probability density function (pdf) based on which the centroids of the dictionary elements are determined. Figure \ref{fig4}(f) is a smoothly rounded version of the cross correlation image in panel (e) pushed towards a binary-valued function. Such contrast enhancement generates a pdf with more chance of generating centroids in critical regions. For example for $z\in[0,1]$, and $\epsilon_r$ being a small positive number, the function $r(z) = 0.5 + \pi^{-1}\arctan\big(\epsilon_r^{-1}(z-0.5)\big)$ is a smooth approximation of the rounding half up function, which was used to generate panel (f) from panel (e). Figure \ref{fig4}(g) shows 400 random centroids generated with reference to the pdf in panel (f). We can see that the majority of the centroids concentrate in regions of the image with more chance of character matching.

In summary, for each basic element (English letter) we follow the proposed process to generate a pdf. We draw 200 samples from the pdf and use them as the centroids where the basic element are placed in the dictionary. We repeat the process for two rotated versions of each character (angles $\pm 15$ degrees). The number of samples can be increased for the basic elements with higher cross correlations and more peaked pdfs (determined by a simple kurtosis test). For the top ten elements with such property we used 400 samples, generating a dictionary of size $n_s = (3\times 26 -10)\times 200 + 10\times 400 = 17600$ elements.

For $\tau=4$ we have presented an instance of the CSC outcome in Figure \ref{fig4}(h). It is noteworthy that CSC identifies the closest dictionary elements to the image characters, which are not necessarily perfectly aligned with the image content. A simple index map allows identifying the characters present in the image as demonstrated in Figure \ref{fig4}(i). Table \ref{tab1} reports the CSC success rate in identifying the entire word as well as each character for the given OCR image with different SNR values (200 experiments performed for each SNR). With the proposed limited number of centroid samples, yet an error-free identification is achieved in almost 86\% of the experiments. The most frequent misidentifications are F$\rightarrow$P, I$\rightarrow$L and S$\rightarrow$B. It is worth noting the rather insensitive nature of CSC to the SNR values which is justifiable by the embedded LP framework. Basically an LP solution can stay the same when the cost coefficient vector and the constraint matrix are perturbed  sufficiently small (image noise is the source of such perturbation in our problem).

\begin{table}[htbp]
    \centering
\caption{Success rate (SR) vs SNR for the initial dictionary}
    \label{tab:mytable}
\begin{tabular}{|c|c|c|c|c|c|}
    \specialrule{1.5pt}{0pt}{0pt}
\diagbox{\scriptsize SR\normalsize}{\scriptsize SNR}
          & 10 dB    & 5 dB     & 0 dB     & -5 dB    & -10 dB     \\
    \specialrule{1.5pt}{0pt}{0pt}
    Entire Word    & 0.855	& 0.86	& 0.84 &	0.87&	0.87    \\ \hline
    F     & 0.9	& 0.905 &	0.88	& 0.91 &	0.91     \\ \hline
    I     & 0.965	& 0.945 &	0.945 &	0.97	& 0.955      \\ \hline
    S     & 0.99 &	0.995	& 0.985	 & 0.975	& 0.975     \\ \hline
    H     & 0.995 &	0.99	& 0.985 &	0.99 &	1      \\ \hline
\specialrule{1.5pt}{0pt}{0pt}
\end{tabular}%
  \label{tab1}%
\end{table}

\begin{table}[htbp]
    \centering
\caption{SR vs SNR for the refined dictionary}
    \label{tab:mytable}
\begin{tabular}{|c|c|c|c|c|c|}
    \specialrule{1.5pt}{0pt}{0pt}
\diagbox{\scriptsize SR\normalsize}{\scriptsize SNR}
          & 10 dB    & 5 dB     & 0 dB     & -5 dB    & -10 dB     \\
    \specialrule{1.5pt}{0pt}{0pt}
    Entire Word    & 0.995&	0.985&	0.985&	0.985&	0.995    \\ \hline
    F     & 0.995	&0.99&	0.99&	0.99&	0.995    \\ \hline
    I     & 1	& 0.995 &	0.995 &	0.995 &	1     \\ \hline
    S     & 1 &	1	&1	&1	&1     \\ \hline
    H     & 1 &	1	&1	&1	&1      \\ \hline
\specialrule{1.5pt}{0pt}{0pt}
\end{tabular}%
  \label{tab2}%
\end{table}

The almost 14\% misidentification rate in Table \ref{tab1} can be significantly improved by populating the dictionary with elements of more closely spaced centroids. Table \ref{tab2} reports a similar experiment applied to a dictionary of 10 basic elements (the top ten frequent elements identified in the experiments of Table \ref{tab1}), each using 1000 centroids generated as above ($n_s = 10000$). We avoided to blow up the dictionary by taking dense samples of every basic element and instead only took dense samples of the most probable elements. Increasing the number of sample centroids in this case improved the accuracy to almost 99\%. Equipping CSC with other dictionary refinement or multi-stage verification tests is a future possible research towards real-world implementations of this framework for challenging and complex OCR problems. 

It is noteworthy that the theory imposes a limited geometric incoherence among the dictionary elements to recover a ``specific'' composition. However, in practice for dictionaries with highly coherent elements (e.g., the last OCR example), several close compositions may exist, identifying any of which is practically a good solution to the problem. In other words, even for dictionaries with geometrically coherent elements, CSC can still identify a solution with the desired level of cardinality, which reveals all the information about the characterized object.


%

\section{Derivation of the Proximal Operator}\label{AppendProx}
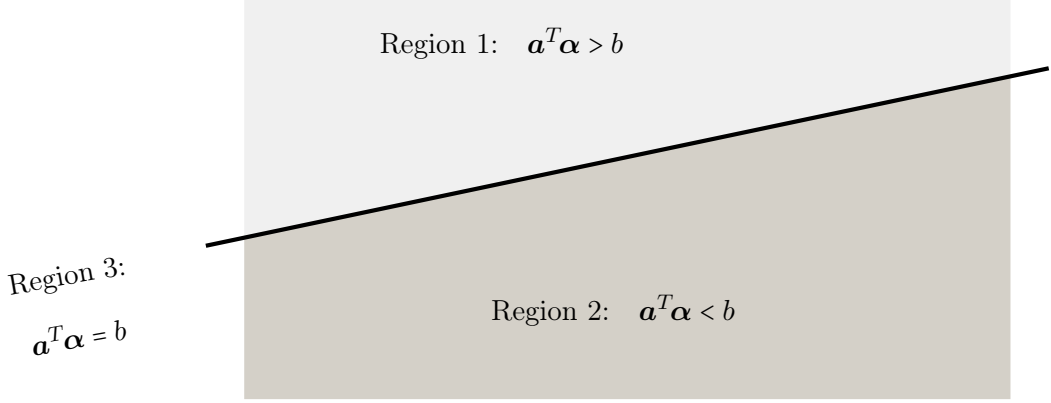
\begin{figure}[t]
\hspace{.3cm}\begin{overpic}[width=.98\textwidth,tics=10]{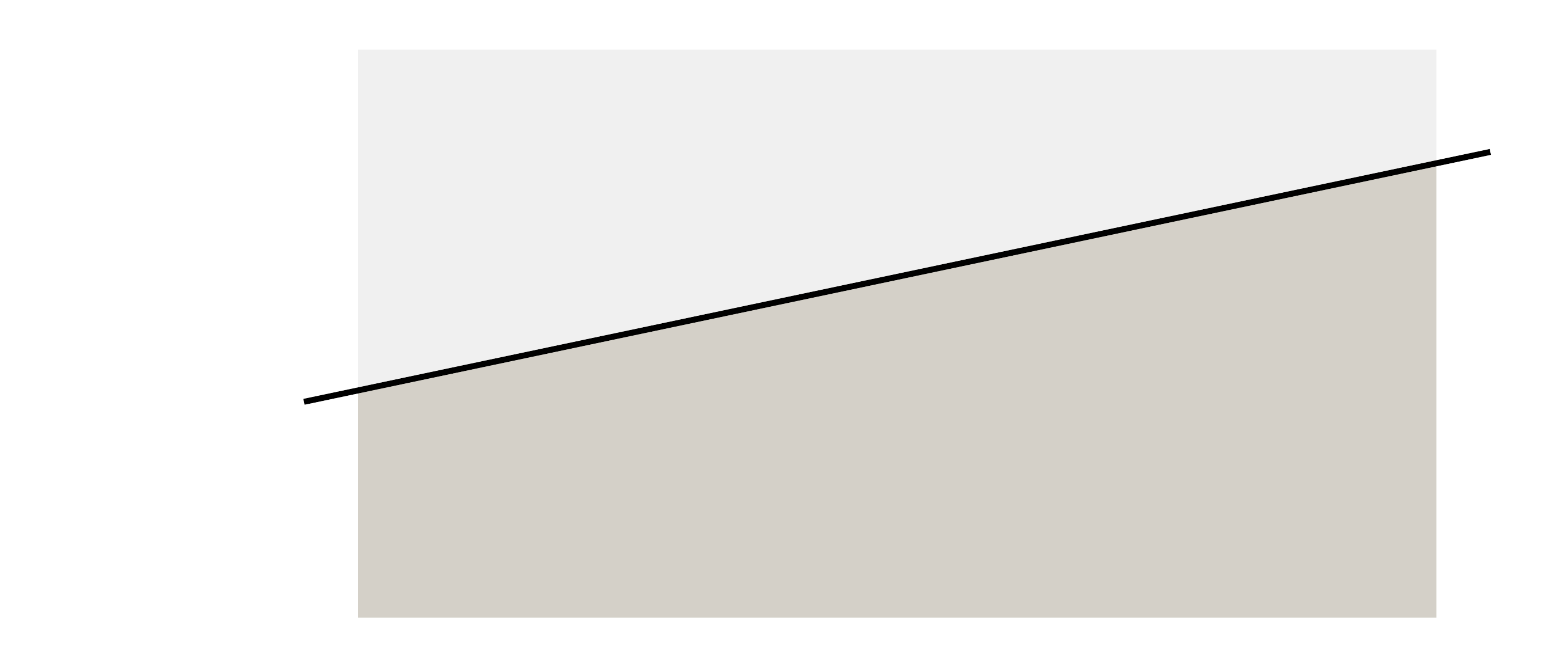}
 \put (35,34) {Region 1:\;\; $\boldsymbol{a}^T\!\balpha>b$} 
\put (45,10) {Region 2:\;\; $\boldsymbol{a}^T\!\balpha<b$} 
\put (1.5,12.3){\rotatebox{10}{Region 3:}}
\put (3.5,6.8){\rotatebox{10}{$\boldsymbol{a}^T\!\balpha=b$}}
 \end{overpic}
 \caption{Splitting the $\balpha$-domain into three regions relative to the hyperplane $\boldsymbol{a}^T\!\balpha=b$}\label{figprox}
\end{figure}

In this section we calculate the proximal operator for $g(\balpha) = \max(\Ba^T\balpha,b)$, based on the definition in (\ref{eqimp4}). Let's denote the minimizer of (\ref{eqimp4}) by $\boldsymbol{\rho}^*$. We consider breaking the $\balpha$-domain into three regions relative to the hyperplane $\boldsymbol{a}^T\!\balpha=b$ as depicted in Fig. \ref{figprox}. We then discuss the conditions under which $\boldsymbol{\rho}^*$ lands in each region. 
\subsubsection*{Region 1:}
In this region $g(\balpha) = \Ba^T\balpha$ and the proximal calculation reduces to ${\boldsymbol{\rho}}^* = \operatorname*{arg\,min}_{\balpha} \;\; \Ba^T\balpha + \frac{1}{2\xi}\|\balpha - {\boldsymbol{\rho}}\|^2$.
The minimizer of this quadratic program is simply ${\boldsymbol{\rho}}^* = {\boldsymbol{\rho}}-\xi\Ba$. The feasibility of ${\boldsymbol{\rho}}^*$ (i.e., the minimizer being in region 1) requires $\Ba^T({\boldsymbol{\rho}} - \xi\Ba)>b$ or simply $\Ba^T{\boldsymbol{\rho}}> b + \xi\Ba^T\Ba$.
\subsubsection*{Region 2:}
In this region the proximal program reduces to
$
{\boldsymbol{\rho}}^* = \operatorname*{arg\,min}_{\balpha} \;\; b + \frac{1}{2\xi}\|\balpha - {\boldsymbol{\rho}}\|^2,
$
which trivially yields ${\boldsymbol{\rho}}^* = {\boldsymbol{\rho}}$. The feasibility of ${\boldsymbol{\rho}}^*$ in this case requires $\Ba^T{\boldsymbol{\rho}} <b$.
\subsubsection*{Region 3:}
When $b\leq \Ba^T{\boldsymbol{\rho}}\leq b + \xi\Ba^T\Ba$, 
neither regions 1 or 2 could be in hold of ${\boldsymbol{\rho}}^*$, and the point must lie on the hyperplane $\Ba^T\balpha = b$. In other words
\begin{equation*}
{\boldsymbol{\rho}}^* = \operatorname*{arg\,min}_{\balpha} \;\; b + \frac{1}{2\xi}\|\balpha - {\boldsymbol{\rho}}\|^2\quad s.t.\quad \Ba^T\balpha = b.
\end{equation*}
To explicitly find ${\boldsymbol{\rho}}^*$, we neglect the constant term and the $\xi^{-1}$ factor, and form the Lagrangian as $L(\balpha,\mu) = \frac{1}{2}\|\balpha - {\boldsymbol{\rho}}\|^2 +\mu (\Ba^T\balpha - b)$. We denote $\mu^*$ as the optimal multiplier, where
$({\boldsymbol{\rho}}^*,\mu^*) = \operatorname*{arg\,min}_{(\balpha,\mu)}\;\;L(\balpha,\mu)$. Imposing the optimality conditions yields
\begin{equation}\label{eqlag1}
\left\{\begin{array}{l}\frac{\partial L}{\partial \balpha}\Big|_{(\balpha,\mu)=({\boldsymbol{\rho}}^*,\mu^*)}={\boldsymbol{\rho}}^*-{\boldsymbol{\rho}}+\mu^*\Ba=\boldsymbol{0}\\
\frac{\partial L}{\partial \mu}\Big|_{(\balpha,\mu)=({\boldsymbol{\rho}}^*,\mu^*)}=\Ba^T{\boldsymbol{\rho}}^* - b=0.
\end{array}.
\right.
\end{equation}
Solving (\ref{eqlag1}) for $({\boldsymbol{\rho}}^*,\mu^*)$ results in $\mu^* = (\Ba^T{\boldsymbol{\rho}}-b)/\|\Ba\|^2$ and ${\boldsymbol{\rho}}^*={\boldsymbol{\rho}}-(1/\|\Ba\|^2) \Ba\Ba^T{\boldsymbol{\rho}}+(b/\|\Ba\|^2) \Ba$. 

We can now combine the conditions and the minimizer evaluations for each region to summarize the proximal operator as presented in (\ref{prox_derived}).

\section{Proof of Theorem \ref{thunique}}
We first reformulate the main CSC problem in (\ref{eq5x}) by rewriting the objective in a separable form. This is performed through the variable change $\Beta=\boldsymbol{B}\balpha$ and using the fact that $\mathcal{L}_{\balpha}(x) = \sum_{i=1}^{n_\Omega}\beta_i\chi_{\Omega_i}(x)$:
\begin{align}\nonumber
G(\balpha) &=\int_{D}\max\Big(\Delta(x) \sum_{i=1}^{n_\Omega}\beta_i\chi_{\Omega_i}(x), \Delta(x)^-   \Big)\;\mbox{d}x\\ \nonumber &=
\sum_{i=1}^{n_\Omega}p_i\max(\beta_i,0)-q_i\min(\beta_i,1)\\ & \triangleq \mathcal{G}(\Beta) \label{eq43}
\end{align}
Taking into account the additional constraint $\Beta=\boldsymbol{B}\balpha$, the main optimization (\ref{eq5x}) can be written in $\mathbb{R}^{n_\Omega+n_s}$ as
\begin{equation}
\min_{(\Beta,\balpha)} \;\;\mathcal{G}(\Beta)\quad s.t. \quad      \left\{
     \begin{array}{l}
       \|\boldsymbol{\alpha}\|_1 \leq  \tau\\
       \boldsymbol{K} \begin{bmatrix}\Beta \\ \balpha \end{bmatrix} = \boldsymbol{0}
     \end{array}
   \right.,    \label{eqp0}
\end{equation}
where $\boldsymbol{K}=\begin{bmatrix}-\boldsymbol{I} & \boldsymbol{B}\end{bmatrix}$. Through a lemma, we first derive sufficient conditions under which a given pair $(\Beta^*,\balpha^*)$ uniquely minimizes (\ref{eqp0}). We then use this result to link the conditions stated in Theorem \ref{thunique} to the unique optimality.

\begin{lemma}\label{lemunique}
\emph{Consider $\Beta^*\in\mathbb{R}^{n_\Omega}$ such that $I_{(0,1)}(\Beta^*)=\emptyset$, and $\balpha^*\in\mathbb{R}^{n_s}$ such that $\|\balpha^*\|_1=\tau$, which together satisfy the linear constraint in (\ref{eqp0}). If there exist a scalar $\Lambda_{\alpha}>0$, and a vector $\Blambda\in \mathbb{R}^{n_\Omega+n_s}$ in the range of $\boldsymbol{K}^T$, such that
\begin{align}
\hspace{-.3cm}\left\{\begin{array}{lc}
\Blambda_\ell \in (-q_\ell,p_\ell-q_\ell) & \ell\in I_0(\Beta^*)\\
\Blambda_\ell \in (p_\ell-q_\ell,p_\ell) & \ell \in I_1(\Beta^*)\\
\Blambda_{\ell+n_\Omega} \in (-\Lambda_{\alpha},\Lambda_{\alpha}) & \ell \in  I_0(\balpha^*)
\end{array}
\right. ,\label{eqp1a}
\end{align}
\begin{align}
\left\{\begin{array}{lc}
\Blambda_\ell =p_\ell & \ell\in I_{(1,\infty)}(\Beta^*)\\
\Blambda_\ell =-q_\ell & \ell \in I_{(-\infty,0)}(\Beta^*)\\
\Blambda_{\ell+n_\Omega} =\Lambda_{\alpha} \sign(\balpha_\ell^*)  & \ell \in I_{\mathbb{R}\setminus\{0\}}(\balpha^*)
\end{array}
\right. \label{eqp1b}
\end{align}
and for
\begin{equation}\Gamma = I_{(-\infty,0)}(\Beta^*)\cup I_{(1,\infty)}(\Beta^*)\cup \big\{n_\Omega + I_{\mathbb{R}\setminus\{0\}}(\balpha^*)\big\}\label{eqp2}
\end{equation}
the matrix $\boldsymbol{K}_{:,\Gamma}$ is full column rank, then the pair $(\Beta^*,\balpha^*)$ is the unique minimizer of the convex program (\ref{eqp0}).}

\end{lemma}

\emph{Proof:}
Given $\Beta$ and $\balpha$, consider a new objective function $\mathcal{H}(.):\mathbb{R}^{n_\Omega+n_s}\to \mathbb{R}$:
\[\mathcal{H}\Big(\begin{bmatrix}\Beta\\ \balpha\end{bmatrix}\Big) = \mathcal{G}(\Beta)+ \mathbbit{I}_{\hspace{.02cm}\mathcal{C}_{\tau}\!}(\balpha),
\]
where $\mathbbit{I}_{\hspace{.02cm}\mathcal{C}_{\tau}\!}(\balpha)$ denotes the indicator function of the convex set $\mathcal{C}_{\tau}=\{\balpha\in\mathbb{R}^{n\!_s}:\;\|\balpha\|_1\leq {\tau}\}$, as defined in (\ref{eqimp8}).

The convex program (\ref{eqp0}) can be equivalently cast as
\begin{equation}\label{eqp3}
\min_{\Beta,\balpha} \;\; \mathcal{H}\Big(\begin{bmatrix}\Beta\\ \balpha\end{bmatrix}\Big) \quad s.t. \quad \boldsymbol{K}\begin{bmatrix}\Beta\\ \balpha\end{bmatrix} = \boldsymbol{0}.
\end{equation}
We first show that under the conditions specified in the lemma, $\Blambda\in \partial \mathcal{H}([\Beta^*;\balpha^*])$, that is, $\Blambda$ is a subgradient of $\mathcal{H}(.)$ at $[\Beta^*;\balpha^*]$.

For this purpose, we split $\Blambda$ into $\Blambda'\in \mathbb{R}^{n_\Omega}$ and $\Blambda''\in \mathbb{R}^{n_s}$ as follows:
\begin{equation}\label{eqp4}
\Blambda_\ell' = \Blambda_\ell, \quad \ell = 1,\cdots,n_\Omega;\;\; \Blambda_\ell'' = \Blambda_{\ell+n_\Omega}, \quad \ell = 1,\cdots,n_s.
\end{equation}
Conditions (\ref{eqp1a}) and (\ref{eqp1b}) on $\Blambda$ can be reinterpreted as 
\begin{align}\label{eqp5}
\left\{\begin{array}{lc}
\!\!\Blambda_\ell' \in (-q_\ell,p_\ell-q_\ell) &\hspace{-.15cm} \ell\in I_0(\Beta^*)\\
\!\!\Blambda_\ell' \in (p_\ell-q_\ell,p_\ell) &\hspace{-.15cm} \ell \in I_1(\Beta^*)
\end{array}
\right. \!\!,
\left\{\begin{array}{lr}
\!\!\Blambda_\ell' =p_\ell &\hspace{-.15cm} \ell\in I_{(1,\infty)}(\Beta^*)\\
\!\!\Blambda_\ell' =-q_\ell &\hspace{-.15cm} \ell \in I_{(-\infty,0)}(\Beta^*)
\end{array}
\right. 
\end{align}
and
\begin{equation}\label{eqp6}
\left\{\begin{array}{lc}
\Blambda_{\ell}'' \in (-\Lambda_{\alpha},\Lambda_{\alpha}) & \ell \in  I_0(\balpha^*)\\ \Blambda_{\ell}'' =\Lambda_{\alpha} \sign(\balpha_\ell^*)  & \ell \in I_{\mathbb{R}\setminus\{0\}}(\balpha^*)\end{array}\right. .
\end{equation}
In view of the separability of $\mathcal{H}(.)$ in $\Beta$ and $\balpha$, to show $\Blambda \in \partial \mathcal{H}([\Beta^*;\balpha^*])$ it suffices to show that $\Blambda'\in \partial \mathcal{G}(\Beta^*)$ and $\Blambda''\in \partial \mathbbit{I}_{\hspace{.02cm}\mathcal{C}_{\tau}\!}(\balpha^*)$.

Using basic subgradient calculus \cite{boyd2004convex}, it can be easily verified that $\Bz \in \partial \mathcal{G}(\Beta)$ when
\begin{align}\label{eqp7}
\left\{\begin{array}{lc}
\!\!\Bz_i \in [-q_i,p_i-q_i] &\hspace{-.1cm} i\in I_0(\Beta)\\
\!\!\Bz_i \in [p_i-q_i,p_i] &\hspace{-.1cm} i\in I_1(\Beta)
\end{array}
\right. \!,
\left\{\begin{array}{lr}
\!\!\Bz_i=p_i &\hspace{-.15cm} i\in I_{(1,\infty)}(\Beta)\\
\!\!\Bz_i=p_i - q_i &\hspace{-.15cm} i\in I_{(0,1)}(\Beta)\\
\!\!\Bz_i =-q_i &\hspace{-.15cm} i\in I_{(-\infty,0)}(\Beta)
\end{array}
\right.\!\!\!.
\end{align}
Since $I_{(0,1)}(\Beta^*)=\emptyset$, (\ref{eqp5}) implies that $\Blambda'\in \partial \mathcal{G}(\Beta^*)$.

For the $\balpha$-dependent term, $\partial \mathbbit{I}_{\hspace{.02cm}\mathcal{C}_{\tau}\!}(\balpha)$ can be represented by the normal cone of $\mathcal{C}_{\tau}$ at $\balpha$. More specifically \cite{rockafellar1979directionally}
\[\partial \mathbbit{I}_{\hspace{.02cm}\mathcal{C}_{\tau}\!}(\balpha) = \big\{\Bz \in \mathbb{R}^{n_s}: \; \Bz^T(\balpha'-\balpha)\leq 0, \forall \balpha'\in \mathcal{C}_{\tau} \big\}.
\]
To show $\Blambda''\in \partial \mathbbit{I}_{\hspace{.02cm}\mathcal{C}_{\tau}\!}(\balpha)$, clearly
\begin{align}\nonumber
{\Blambda''}^T\!\balpha^* &= \sum_{\ell =1}^{n_s} {\Blambda}''_\ell \balpha_\ell^* = \!\!\!\!\sum_{\ell\in I_{\mathbb{R}\setminus\{0\}}(\balpha^*)} \!\!\!\! \Lambda_{\alpha}\sign(\balpha_\ell^*)\balpha_\ell^* \\  & = \Lambda_{\alpha} \|\balpha^*\|_1 =\Lambda_{\alpha}{\tau},\label{eqp8}
\end{align}
and for an arbitrary vector $\balpha'\in \mathcal{C}_{\tau}$, we can use the H\"{o}lder's inequality to upper bound ${\Blambda''}^T\!\balpha'$ as
\begin{align}
{\Blambda''}^T\!\balpha' \:\leq\; |{\Blambda''}^T\!\balpha'| \;\leq\; \|{\Blambda''}\|_\infty \|\balpha'\|_1\;\leq\; \Lambda_{\alpha} {\tau}.\label{eqp9}
\end{align}
A direct result of (\ref{eqp8}) and (\ref{eqp9}) is ${\Blambda''}^T\!(\balpha' - \balpha^*)\leq 0$, which asserts that $\Blambda'' \in \partial \mathbbit{I}_{\hspace{.02cm}\mathcal{C}_{\tau}\!}(\balpha^*)$.

In order to prove the uniqueness of the minimizing pair, we show that any vector $[\hat\Beta;\hat\balpha] \neq [\Beta^*;\balpha^* ]$ that satisfies the linear constraint in (\ref{eqp3}) yields a greater cost:
\[\mathcal{H}\Big(\begin{bmatrix}\hat\Beta\\ \hat\balpha\end{bmatrix}\Big)>\mathcal{H}\Big(\begin{bmatrix}\Beta^*\\ \balpha^*\end{bmatrix}\Big).
\]
For this purpose, we proceed by defining a difference vector
\[\Bh=\begin{bmatrix}\hat\Beta\\ \hat\balpha\end{bmatrix} - \begin{bmatrix}\Beta^*\\ \balpha^*\end{bmatrix}.
\]
As a basic property of the subgradient \cite{boyd2004convex}, the cost values are related as
\begin{equation}\label{eqp10}
\forall \Bz\in\partial \mathcal{H}\Big(\begin{bmatrix}\Beta^*\\ \balpha^*\end{bmatrix}\Big): \qquad \mathcal{H}\Big(\begin{bmatrix}\hat\Beta\\ \hat\balpha\end{bmatrix}\Big)\geq \mathcal{H}\Big(\begin{bmatrix}\Beta^*\\ \balpha^*\end{bmatrix}\Big) + \boldsymbol{h}^T \Bz.
\end{equation}
Similar to (\ref{eqp4}), consider splitting $\boldsymbol{h}$ into $\boldsymbol{h}'\in \mathbb{R}^{n_\Omega}$ and $\boldsymbol{h}''\in \mathbb{R}^{n_s}$, and $\Bz$ into $\Bz'\in \mathbb{R}^{n_\Omega}$ and $\Bz''\in \mathbb{R}^{n_s}$. We make use of the following disjoint index sets (note that $I_{(0,1)}(\Beta^*)=\emptyset$):
\[\Gamma' = I_{(-\infty,0)}(\Beta^*)\cup I_{(1,\infty)}(\Beta^*),\;\;\mbox{and}\;\;{\Gamma'}^c = I_{0}(\Beta^*)\cup I_{1}(\Beta^*),
\]
and decompose $\Blambda'$ and $\Bz'$ as $\Blambda' = \mathcal{P}_{\Gamma'} (\Blambda') + \mathcal{P}_{{\Gamma'}^c} (\Blambda')$ and $\Bz' = \mathcal{P}_{\Gamma'} (\Bz') + \mathcal{P}_{{\Gamma'}^c} (\Bz')$. Here $\mathcal{P}_A(.)$ is the projection operator onto the index set $A$, i.e.,
\[ [\mathcal{P}_A(\boldsymbol{x})]_i = \left\{\begin{array}{lc}
\boldsymbol{x}_i & i\in A\\
0 & i\notin A
\end{array}
\right..
\]
From (\ref{eqp5}) and (\ref{eqp7}) we observe that $\mathcal{P}_{\Gamma'}(\Bz') = \mathcal{P}_{\Gamma'}(\Blambda')$. Subsequently, $\Bz' = \Blambda' - \mathcal{P}_{{\Gamma'}^c}(\Blambda') + \mathcal{P}_{{\Gamma'}^c} (\Bz')$  and 
\begin{align}\nonumber
\mathcal{H}\big(\![\hat\Beta;\hat\balpha]\!\big) & \!\geq  \mathcal{H}\big(\![\Beta^*\!\!;\!\balpha^*\!]\!\big)\!+\! \boldsymbol{h}^T \!\Bz \\ \nonumber & \!=\mathcal{H}\big(\![\Beta^*\!\!;\!\balpha^*\!]\!\big)\! +\! \boldsymbol{h'}^T \!\Bz' \!+\! \boldsymbol{h''}^T \!\Bz'' \\ \nonumber & \!= \mathcal{H}\big(\![\Beta^*\!\!;\!\balpha^*\!]\!\big) \!+\! \boldsymbol{h'}^T\! \!\Big(\!\Blambda' \!- \mathcal{P}_{{\Gamma'}^c}(\Blambda') \!+\! \mathcal{P}_{{\Gamma'}^c} (\!\Bz')\!\Big)\!+\!\boldsymbol{h''}^T \!\!\Bz''    \\ & \!=      \mathcal{H}\big(\![\Beta^*\!\!;\!\balpha^*\!]\!\big) \!+\! \boldsymbol{h'}^T \!\Big(  \mathcal{P}_{{\Gamma'}^c}(\!\Bz') \!-\!    \mathcal{P}_{{\Gamma'}^c}(\Blambda')\!\Big)\nonumber \\ &\hspace{1.85cm}\!+\! \boldsymbol{h''}^T (\!\Bz''\!-\!  \Blambda'').\label{eqp11}
\end{align}
To establish the last equality we used the fact that $\boldsymbol{K}\boldsymbol{h}=\boldsymbol{0}$ and $\Blambda$ is in the range of $\boldsymbol{K}^T$, hence, $\boldsymbol{h'}^T\Blambda' + \boldsymbol{h''}^T\Blambda'' = 0$. In the remainder of the proof, we show that by a careful selection of the subgradient vector $\Bz$, we can lower bound $\boldsymbol{h}^T \!\Bz$ by a positive quantity. 

For the proposed support definition in (\ref{eqp2}), the off-support elements are identified by
\[\Gamma^c = I_{0}(\Beta^*)\cup I_{1}(\Beta^*)\cup \big\{ n_\Omega + I_0(\balpha^*)\big\}.
\]
Based on the sign of the entries in $\boldsymbol{h}'$ and $\boldsymbol{h}''$, we can partition $\Gamma^c$ into disjoint index sets, denoted as follows:
\begin{align*}
\Gamma^c =& \underbrace{\big( I_0(\Beta^*) \cap I_{[0,\infty)}(\boldsymbol{h}')\big)}_{\mathcal{I}_{+}^{\beta,0}} \;\cup\; \underbrace{\big( I_0(\Beta^*) \cap I_{(-\infty,0)}(\boldsymbol{h}')\big)}_{\mathcal{I}_{-}^{\beta,0}}  \\[-.13cm]& \cup \; \underbrace{\big( I_1(\Beta^*) \cap I_{[0,\infty)}(\boldsymbol{h}')\big)}_{\mathcal{I}_{+}^{\beta,1}} \;\cup\; \underbrace{\big( I_1(\Beta^*) \cap I_{(-\infty,0)}(\boldsymbol{h}')\big)}_{\mathcal{I}_{-}^{\beta,1}} \\[-.13cm] & \cup \; \Big\{n_\Omega + \underbrace{\big( I_0(\balpha^*) \cap I_{[0,\infty)}(\boldsymbol{h}'')\big)}_{\mathcal{I}_{+}^{\alpha,0}}\Big\}\\[-.13cm] &\cup\; \Big\{n_\Omega + \underbrace{\big( I_0(\balpha^*) \cap I_{(-\infty,0)}(\boldsymbol{h}'')\big)}_{\mathcal{I}_{-}^{\alpha,0}}\Big\}.
\end{align*}
For the general inequality (\ref{eqp11}) being valid for any subgradient vector, we narrow our choice to a specific vector $\By = [\By'; \By'']\in \partial \mathcal{H}([\Beta^*;\balpha^*])$ such that $\By_\Gamma =\Blambda_\Gamma$ and
\begin{equation}\hspace{-.06cm}\By_\ell' = \left\{\begin{array}{lr}\!\!p_\ell - q_\ell & \ell\in \mathcal{I}_{+}^{\beta,0}\cup \mathcal{I}_{-}^{\beta,1}\\
\!\!- q_\ell & \ell\in \mathcal{I}_{-}^{\beta,0}\\
\!\!p_\ell & \ell \in \mathcal{I}_{+}^{\beta,1}
\end{array}
\right.\!\!, \By_\ell'' = \left\{\begin{array}{lr}\!\!\Lambda_{\alpha} & \ell\in \mathcal{I}_{+}^{\alpha,0}\\
\!\!- \Lambda_{\alpha} & \ell\in \mathcal{I}_{-}^{\alpha,0}
\end{array}
\right.\!\!.\label{eqp12}
\end{equation}
By referring to (\ref{eqp7}) and tracing a similar path as (\ref{eqp8}) and (\ref{eqp9}), it is straightforward to see that $\By$ meets all the criteria of being a subgradient vector at $[\Beta^*;\balpha^*]$. Moreover, since every entry of $\By$ restricted to $\Gamma^c$ is set to an extreme possible value, (\ref{eqp5}), (\ref{eqp6}) and (\ref{eqp12}) trivially reveal that
\[\left\{\begin{array}{cc}\By_\ell' - \Blambda_\ell'>0 & \ell\in  \mathcal{I}_{+}^{\beta,0}\cup \mathcal{I}_{+}^{\beta,1}\\
\By_\ell' - \Blambda_\ell'<0  & \ell\in  \mathcal{I}_{-}^{\beta,0}\cup \mathcal{I}_{-}^{\beta,1}\end{array}
\right.
, \left\{\begin{array}{cc}\By_\ell'' - \Blambda_\ell''>0 & \ell\in  \mathcal{I}_{+}^{\alpha,0}\\
\By_\ell'' - \Blambda_\ell''<0  & \ell\in  \mathcal{I}_{-}^{\alpha,0}\end{array}
\right..
\]
We can set a strictly positive quantity $\epsilon$ to be
\begin{align*}\epsilon = \min\Big\{\! &\{\By_\ell' - \Blambda_\ell'\!: \ell\in  \mathcal{I}_{+}^{\beta,0}\!\cup\! \mathcal{I}_{+}^{\beta,1}\}\cup \{ \Blambda_\ell' - \By_\ell'\!: \ell\in  \mathcal{I}_{-}^{\beta,0}\cup \mathcal{I}_{-}^{\beta,1}\}\\ & \cup \{\By_\ell'' - \Blambda_\ell'': \ell\in  \mathcal{I}_{+}^{\alpha,0} \}\cup \{ \Blambda_\ell'' - \By_\ell'': \ell\in  \mathcal{I}_{-}^{\alpha,0} \}\!\Big\},
\end{align*}
and lower bound the constituting components of $\boldsymbol{h}^T \By$ as 
\begin{align}\nonumber
\boldsymbol{h'}^T &\big(  \mathcal{P}_{{\Gamma'}^c}(\By') -    \mathcal{P}_{{\Gamma'}^c}(\Blambda')\big)  = \sum_{\ell\in {\Gamma'}^c}\boldsymbol{h}_\ell'(\By_\ell ' - \Blambda_\ell' )\\& = \sum_{\ell\in  \mathcal{I}_{+}^{\beta,0}\cup \mathcal{I}_{+}^{\beta,1}}\!\!\!\!\!\!\!\! \boldsymbol{h}_\ell'(\By_\ell ' - \Blambda_\ell' ) - \sum_{\ell\in  \mathcal{I}_{-}^{\beta,0}\cup \mathcal{I}_{-}^{\beta,1}}\!\!\!\!\!\!\!\! \boldsymbol{h}_\ell'(  \Blambda_\ell' - \By_\ell ' ) \nonumber\\& = \sum_{\ell\in {\Gamma'}^c}|\boldsymbol{h}_\ell'||\By_\ell ' - \Blambda_\ell' | \;\;\;\;\;\;\geq\;\; \epsilon \sum_{\ell\in {\Gamma'}^c}|\boldsymbol{h}_\ell'|,\label{eqp13}
\end{align}
and in a similar fashion
\begin{align}
\boldsymbol{h''}^T\!\! (\By''-  \Blambda'') = \!\!\!\!\!\sum_{\ell\in  \mathcal{I}_{+}^{\alpha,0}\cup \mathcal{I}_{-}^{\alpha,0} } \!\!\!|\boldsymbol{h''}_\ell||\By_\ell '' - \Blambda_\ell'' |\geq \epsilon\!\!\!\!\!\!\!\!\! \sum_{\ell\in  \mathcal{I}_{+}^{\alpha,0}\cup \mathcal{I}_{-}^{\alpha,0} } \!\!\!|\boldsymbol{h''}_\ell|.\label{eqp14}
\end{align}
Now, in view of (\ref{eqp13}) and (\ref{eqp14})  
\begin{align*}
\boldsymbol{h'}^T \Big(\!  \mathcal{P}_{{\Gamma'}^c}(\By') -    \mathcal{P}_{{\Gamma'}^c}(\Blambda')\!\Big)+ \boldsymbol{h''}^T (\By''-  \Blambda'') &= \!\!\!\sum_{\ell\in  \Gamma^c} |\boldsymbol{h}_\ell||\By_\ell  - \Blambda_\ell |\\ \nonumber & \geq \epsilon \|\boldsymbol{h}_{\Gamma^c}\|_1,
\end{align*}
which yields
\begin{equation}\label{eqp14-1}
\mathcal{H}\Big(\begin{bmatrix}\hat\Beta\\ \hat\balpha\end{bmatrix}\Big)\geq \mathcal{H}\Big(\begin{bmatrix}\Beta^*\\ \balpha^*\end{bmatrix}\Big) + \epsilon \|\boldsymbol{h}_{\Gamma^c}\|_1.
\end{equation}
Since $\epsilon > 0$, we always have $\mathcal{H}\big([\hat\Beta;\hat\balpha]\big)  >  \mathcal{H}\big([\Beta^*;\balpha^*]\big)$ except when $\boldsymbol{h}_{\Gamma^c}=\boldsymbol{0}$ that an equality can happen. As $\boldsymbol{K}_{:,\Gamma}$ has full column rank, the linear equation $\boldsymbol{K}\boldsymbol{h}=\boldsymbol{0}$ subject to $\boldsymbol{h}_{\Gamma^c}=\boldsymbol{0}$ has the unique solution $\boldsymbol{h}=\boldsymbol{0}$. In other words, the cost equality only happens when $[\hat\Beta;\hat\balpha]=  [\Beta^*;\balpha^*]$. Thanks to the equivalence of (\ref{eqp0}) and (\ref{eq5x}), the conditions stated in Lemma \ref{lemunique} also warrant the unique optimality of $\balpha^*$ for the original program in (\ref{eq5x}).$\square$

In the sequel, we show the conditions stated in Theorem \ref{thunique} can be directly linked to those imposed by Lemma \ref{lemunique}, which in turn establish the unique optimality of $\balpha^*$.

To show the full column rank property of $\boldsymbol{K}_{:,\Gamma}$ we have
\[\boldsymbol{K}_{:,\Gamma} = \begin{bmatrix}-\boldsymbol{I}_{:,\Gamma_{0^-}\cup\Gamma_{1^+}}& \boldsymbol{B}_{:,\Gamma_{\!\balpha^*}}
  \end{bmatrix}.
\]
By applying basic rank preserving operations we can reform $\boldsymbol{K}_{:,\Gamma}$ as
\[\boldsymbol{K}_{:,\Gamma} \leftrightarrow \begin{bmatrix}-\boldsymbol{I} & \boldsymbol{0}\\ \boldsymbol{0}&\boldsymbol{B}_{\Gamma_0\cup\Gamma_1,\Gamma_{\!\balpha^*}}
  \end{bmatrix},
\]
where the identity block has a width $|\Gamma_{0^-}\cup\Gamma_{1^+}|$. Clearly, if $\boldsymbol{B}_{\Gamma_0\cup\Gamma_1,\Gamma_{\!\balpha^*}}$ has a full column rank, $\boldsymbol{K}_{:,\Gamma'}$ will also be full column rank.

Further, suppose there exist $\boldsymbol{\eta}$ and $\eta_c$ that satisfy (\ref{eqthunique1}). We introduce a vector $\tilde{\boldsymbol{\eta}}$, the entries of which are set to 
\begin{equation}\label{eqp15}
\tilde{\boldsymbol{\eta}}_\ell  = \left \{\begin{array}{ll}   q_\ell & \ell\in \Gamma_{0^-} \\  -p_\ell & \ell\in \Gamma_{1^+} \\   \boldsymbol{\eta}_{m(\ell)} & \ell\in \Gamma_{0} \cup \Gamma_{1}\end{array} \right. .
\end{equation}
We assert that by setting $\boldsymbol{\Lambda} = \boldsymbol{K}^T\tilde{\boldsymbol{\eta}}$ and $\Lambda_\alpha = \eta_c$, the uniqueness criteria in Lemma \ref{lemunique} are met. Clearly,
\begin{equation*}
\boldsymbol{K}^T\tilde{\boldsymbol{\eta}} = \begin{bmatrix} -\boldsymbol{I}\\\boldsymbol{B}^T \end{bmatrix}\tilde{\boldsymbol{\eta}} = \begin{bmatrix} - \tilde{\boldsymbol{\eta}} \\ \boldsymbol{B}^T \tilde{\boldsymbol{\eta}} \end{bmatrix}.
\end{equation*}
For the first block of $\boldsymbol{K}^T\tilde{\boldsymbol{\eta}}$, by the construction (\ref{eqp15}), it is straightforward to verify that $- \tilde{\boldsymbol{\eta}}_\ell$ satisfies the conditions specified in (\ref{eqp1a}), (\ref{eqp1b}) for $\ell=1,2,\cdots,n_\Omega$ and we are only left to show
\begin{equation}\label{eqp16}
(\boldsymbol{B}^T \tilde{\boldsymbol{\eta}})_j\left \{\begin{array}{lc} \in (-\eta_c,\eta_c) & j \in I_0(\balpha^*)\\ =\eta_c\sign(\balpha_j^*) & j \in I_{\mathbb{R}\setminus\{0\}}(\balpha^*) \end{array}\right. .
\end{equation}
However, 
\begin{align*}\nonumber
 \boldsymbol{B}^T \tilde{\boldsymbol{\eta}}  &=  \big(\boldsymbol{B}_{\Gamma_1\cup\Gamma_0,:}\big)^T \tilde{\boldsymbol{\eta}}_{\Gamma_1\cup\Gamma_0} + \big(\boldsymbol{B}_{\Gamma_{1^+}\cup\Gamma_{0^-},:}\big)^T\tilde{\boldsymbol{\eta}}_{\Gamma_{1^+}\cup\Gamma_{0^-}} \\\nonumber  & =  \big(\boldsymbol{B}_{\Gamma_1\cup\Gamma_0,:}\big)^T \boldsymbol{\eta} \qquad\;\;+ \sum_{\ell\in \Gamma_{1^+}\cup\Gamma_{0^-}}\tilde{\boldsymbol{\eta}}_\ell \big(\boldsymbol{B}_{\ell,:}\big)^T \\[-.25 cm] &= \eta_c\boldsymbol{c},
\end{align*}
where the last equality is thanks to the assumption (\ref{eqthunique1}) and $\sum_{\ell\in \Gamma_{1^+}\cup\Gamma_{0^-}}\tilde{\boldsymbol{\eta}}_\ell \big(\boldsymbol{B}_{\ell,:}\big)^T=-\boldsymbol{e}$ (a detailed derivation of this equality is presented in the proof of Theorem 4.7 in \cite{aghasi2015convex}).
Subsequently, $\eta_c\boldsymbol{c}$ satisfies (\ref{eqp16}) and the proof is complete.

\section{Proof of Theorem \ref{thglast}}
We will follow the notational conventions stated in Section \ref{sec:accrecovery} throughout the proof. 

Accordingly, suppose that $\Beta^*\in \mathbb{R}^{n_\Omega}$ contains the values of $\mathcal{L}_{\balpha^*}(x)$ over the cells $\{\Omega_i\}_{i=1}^{n_\Omega}$. Let $T$ denote the index set defined in (\ref{Tdef}). By looking at the image of $\balpha^*$ in the $\Beta$-domain, we expect $\Beta^*$ to match the values of $\mathcal{L}_{\balpha_\mathpzc{R}}(x)$ over the cells within $\bigcup_{j=1}^{n_\oplus+n_\ominus} \mathcal{S}_j$ and to vanish over the remaining cells. More specifically,
\begin{equation*}
 \Beta^*_i = \left \{\begin{array}{lc}
 \Beta_\ell^\mathpzc{R} & i\in \mathcal{J}_\ell , \quad \ell \in \Gamma_0^\mathpzc{R}, \Gamma_1^\mathpzc{R}, \Gamma_{0^-}^\mathpzc{R}, \Gamma_{1^+}^\mathpzc{R}\\
 0 & i\in T
\end{array}.
\right.
\end{equation*}
Accordingly, the index sets of the unit and null-valued cells associated with $\Beta^*$ may be indicated by
\begin{equation*}
\Gamma_0 = T\cup \bigcup_{\ell\in \Gamma_0^\mathpzc{R}} \mathcal{J}_\ell \qquad \mbox{and}\qquad \Gamma_1 =  \bigcup_{\ell\in \Gamma_1^\mathpzc{R}} \mathcal{J}_\ell.
\end{equation*}

Let us assume the cell index assignment is performed in a way that $\Gamma_0\cup\Gamma_1 = \{1,2,\cdots, |\Gamma_0\cup\Gamma_1|\}$. Such assumption would avoid index mapping complications. Now, consider the binary matrix $\boldsymbol{B}'\in\{0,1\}^{(n_\oplus+n_\ominus)\times|\Gamma_0\cup\Gamma_1|}$ constructed as
\begin{equation*}
\boldsymbol{B}_{j,i}'= 1_{\{\Omega_i\subset\mathcal{S}_j\}} = \left\{\begin{array}{lc}\!1& \Omega_i\subset\mathcal{S}_j\\ \! 0 & \Omega_i\nsubset\mathcal{S}_j \end{array} \right.\!\!\!,\; j=1,\cdots, n_\oplus+n_\ominus, \; i\in \Gamma_0\cup\Gamma_1,
\end{equation*}
and follow a similar pattern for the exterior shapes to construct a matrix $\boldsymbol{B}''$ as
\begin{equation*}
\boldsymbol{B}_{j-n_\oplus-n_\ominus,i}''= 1_{\{\Omega_i\subset\mathcal{S}_j\}}, \qquad j=n_\oplus+n_\ominus+1, \cdots,  n_s, \quad i\in \Gamma_0\cup\Gamma_1.
\end{equation*}

Using Theorem \ref{thunique}, aside from a rank requirement, we need to verify the possibility of finding a vector $\boldsymbol{\eta}\in\mathbb{R}^{|\Gamma_0\cup\Gamma_1|}$ and a scalar $\eta_c>0$ such that
\begin{equation}\label{eq-g1}\begin{bmatrix}\boldsymbol{B}'\\ \boldsymbol{B}''\end{bmatrix}\boldsymbol{\eta} = \eta_c\boldsymbol{c}+\boldsymbol{e},
\end{equation}
where $\boldsymbol{c}_{\Ip} = \boldsymbol{1}$, $\boldsymbol{c}_{\In} = -\boldsymbol{1}$ and $\|\boldsymbol{c}_{(\Ip\cup\In)^c}\|_\infty < 1$. The entries of $\boldsymbol{\eta}$ need to satisfy
\begin{equation}\label{eqeta}
\left\{\begin{array}{lc}  q_i-p_i <\boldsymbol{\eta}_i <q_i & i\in \Gamma_0\\ -p_i <\boldsymbol{\eta}_i <q_i-p_i & i\in \Gamma_1
\end{array}\right..
\end{equation}

Regarding the rank requirement, the columns of $\boldsymbol{B}'$ are zero over $T$, and we can easily verify that the remaining columns are multiple replications of the columns of $(\boldsymbol{B}_{\Gamma_0^\mathpzc{R}\cup\Gamma_1^\mathpzc{R},:}^\mathpzc{R})^T$. In other words \vspace{-.25cm}
\begin{equation*}
\rank(\boldsymbol{B}') = \rank(\boldsymbol{B}_{\Gamma_0^\mathpzc{R}\cup\Gamma_1^\mathpzc{R},:}^\mathpzc{R}) = n_\oplus + n_\ominus,
\end{equation*}
where the last equality is thanks to the full-rank property of the underlying matrix (see the proof of Proposition \ref{lemcons} in \cite{aghasi2015convex}). As a result, the rank requirement by Theorem \ref{thunique} is automatically satisfied.

Clearly, the strict inequalities in (\ref{eqeta}) require us to have $p_i>0$ for $i\in\Gamma_0$ and $q_i>0$ for $i\in\Gamma_1$. We choose $\epsilon$ to be a number in the interval $(0,\epsilon_T)$, where
\begin{equation}
\epsilon_T = \inf \;\big\{p_i: i\in T\big\}\cup \big\{p_i-q_i: q_i< p_i, i\in T\big\}.
\end{equation}
Using the definition of $\boldsymbol{\varepsilon}^{LV}$ in (\ref{locviol}) and following the requirements imposed by (\ref{eq-g1}) and (\ref{eqeta}), we suggest the following $\boldsymbol{\eta}$ for the certificate of duality:
\begin{equation}\label{eq-g3}
 \boldsymbol{\eta}_i = \left\{\begin{array}{cll}
 \frac{1}{|\mathcal{J}_\ell|}(\eta_c\boldsymbol{w}_\ell+  \boldsymbol{\varepsilon}^{LV}_\ell) & i\in \mathcal{J}_\ell,& \ell \in \Gamma_0^\mathpzc{R}, \Gamma_1^\mathpzc{R}\\[.1cm]
-\epsilon  & i\in T,& q_i< p_i\\
q_i-p_i+\epsilon  & i\in T, &q_i\geq p_i
\end{array}
\right. .
\end{equation}
A simple calculation shows that under (\ref{eqcond}), by setting
\begin{equation}
\epsilon_i = \left\{\begin{array}{rll}
 \frac{1}{|\boldsymbol{w}_\ell|}(  \boldsymbol{\varepsilon}^{LV}_\ell - q_i|\mathcal{J}_\ell|) & i\in \mathcal{J}_\ell,& \ell \in \Gamma_0^\mathpzc{R}\\
-\frac{1}{|\boldsymbol{w}_\ell|}(  \boldsymbol{\varepsilon}^{LV}_\ell + p_i|\mathcal{J}_\ell|) & i\in \mathcal{J}_\ell,& \ell \in \Gamma_1^\mathpzc{R}
 \end{array}
\right.
\end{equation}
and setting $\eta_c> \max \big\{\epsilon_i \big\}_{i\in\mathcal{J}_\ell, \ell\in \Gamma_1^\mathpzc{R}\cup\Gamma_0^\mathpzc{R}}$, the quantities $\boldsymbol{\eta}_i$, $i\in \mathcal{J}_\ell, \ell \in \Gamma_0^\mathpzc{R}\cup  \Gamma_1^\mathpzc{R}$, are confined within the designated bounds imposed by (\ref{eqeta}). Similarly, when $\epsilon \in(0,\epsilon_T)$, verifying that for $i\in T$, $\boldsymbol{\eta}_i$ is in agreement with (\ref{eqeta}) is straightforward.

To demonstrate that (\ref{eq-g1}) holds, for $j\in \Ip$ we have
\begin{align}\label{eq-g4}
\boldsymbol{B}_{j,:}'\boldsymbol{\eta} &=\!\! \sum_{i\in \Gamma_0\cup\Gamma_1}\!\! \boldsymbol{\eta}_i 1_{\{\Omega_i\subset\mathcal{S}_j\}}  \\&=\nonumber \!\sum_{i\in\mathcal{J}_\ell}   \sum_{\ell\in \Gamma_0^\mathpzc{R}\cup \Gamma_1^\mathpzc{R}} \! \big(\frac{\eta_c}{|\mathcal{J}_\ell|}\boldsymbol{w}_\ell+\frac{\boldsymbol{\varepsilon}^{LV}_\ell}{|\mathcal{J}_\ell|}\big)1_{\{\Omega_i\subset\mathcal{S}_j\}}  + \sum_{i\in T} \boldsymbol{\eta}_i1_{\{\Omega_i\subset\mathcal{S}_j\}}.
\end{align}
The second term on the right hand side of (\ref{eq-g4}) is simply zero. Using the fact that $1_{\{\Omega_i\subset\mathcal{S}_j\}} = 1_{\{\Omega_i\subset\Omega_\ell^\mathpzc{R}\}} 1_{\{\Omega_\ell^\mathpzc{R}\subset\mathcal{S}_j\}}$ we get
\begin{align*}
\boldsymbol{B}_{j,:}'\boldsymbol{\eta} &=   \sum_{\ell\in \Gamma_0^\mathpzc{R}\cup \Gamma_1^\mathpzc{R}} \big(\eta_c\boldsymbol{w}_\ell+ \boldsymbol{\varepsilon}^{LV}_\ell \big)      1_{\{\Omega_\ell^\mathpzc{R}\subset\mathcal{S}_j\}}  \frac{1}{|\mathcal{J}_\ell|}\sum_{i\in\mathcal{J}_\ell}   1_{\{\Omega_i\subset\Omega_\ell^\mathpzc{R}\}}
\\&= \eta_c \sum_{\ell\in \Gamma_0^\mathpzc{R}\cup \Gamma_1^\mathpzc{R}} 1_{\{\Omega_\ell^\mathpzc{R}\subset\mathcal{S}_j\}} \boldsymbol{w}_\ell + \sum_{\ell\in \Gamma_0^\mathpzc{R}\cup \Gamma_1^\mathpzc{R}} 1_{\{\Omega_\ell^\mathpzc{R}\subset\mathcal{S}_j\}}\boldsymbol{\varepsilon}^{LV}_\ell  \\ &=   \eta_c \boldsymbol{w}^T \boldsymbol{B}_{\Gamma_0^\mathpzc{R}\cup \Gamma_1^\mathpzc{R},j}^\mathpzc{R} + {\boldsymbol{\varepsilon}^{LV}}^T \boldsymbol{B}_{\Gamma_0^\mathpzc{R}\cup \Gamma_1^\mathpzc{R},j}^\mathpzc{R} \\& =\eta_c+\boldsymbol{e}_j.
\end{align*}
A similar line of argument shows that $\boldsymbol{B}_{j,:}'\boldsymbol{\eta} = -\eta_c+\boldsymbol{e}_j$ for $j\in \In$.

We are only left to show the possibility of finding $\boldsymbol{c}_j \in (-1,1)$ such that
\begin{align*}
\boldsymbol{B}_{j-n_\oplus-n_\ominus,:}''\boldsymbol{\eta} = \eta_c \boldsymbol{c}_j  + \boldsymbol{e}_j, \qquad j=n_\oplus+n_\ominus +1, \cdots, n_s.
\end{align*}
By setting $\boldsymbol{c}_j =  (\eta_c)^{-1}(\boldsymbol{B}_{j-n_\oplus-n_\ominus,:}''\boldsymbol{\eta}-\boldsymbol{e}_j)$, we have
\begin{align*}
| \boldsymbol{c}_j| &= \frac{1}{\eta_c}\Big|-\boldsymbol{e}_j+ \sum_{i\in\Gamma_0\cup\Gamma_1}\boldsymbol{B}_{j-n_\oplus-n_\ominus,i}''\boldsymbol{\eta}_i \Big|\\
&= \frac{1}{\eta_c}\Big|-\boldsymbol{e}_j+
\sum_{\ell\in \Gamma_0^\mathpzc{R}\cup \Gamma_1^\mathpzc{R}} \big(   \frac{\boldsymbol{\varepsilon}^{LV}_\ell}{|\mathcal{J}_\ell|} + \frac{\eta_c}{|\mathcal{J}_\ell|}\boldsymbol{w}_\ell \big)    \sum_{i\in\mathcal{J}_\ell}   1_{\{\Omega_i\subset\mathcal{S}_j\}}
\\& \hspace{.95cm}+ \sum_{i\in T} \boldsymbol{\eta}_i 1_{\{\Omega_i\subset\mathcal{S}_j\}} \Big|
\\
& = \frac{1}{\eta_c}\Big|-\boldsymbol{e}_j
+   \sum_{\ell\in \Gamma_0^\mathpzc{R}\cup \Gamma_1^\mathpzc{R}} \gamma_{\ell,j}\boldsymbol{\varepsilon}^{LV}_\ell
+ \eta_c\sum_{\ell\in \Gamma_0^\mathpzc{R}\cup \Gamma_1^\mathpzc{R}} \gamma_{\ell,j}\boldsymbol{w}_\ell
\\
&
     \qquad\quad    +
 \sum_{i\in T} (q_i-p_i)1_{\{\Omega_i\subset\mathcal{S}_j,q_i> p_i\}} +\epsilon \sum_{i\in T} 1_{\{\Omega_i\subset\mathcal{S}_j,q_i\geq p_i\}} \\&\hspace{1.05cm} -\epsilon \sum_{i\in T} 1_{\{\Omega_i\subset\mathcal{S}_j,q_i<p_i\}}
  \Big|,   
\end{align*}  
and therefore, 
\begin{align*}
      | \boldsymbol{c}_j| &\leq \frac{1}{\eta_c}\Big|
        -\boldsymbol{e}_j
        +   \sum_{\ell\in \Gamma_0^\mathpzc{R}\cup \Gamma_1^\mathpzc{R}} \gamma_{\ell,j}\boldsymbol{\varepsilon}^{LV}_\ell\Big| + \Big|  \sum_{\ell\in \Gamma_0^\mathpzc{R}\cup \Gamma_1^\mathpzc{R}} \gamma_{\ell,j}\boldsymbol{w}_\ell  \Big|
 \\
&
     \quad\;  +
\frac{1}{\eta_c} \sum_{i\in T} (q_i-p_i)1_{\{\Omega_i\subset\mathcal{S}_j,q_i> p_i\}}  + \frac{\epsilon}{\eta_c} \sum_{i\in T} 1_{\{\Omega_i\subset\mathcal{S}_j\}}.
\end{align*}
Clearly, $\sum_{i\in T} (q_i-p_i)1_{\{\Omega_i\subset\mathcal{S}_j,q_i> p_i\}} = \sum_{i\in T\cap\mathcal{I}_j} (q_i-p_i)^+$.
Now under the assumption
\begin{align*}
\Big|  \sum_{\ell\in \Gamma_0^\mathpzc{R}\cup \Gamma_1^\mathpzc{R}} \gamma_{\ell,j}\boldsymbol{w}_\ell  \Big| < 1 -  \frac{1}{\eta_c}\bigg(\Big|
        -\boldsymbol{e}_j
       &+   \sum_{\ell\in \Gamma_0^\mathpzc{R}\cup \Gamma_1^\mathpzc{R}} \gamma_{\ell,j}\boldsymbol{\varepsilon}^{LV}_\ell\Big|  \\& + \sum_{i\in T\cap\mathcal{I}_j} (q_i-p_i)^+\bigg),
\end{align*}
we can select $\epsilon$ to be sufficiently small to assure $|\boldsymbol{c}_j|<1$. In other words, the condition posed by (\ref{eqcoh}) could be met by 
choosing $\delta_j$ as in (\ref{eqdeltaj}).

%
%



%

\bibliographystyle{IEEEtran}

\end{document}